\documentclass[10pt,journal,compsoc]{IEEEtran}
\ifCLASSOPTIONcompsoc
  \usepackage[nocompress]{cite}
\else
  \usepackage{cite}
\fi

\usepackage{microtype}
\usepackage{graphicx}
\usepackage{bm}
\usepackage{algorithm}
\usepackage{algorithmic}
\usepackage{multirow}
\usepackage{amsthm}
\usepackage{amsmath}
\usepackage{subfigure}
\usepackage{amssymb}
\usepackage{booktabs} 

\usepackage{balance}
\usepackage{float}
\usepackage{subeqnarray}
\usepackage{cases}
\usepackage{tabularx}
\usepackage{pifont}
\usepackage{makecell}
\usepackage{textcomp}
\usepackage{xcolor}

\def\x{\bm{\lambda}}
\def\y{\bm{w}}

\def\s{\mathbf{s}}
\def\t{\mathbf{t}}
\def\D{\mathcal{D}}
\def\E{\mathcal{E}}
\begin{document}
%
\title{Learning Deformable Image Registration \\ from Optimization: Perspective, Modules, \\ Bilevel Training and Beyond}
%
%
%
%

\author{Risheng~Liu,~\IEEEmembership{Member,~IEEE,}
	Zi~Li,
	Xin~Fan,~\IEEEmembership{Senior Member,~IEEE,}
	Chenying~Zhao,
	Hao~Huang,
	and~Zhongxuan~Luo
	\IEEEcompsocitemizethanks{
	    \IEEEcompsocthanksitem R. Liu, Z. Li, X. Fan and Z. Luo are with the DUT-RU International School of Information Science $\&$ Engineering and the Key Laboratory for Ubiquitous Network and Service Software of Liaoning Province, Dalian University of Technology, Dalian 116024, China. Z. Luo is also with the Institute of Artificial Intelligence, Guilin University of Electronic Technology, Guilin 541004, China. (E-mail: rsliu@dlut.edu.cn; alisonbrielee@gmail.com; xin.fan@ieee.org; zxluo@dlut.edu.cn). (Corresponding author: X. Fan). 
	    \IEEEcompsocthanksitem C. Zhao and H. Huang are with the Department of Radiology, Children's Hospital of Philadelphia, Philadelphia, PA, United States. C. Zhao is also with the Department of Bioengineering, School of Engineering and Applied Science, University of Pennsylvania, Philadelphia, PA, United States. H. Huang is also with the Department of Radiology, Perelman School of Medicine, University of Pennsylvania, Philadelphia, PA, United States (E-mail: chenyzh@seas.upenn.edu; huangh6@email.chop.edu).
    }}

\markboth{Journal of \LaTeX\ Class Files,~Vol.~14, No.~8, August~2015}%
{Shell \MakeLowercase{\textit{et al.}}: Bare Demo of IEEEtran.cls for Computer Society Journals}
%


\IEEEtitleabstractindextext{%
\begin{abstract}
	Conventional deformable registration methods aim at solving an optimization model carefully designed on image pairs and their computational costs are exceptionally high. In contrast, recent deep learning-based approaches can provide fast deformation estimation.
	These heuristic network architectures are fully data-driven and thus lack explicit geometric constraints which are indispensable to generate plausible deformations, e.g., topology-preserving.
	Moreover, these learning-based approaches typically pose hyper-parameter learning as a black-box problem and require considerable computational and human effort to perform many training runs. 
	To tackle the aforementioned problems, we propose a new learning-based framework to optimize a diffeomorphic model via multi-scale propagation. Specifically, we introduce a generic optimization model to formulate diffeomorphic registration and develop a series of learnable architectures to obtain propagative updating in the coarse-to-fine feature space. Further, we propose a new bilevel self-tuned training strategy, allowing efficient search of task-specific hyper-parameters. This training strategy increases the flexibility to various types of data while reduces computational and human burdens.
	We conduct two groups of image registration experiments on 3D volume datasets including image-to-atlas registration on brain MRI data and image-to-image registration on liver CT data. Extensive results demonstrate the state-of-the-art performance of the proposed method with diffeomorphic guarantee and extreme efficiency. 
	We also apply our framework to challenging multi-modal image registration, and investigate how our registration to support the down-streaming tasks for medical image analysis including multi-modal fusion and image segmentation.
\end{abstract}


}
\maketitle

\IEEEdisplaynontitleabstractindextext

%
\IEEEpeerreviewmaketitle

\IEEEraisesectionheading{\section{Introduction}\label{sec:introduction}}

\IEEEPARstart{R}{egistration} plays a critical role in medical image analysis, which transforms different images into one common coordinate system with matched contents by finding the spatial correspondence between images~\cite{MaintzV98}. It is fundamental to many clinical tasks such as image fusion of different modalities, anatomical change diagnosis, motion extraction, and population modeling. 
Traditional image registration is formulated as an optimization problem to minimize image mismatching between a target and a warped source image, subject to transformation constraints. 
Deformable registration method computes a dense correspondence between image pairs~\cite{SotirasDP13}. The high degrees of freedom for the solution space (deformation maps) and great variations on source/target image pairs are major challenges for this issue.

Conventional deformable registration techniques aim at solving the optimization problem and offer rigorous theoretical treatments. 
However, the optimization is typically computationally expensive and time-consuming as the iterative process involves gradient computations over the high dimensional parameter and image spaces~\cite{Ashburner07, BegMTY05,Avants08,SunNK14}.
Recent learning-based methods replace the costly numerical optimization with one step of prediction by learned deep networks so that they can provide fast deformation estimation. 
Balakrishnan~\emph{et~al.} propose a UNet network structure, called VoxelMorph, to address deformable image registration~\cite{BalakrishnanZSG19}.
Later, some researchers further combine the learning with the diffeomorphic constraint to provide topology-preserving deformations~\cite{DalcaBGS19, ShenHXN19}. 
These methods mostly learn parameters upon a pre-defined training loss to output deformation fields. Hence, it is difficult to adaptively include registration information for the front-end feature learning phase.
Some existing coarse-to-fine approaches~\cite{LiuLZFL20,WangZ20}, predicting deformation fields at different scales, may lead to more accurate registration and a controllable training procedure compared to those single scale approaches.

Regularization on hyper-parameters,~\emph{e.g.,} trade-off parameters, weight decay, and dropout, are crucial to the generalization of registration networks~\cite{BalakrishnanZSG19,DalcaBGS19}. The quality of output deformation fields for different deep networks highly depends on the choice of hyper-parameters. However, hyper-parameter choosing typically involves training many separate models with various hyper-parameter configurations, posing a significant computational challenge and potentially leading to sub-optimal results. For example, the grid search and random search work well only when ample computational resources are available. Generally, previous learning-based registration approaches pose hyper-parameter optimization as a black-box optimization problem, 
%
%
ignoring information that is important for faster convergence, thus require many training runs. 

\subsection{Our Contributions}
To address the limitations of both optimization-based and learning-based approaches, we design a new deep propagation framework to optimize a diffeomorphic model via multi-scale propagation for deformable registration.
First, we introduce a generic optimization model to formulate the diffeomorphic deformation problem. Rather than performing the optimization over the image domain, we learn a more discriminative feature space that handles deformations more powerfully.
Then we employ deep modules to propagate deformation fields on the learned multi-scale feature space, efficiently optimizing the diffeomorphic energy. This optimization perspective differentiates our scheme from naively cascading deep networks in most existing learning-based approaches, and provides a computational interpretation of network architectures that guarantees diffeomorphism.
Moreover, to tackle the inefficiency of hyper-parameter tuning, we introduce a bilevel self-tuned training strategy for our registration model.
This bilevel training takes the hyper-parameter learning as the upper-level objective while formulates learning for model parameters as the lower-level objective.
With the help of upper-level and lower-level objectives, the model parameters and hyper-parameters can be obtained collaboratively.
The main contributions of this work can be summarized in the following aspects:
\begin{enumerate}
	\item We establish a deep propagation framework to optimize the diffeomorphic registration energy on the learned multi-scale feature space. Circumventing expensive computations of iterative gradients on the image domain and performing the optimization over the discriminative feature space, this framework renders fast and efficient registration.
	\item We develop the error-based data matching, context-based regularization and constraint modules to yield the propagating process and to design the training loss. Each module has physical inspiration or geometrical priors so that helps to obtain the solutions stably and controllably. Moreover, we may interpret our learning-based registration as the optimization of an energy with explicit diffeomorphism constraints.
	\item We devise a new self-tuned training strategy that simultaneously learns optimal hyper-parameters of the loss function and network parameters of deep modules. We pose this joint training as bilevel optimization and propose an approximation algorithm to tackle its computation difficulties. This strategy allows flexible and efficient training rather than intensive labors of manually tuning in order to accommodate multiple types of medical data presenting significant variations.
	\item Comprehensive evaluations on challenging multiple registration tasks of image-to-atlas, uni-modal and cross-modal image-to-image demonstrate that our approach achieves state-of-the-art performance. We also investigate the effectiveness of our approach to support the down-streaming medical image analysis including fusion and segmentation.
\end{enumerate}

The paper is organized as follows. Section 2 describes related work.  Section 3 introduces our optimization-inspired propagation framework and Section 4 describes our training strategies. We demonstrate experimental results in Section 5 and conclude the paper in Section 6.

\section{Related Work}\label{sec:relatedwork}
In this section, we describe registration methods based on different mechanisms, e.g, simple low-dimensional parametric models, model-based registration, prior-based registration
deep learning-based methods and related optical flow task.

\textbf{Low-dimensional parametric registration. }~Simple, low-dimensional parametric models, e.g., rigid~\cite{HZ2004}, affine~\cite{BuergerSK11}, or homography transformations~\cite{ZhangWLJYWZS20}, try to find a matrix that achieves the best possible agreement between a transformed source and a target image, which consists of 6 or 8 degrees of freedom.
These parametric models usually serve as an initial alignment followed by more advanced, high-dimensional parametric or non-parametric registration models, such as deformable transformations that are with more degrees of freedom and the ability to capture subtle, localized images deformations.
In our paper, we concentrate on the latter step, in which we compute a dense, nonlinear correspondence for all voxels and things become more complex and challenging.

\textbf{Model-based registration.}~
Physical models can be typically separated into elastic body models~\cite{PennecSAFA05}, viscous fluid flow models~\cite{ChiangLKDBRMZTT08}, diffusion models~\cite{MansiPSDA11}, curvature registration~\cite{BeuthienKF10}, and flows of diffeomorphisms~\cite{Avants08}. In these cases, the transformation is governed by different types of Partial Differential Equation (PDE). 
These physical principle inspired methods generally ensure desirable properties such as inverse consistency and topology preservation.
Among them, diffeomorphic frameworks~\cite{Ashburner07,BegMTY05,Avants08,YangLRJ15,PaiSSDSN16} use smooth velocity fields to represent the deformation, have shown remarkable success in various computational anatomy applications.
However, the methods estimate transformation through the solution of the PDE that can be computationally demanding and too complicated to work with.

\textbf{Prior-based registration.}~
It is possible to introduce prior knowledge about deformations when registration involves image acquisitions of specific anatomical organs, such as the tumor, prostate, breast, brain, lung and cardiac.
Prior-based approaches that exploit our knowledge regarding the problem through the use of more informed priors at the cost of being constrained to well-defined settings, including statistically-constrained methods and biophysical model inspired methods.
Statistically-constrained registration frameworks~\cite{TangFWKS09,KimWYS12,BrunLPCLZMWGT11} can capture statistical information about deformation fields across a population of subjects but are generally limited by previously-observed deformations.
The biomechanical inspired methods~\cite{KonukogluCMSWMDA10,PhatakMVPBW09} highly rely on complicated priors. 
With the complexity of different clinical scenes, it is difficult to introduce priors flexibly that ideally regularize the transformation. 

\begin{figure*}[t] 
	\centering 
	\includegraphics[width=1.0\textwidth]{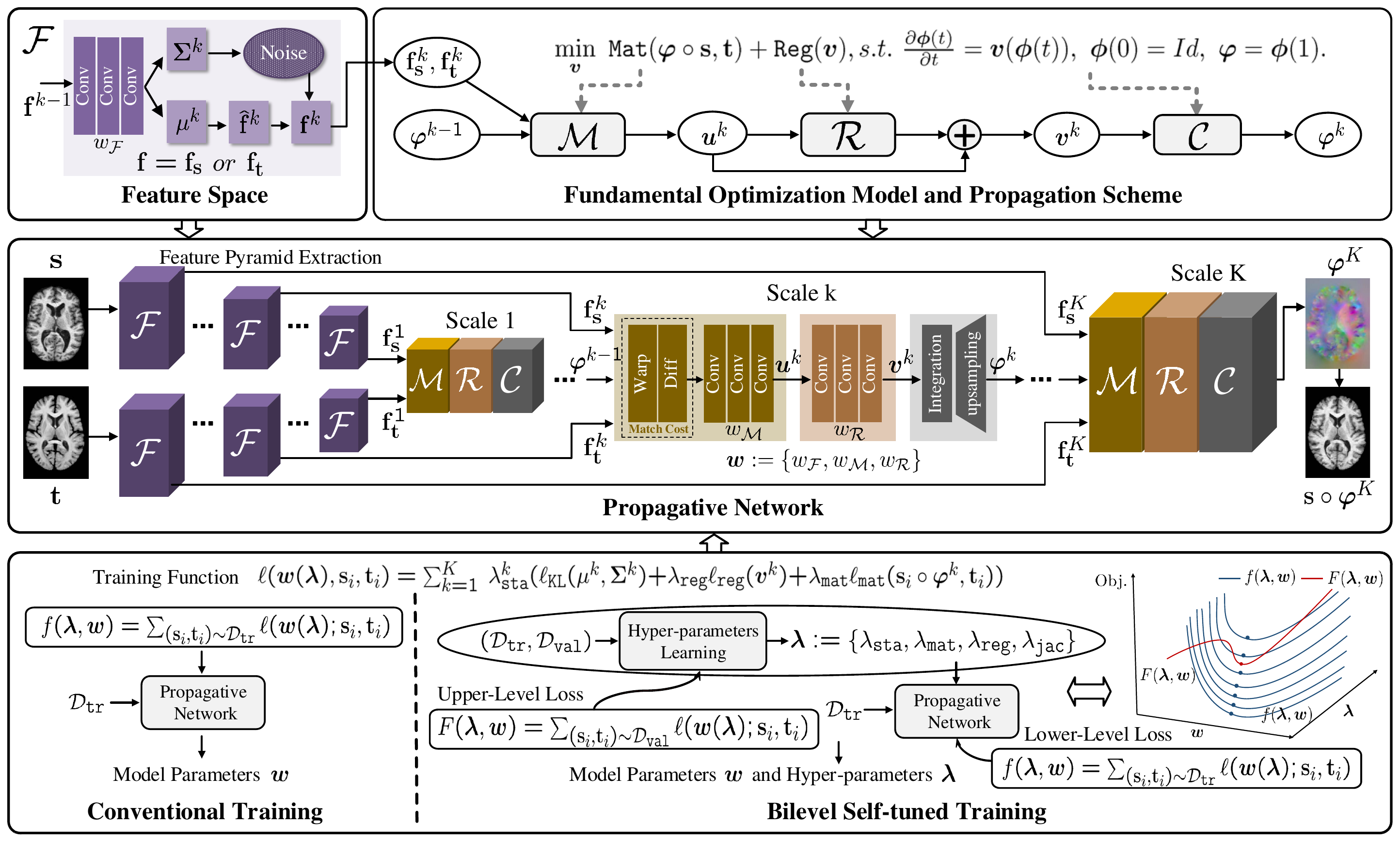} 
	\vspace{-1.2em}
	\caption{ 
		The first row shows our optimization learning perspective, which is to cascade three modules to propagate the optimization of registration fields on feature space.
		In the second row, the feature extraction $\mathcal{F}$, error-based matching $\mathcal{M}$, regularization $\mathcal{R}$, constraint $\mathcal{C}$ modules form one iteration of our framework.
		Different from conventional training, thanks to upper-level and lower-level objectives, our bilevel training in the third row could learn model parameters $\y$ and hyper-parameters $\x$ collaboratively. We also give its illustration. Blue cures represent the lower-level objective and their minimal values are shown as blue dots. The red cure denotes the upper-level objective, whose minimal value is the red dot. 
	}  
	\label{fig:pipline}
\end{figure*}

\textbf{Learning-based registration.}~
Deep learning based methods~\cite{BalakrishnanZSG19,DalcaBGS19,YangKSN17,IlgMSKDB17,liu2019theoretically,9098526} taking advantage of neuron networks have shown impressive results, especially in terms of runtime. 
Inspired by the work of spatial transformer~\cite{JaderbergSZK15}, plenty of works~\cite{BalakrishnanZSG19,DalcaBGS19,ShenHXN19,ZhaoDCX19,HeringGH19} have focused on replacing costly numerical optimization with global function optimization over the training data in an unsupervised way.
Recently, some researchers~\cite{DalcaBGS19,ShenHXN19,MokC20,LiuLZFL20,WangZ20} propose to estimate the velocity fields or momentum fields, which can be used to obtain diffeomorphic transformations.
Typically, hyper-parameters such as regularization parameters in loss function exist in registration networks, which are crucial to the registration performance and generalization capability.
However, learning-based approaches pose hyper-parameter learning as a black-box problem and require considerable computational and human effort to perform many training runs, especially when switching to other registration tasks.

\textbf{Related tasks.}~
Optical flow estimation is a related registration problem for 2D images which returns a dense displacement vector field depicting small displacements between image pairs.
The interest of optical flow estimation is to recover the apparent motion of objects between sequences of successive images, where its spatial correspondences/displacement field are associated with different time points.
Learning-based optical flow approaches take a pair of images as input and use a convolutional neural network to learn the optical flow from data.
Most of these works~\cite{IlgMSKDB17,SunY0K18,hui18liteflownet,Hur019} require supervision in the form of ground truth flow fields while using an unsupervised objective~\cite{MeisterH018,abs-2012-00212} has emerged as a trend in recent.
PWC-Net~\cite{SunY0K18} and IRR~\cite{Hur019} apply iterative refinement using coarse-to-fine pyramids. 
Other related  applications include from tracking to depth prediction, stereo reconstruction and so on.
However, on memory-costly 3D image registration tasks,  things become more complex compared to those of 2D tasks. That is, designing effective architectures that perform better, train more easily and generalize well to novel scenes is more difficult and challenging, which is exactly the key contributions of this work.

\section{Learning Registration from Optimization}\label{sec:method}

To capture large deformations, diffeomorphic registrations have been frequently employed. These diffeomorphic methods ~\cite{BegMTY05,NiethammerKV19,YangKSN17,LiuLZFL20} have many desirable mathematical properties, such as invertibility, one-to-one smoothness, and topology-preserving. 
However, these physical model inspired methods~\cite{Ashburner07,BegMTY05,Avants08,YangLRJ15,PaiSSDSN16} generally solve the optimization problem on the image domain, while the high dimensionality of the registration field parameters as well as the non-linear relationship between the images and the parameters pose a significant computational challenge.
Prior-based methods~\cite{KonukogluCMSWMDA10,KonukogluCMSWMDA10} introduce prior knowledge regarding the physical properties of the underlying anatomical structure. The informed priors may help to render the registration method more robust and stable, but it is challenging to introduce priors flexibly under different clinical scenes. These limitations make purely optimization-based registration methods hard to obtain the solutions efficiently and flexibly.
The goal of this work is to learn a powerful solver to aggregate a variety of mechanisms to address the deformable registration problem efficiently.  
We first introduce a general optimization model to formulate deformable image registration, inspired by which we present our optimization propagation framework with a series of modules on multi-scale feature space.

\textbf{Fundamental optimization formulation of registration.} 
Given a source image $\mathbf{s}$ and a target image $\mathbf{t}$ with a spatial domain ${\Omega \in \mathbb{R}^{d}}$, specifically, we aim at minimizing the following constrained optimization model:
\begin{equation}
\begin{array}{l}
\quad\min\limits_{\bm{v}} \ {\mathtt{Mat}}(\bm{\varphi} \circ \mathbf{s},\mathbf{t})+ {\mathtt{Reg}}(\bm{v}),\\s.t. \ \frac{\partial \bm{\phi}(t)}{\partial t} = \bm{v}(\bm{\phi}(t)), \ \bm{\phi}(0) = Id, \ \bm{\varphi}=\bm{\phi}(1),
\end{array}
\label{eq:1}
\end{equation}
where the $\circ$ represents warping operation, $\bm{\varphi} : \mathbb{R}^{d} \to \mathbb{R}^{d}$ is the final deformation field,  $\bm{v}$ is the stationary velocity fields for unit time. ${\mathtt{Mat(\cdot,\cdot)}}$ is data matching term, forcing the similarity of image pairs. 
${\mathtt{Reg(\cdot)}}$ imposes regularization on the deformation and guarantees its smoothness, by constraining on the velocity fields $\bm{v}$. 
Govern by the ordinary differential equation constraint, $\bm{\phi}(0) = Id$ is the identity transformation, $t \in [0,1]$ represents the time, such that 
generating final registration field involves starting with an identity transform $\bm{\phi}(0) = Id$ and integrating of a stationary velocity field over unit time to obtain $ \bm{\varphi}=\bm{\phi}(1)$. The time $t$ is the time step for the integration process, and the stationary velocity field at each time step can be conceptualized as an integration with a single time step.

We try to bridge the correspondences between well-established principles in conventional methods and registration networks. 
As the first line of Fig.~\ref{fig:pipline} shows, we unroll the optimization process on the discriminative multi-scale feature space, then design the error-based data matching, context-based regularization, and constraint modules, corresponding to data matching, regularization, and constraint in Eq.~\eqref{eq:1}.
Note that although our framework uses insights from conventional registration methods, the proposed modules do not \emph{exactly} solve the corresponding optimization/energy.
Next, we will elaborate on our different modules.

\subsection{Feature Pyramid Extraction}
\label{sec:RFPE}
To learn features that are invariant to noise and uninformative intensity-variations, we propose a generative feature module, which involves a latent variable model (as in VAE~\cite{KingmaW13}). 
We assume a probabilistic distribution for feature representations $p(\mathbf{f}|\mathbf{I})$, where $\mathbf{I} \!: \! \{\mathbf{s}, \!\mathbf{t}\}$, $ \mathbf{f} \!: \! \{\mathbf{f}_{\mathbf{s}}, \mathbf{f}_{\mathbf{t}}\}$, the later means feature representations of the source and target image. 
The prior probability of the feature $\mathbf{f}$  can be modeled as the multivariate normal distribution with spherical covariance $I$:
\begin{equation}
p(\mathbf{f})  = \mathcal{N}( 0 , I).
\end{equation}
We formulate the process of feature preparation as:
\begin{equation}
\begin{array}{l}
\mathbf{f}^{k+1}_{\mathbf{s}}, \mathbf{f}^{k+1}_{\mathbf{t}} \!=\!  \left\{
\begin{array}{ll}
\mathcal{F}(\mathbf{s}, \mathbf{t};\  w^{k+1}_{\mathcal{F}}),    & \!\text{if}~k=0,\\
\mathcal{F}(\mathbf{f}^{k}_{\mathbf{s}}, \mathbf{f}^{k}_{\mathbf{t}};\  w^{k+1}_{\mathcal{F}})    & \!\text{otherwise}.\\
\end{array} \right.
\end{array}
\label{eq:2}
\end{equation}
where $w_{\mathcal{F}}^{k+1}$ denotes trainable parameters of the feature extraction network at the $(k+1)$-th scale. The calculation of the mapping $\mathcal{F}$ will be discussed in Section~\ref{sec:loss}.

\subsection{Deep Propagative Modules}
\label{sec:DPM}
Based on learned feature pyramids, we further unroll the optimization process and design a series of deep modules to propagate the optimization of registration fields.
The second line of Fig.~\ref{fig:pipline} demonstrates the cascade of all ingredients at each iteration of our propagative network.

\textbf{Error-based data matching module.}~
At each iteration, to establish accurate voxel-to-voxel correspondence and reduce the feature space distance between image pairs, we propose to generate the error-based reconstructed registration field $\bm{u}^{k+1}$as:
\begin{equation}
\begin{array}{l}
\bm{u}^{k+1} = \mathcal{M}(\bm{\varphi}^{k}, \mathbf{f}^{k+1}_{\mathbf{s}}, \mathbf{f}^{k+1}_{\mathbf{t}}, \mathbf{e}^{k+1});\  w^{k+1}_{\mathcal{M}}),
\end{array}
\label{eq:3}
\end{equation}
where $w_{\mathcal{M}}^{k+1}$ denotes the parameters of the matching network, $\mathbf{e}$ stores the matching error of the corresponding voxels of the two feature presentations. Inspired by image warping, we directly perform feature warping using the spatial transform function~\cite{BalakrishnanZSG19}~\cite{JaderbergSZK15}, then construct this error/misalignment as:
\begin{equation}
\begin{array}{l}
\mathbf{e}^{k+1}  = \| \mathbf{f}_{\mathtt{t}}^{k+1} - \mathbf{f}_{\mathtt{s}}^{k+1} \circ \bm{\varphi}^k \|_1.
\end{array}
\end{equation}

\textbf{Regularization module. }~
Using contextual information to regularize the flow/registration field has been widely used in traditional flow methods.  At each iteration, we thus apply a context-based regularization network to produce refined registration field $\bm{v}^{k+1}$ as:
\begin{equation}
\begin{array}{l}
\bm{v}^{k+1} = \mathcal{R}(\bm{u}^{k+1};\  w^{k+1}_{\mathcal{R}}),
\end{array}
\label{eq:4}
\end{equation}
where $w_R^{k+1}$ are the learnable parameters of the regularization network. 
To employ the contextual information, this sub-network applies dilated convolution, which effectively enlarges the receptive field size.

\textbf{Constraint module.}~
At each iteration, to provide the diffeomorphism guarantee, we define the deformation field through the ordinary differential equation constraint. Thus, we append a numerical integration module to generate the deformation field $\bm{\varphi}^{k+1}$ as:
\begin{equation}
\begin{array}{l}
\bm{\varphi}^{k+1} = \mathcal{C}(\bm{v}^{k+1};\  w_{\mathcal{C}}),
\end{array}
\label{eq:5}
\end{equation}
where $ w_{\mathcal{C}}$ are the parameters in this module.
We compute this integration using scaling and squaring~\cite{Ashburner07}~\cite{BalakrishnanZSG19} method.
Specifically, it recursively computes the solution in successive small time-steps $h$ as:
$ \bm{\phi}(t+h) = \bm{\phi}(t) + h\bm{v}(\bm{\phi}(t)) = (x+h\bm{v}) \circ \bm{\phi}(t) $.
In our experiments, we use seven steps.


\section{Bilevel Self-tuned Training }

Conventional training involves choosing and tuning hyper-parameters that significantly affect model performance, especially when switching to other data or applications. 
General training generally uses a grid-search algorithm or manually tuning to obtain task-specific hyper-parameters, requiring many training runs with various hyper-parameter configurations, potentially leading to sub-optimal results and requiring considerably computational and human effort.
In the following, we first introduce our training objective, which regards model parameters and hyper-parameters.
Then, to tackle the inefficiency of hyper-parameter tuning, we propose our bilevel self-tuned training and give the solution strategy.

\subsection{Training Objective} \label{sec:loss}
The objectives regarding model parameters $\y$ and hyper-parameters $\x$ consist of three components: KL loss, matching loss, and regularization loss.
Let $\y$ be the set of learnable parameters in the proposed framework, which includes the feature extraction network $w_{\mathcal{F}}$, error-based data matching network $w_{\mathcal{M}}$ and regularization network $w_{\mathcal{R}}$ at different scales (the integration module has no learnable parameters). And, the $\x$ repersent the hyper-parameters to trade off the different loss terms, including $\lambda_{\mathtt{sta}}, \lambda_{\mathtt{mat}}, \lambda_{\mathtt{reg}}, \lambda_{\mathtt{jac}}$.
Specifically, we compute the sum of the losses at different scales as: 
\begin{align}
&\ell(\y(\x), \s_i,\t_i) \\ \nonumber
&\!=  \! \sum_{k=0}^{K-1} \! \ \lambda_{\mathtt{sta}}^{k} ( \ell_{\mathtt{KL}}(\mathbf{\mu}^{k},\! \mathbf{\Sigma}^{k}\!) +  \! \lambda_{\mathtt{mat}} \ell_{\mathtt{mat}}(\mathbf{s}_i \! \circ \! \bm{\varphi}^{k}, \mathbf{t}_i) + \! \lambda_{\mathtt{reg}} \ell_{\mathtt{reg}}(\bm{v}^{k})  ) ,
\end{align}
where  $\mathbf{\mu}, \mathbf{\Sigma},\bm{\varphi}, \bm{v}$ depends on  $w_{\mathcal{F}}, w_{\mathcal{M}}, w_{\mathcal{R}}$ (cf Section~\ref{sec:RFPE} and Section~\ref{sec:DPM} ).
Then, we elaborate on detailed computations. 

\textbf{KL loss.}~
Computing the posterior probability $p(\mathbf{f}|\mathbf{I})$ of feature extraction module is intractable so that we utilize an approximation of the intractable true posterior probability,  $q_{\theta}(\mathbf{f} | \mathbf{I})$ parametrized by $\theta$~\cite{KingmaW13}. We minimize the KL divergence:
\begin{align}
& \min\limits_{\theta} \  \mathrm{KL}[q_{\theta}(\mathbf{f} | \mathbf{I}) \| p(\mathbf{f}|\mathbf{I} )],  \nonumber\\
= &  \ \min\limits_{\theta}  \ \mathrm{KL}[q_{\theta}(\mathbf{f} | \mathbf{I}) || p(\mathbf{f})] - \mathbb{E}_{\mathbf{f} \sim q }[\ln p(\mathbf{I} | \mathbf{f} )] ,
\label{eq:kl}
\end{align}
where the first term acts as a regularizer, while the second term is an expected negative reconstruction error. In this work, the reconstruction error corresponds to the registration loss, including matching loss and regularization loss. 
Then, we model the approximate posterior $q_{\theta}(\mathbf{f} | \mathbf{I})$ as a multivariate normal: 
\begin{equation}
q_{\theta}(\mathbf{f} | \mathbf{I}) = \mathcal{N}( \mathbf{f};  \mathbf{\mu}(\mathbf{f} |  \mathbf{I}), \mathbf{\Sigma}(\mathbf{f} |  \mathbf{I})),
\end{equation}
and apply feature extraction network to predict approximate posterior probability parameters, mean $ \mathbf{\mu}(\mathbf{f} |  \mathbf{I})$ and covariance $\mathbf{\Sigma}(\mathbf{f} |  \mathbf{I})$, from which we then sample the feature to generate the deformation fields. The KL-term can be computed in closed form:
\begin{align}
\ell_{\mathtt{KL}}(\mathbf{\mu}, \mathbf{\Sigma}) = 1/2(\mathrm{tr}(\mathbf{\Sigma}) + ||\mathbf{\mu} || - \log \det(\mathbf{\Sigma}) - m) , \! 
\end{align}
where $\mathbf{\Sigma}_{\mathbf{f} |  \mathbf{I}}$ need to be diagonal, and $m$ is a const.

\textbf{Matching loss.}~
We inherit similarity metric from energy-based approaches~\cite{BegMTY05,Avants08,SunNK14} to define the data matching loss.
Specifically, we use local normalized cross correlation coefficient to penalize differences in appearance.
Note that, when computing matching loss, we scale the image pairs and warp the downsampled images with the deformation field at each scale.
At different scales, we use different window sizes to compute the local normalized correlation coefficient. From the zeroth to the third scale, the window sizes are set to $3, 5, 7, 9$.

\textbf{Regularization loss.}~
We employ the diffusion regularizer on spatial gradients of the velocity fields and apply the Jacobian determinant loss to further constrain the smoothness of the deformation fields as:
\begin{equation}
\ell_{\mathtt{reg}}(\bm{v}) = \sum_{x \in \Omega}  \  \| \bigtriangledown \bm{v} (x) \|^{2}_{2} +  \lambda_{\mathtt{jac}} \max (0, -| J_{\bm{v}}(x)|),
\label{eq:11}
\end{equation}
where the Jacobian matrix $J_{\phi}(x) = \nabla \phi(x) $ captures the local properties of $\phi$ around voxel $x$. We penalize these negative determinants to enforce topology-preservation~\cite{Ashburner07}.

\subsection{Self-tuned Bilevel Formulation}
To tackle the inefficiency of hyper-parameter tuning, we propose our new bilevel self-tuned training, allowing the efficient search of the task-specific hyper-parameter, as shown in the right-bottom part of Fig.~\ref{fig:pipline}.
We start by denoting $\D_{\mathtt{tr}}$ and $\D_{\mathtt{val}}$ as the training and validation sets, respectively.
In our setting, we formulate the learning of the model parameter as $f$:
\begin{equation}
f(\x,\y)=\sum_{(\s_i,\t_i)\sim \D_{\mathtt{tr}}}\ell(\y(\x);\s_i,\t_i),
\label{eq:ll}
\end{equation}
where the $\ell(\y(\x);\s_i,\t_i)$ denotes the loss function with the model parameter $\y$ coupled with  hyper-parameters $\x$, the $(\s_i,\t_i)$ corresponding to source and target image. 

To identify optimal task-specific hyper-parameters $\x$, we minimize the empirical loss on a validation set, which represents a proxy for the generalization error of $\y$. We define the objective of the hyper-parameter learning problem as $F$:
\begin{equation}
F(\x,\y)=\sum_{(\s_i,\t_i)\sim \D_{\mathtt{val}}}\ell(\bm{w}(\bm{\lambda});\s_i,\t_i).
\end{equation}
The $F$ does not depend explicitly on the hyper-parameter $\x$, since in our setting $\x$ is instrumental in finding a good model $\y$, which is our final goal.

We then consider bilevel optimization problems~\cite{FranceschiDFP17,MacKayVLDG19,LiuMYZZ20,LiuCHFLL20,9293146,abs-2012-05609,abs-2101-11517} of the form to formulate our self-tuning learning. Specifically, we introduce a hierarchical optimization model governed by the constraint $\mathcal{C}(\x)$ as:
\begin{equation}
\min\limits_{\x}F(\x,\y), \ s.t. \ \y \in \mathcal{C}(\x), 
\label{eq:bo}
\end{equation}
where we take optimization of task-specific hyper-parameters $\x$ as the upper-level subproblem,
constraint $\mathcal{C}(\x)$ as the solution set of the lower-level subproblem, while lower-level subproblem denotes optimization of model parameters $\y$:
\begin{equation}
\mathcal{C}(\x) := \{ \arg\min\limits_{\y}f(\x,\y) \},
\label{eq:LP}
\end{equation}
where we try to find optimal parameters $\y$ by minimizing the objective $f$ on training data $\D_{\mathtt{tr}}$.
We follow methods~\cite{FranceschiDFP17,MacKayVLDG19,9293146,abs-2101-11517} to rely on the uniqueness of $C(\x)$, where $ \mathcal{C}(\x) = \arg\min f(\x, \y)$ and the lower-level subproblem only has one single optimal solution $\y^*$ for a given $\x$.

\begin{algorithm}[t]
		\caption{Bilevel Self-Tuned Training Algorithm}\label{alg:search}
		\begin{algorithmic}[1]
			\REQUIRE 
			The training and validation datasets $\mathcal{D}_{\mathtt{tr}}$ and $\mathcal{D}_{\mathtt{val}}$ and initialization parameters.
			\ENSURE The optimal hyper-parameters and model parameters. 
			\WHILE {not converged}
			\STATE Calculate the practical Jacobian $\frac{\partial F(\x,\y^*)}{\partial \x}$ in Eq.~\eqref{eq:Ja} via one-step first-order gradient approximation.
			\STATE Perform gradient descent to update $\x$ based on $\frac{\partial F(\x,\y^*)}{\partial \x}$.
			\STATE Calculate $\y$  based on Eq.~\eqref{eq:LP}.
			\ENDWHILE
			\RETURN  The optimal $(\x,\y(\x))$. 
		\end{algorithmic}
\end{algorithm}

\subsection{First-order Solution Strategy}

Due to the nested structure of the bilevel problem in Eq.~\eqref{eq:bo},
evaluating exact gradients of $\x$ for the upper-level problem is difficult and computationally challenging.
Moreover, this hierarchical structure between two levels greatly challenge traditional optimization techniques especially when the dimension of either level is huge. It is the case of our training where the lower level involves hundreds of thousands of deep network parameters. Moreover, computing gradients w.r.t. 3D medical volumes demands additional efforts. Existing solutions~\cite{FranceschiDFP17,MacKayVLDG19,LiuMYZZ20,9293146} are inapplicable to our self-tuned training. 
Next, we introduce an efficient solution to compute the gradient of $\x$ via one-step first-order gradient approximation.

\textbf{Derivation of computing the gradient of $\x$. }~
Our goal is to calculate the derivatives of $F(\x,\y^*)$ with respect to $\x$ as~\footnote{Please notice that we actually do not distinguish between the operation of the derivatives and partial derivatives to simplify our presentation.}:
\begin{equation}
\frac{\partial F(\x,\y^*)}{\partial \x}=\frac{\partial F(\x,\y^*)}{\partial\x}
+   \left( \frac{\partial \y^*}{\partial \x^\prime} \right)^\prime  \frac{\partial F(\x,\y^*)}{\partial\y^*},
\label{eq:Ja}
\end{equation}
where we denote the transpose operation as ``$\prime$''.

To solve the aforementioned problem, firstly, we approximate the solutions of lower-level optimization $\y^*$ in Eq.~\eqref{eq:LP} by the $T$-step iterate of a dynamical system. Given an initialization $\y_0=\E_{0}(\x)$ at $t=0$, the iteration process can be written as:
\begin{equation}
\begin{array}{l}
\y_{t}=  \E_{t}(\y_{t-1}; \x), \ t=1,\cdots,T
\end{array}
\end{equation}
where $\E_{t}$ denotes the operation performed at the $t$-th step and $T$ is the number of iterations. 
For example, we formulate $\E_{t}$ based on the gradient descent rule as:
\begin{equation}
\E_{t}(\y_{t-1}; \x) =  \y_{t-1} - s_t \frac{\partial f(\x,\y_{t-1})}{ \partial\y_{t-1}},
\label{eq:lower-level_ds_discrete}
\end{equation}
where $s_t$ denotes the learning rate.
To reduce the computational burden further, we set $T=1$. 
Similar to the work of~\cite{LiuSY19}, we perform one-step iteration. 
So that we propose a simple approximation scheme as:
\begin{equation}
\frac{\partial F(\x,\y^*)}{\partial \x} = \frac{\partial F(\x,\y_0 - s_1 \frac{\partial f(\x,\y_{0})}{ \partial\y_{0}})}{\partial \x}.
\end{equation}

Now, by formulating the dynamical system as that in Eq.~\eqref{eq:lower-level_ds_discrete}, we then write $\frac{\partial\y^*}{\partial\x}$ as:
\begin{equation}
\frac{\partial\y^*}{\partial\x} = \frac{\partial \left(\y_0 - s_1 \frac{\partial f(\x,\y_0)}{ \partial\y_0} \right)}{\partial \x}  = -s_1 \frac{\partial^2f(\x,\y_0)}{\partial\x \partial\y} .
\label{eq:He}
\end{equation}
The expression above contains expensive matrix-vector product for Hessian calculation. We then approximate the Hessian calculation with the first order gradients. We introduce the following central difference to approximate it as:
\begin{equation}
\frac{\partial F(\x,\y^*)}{\partial\y^*}\frac{\partial^2f(\x,\y_0)}{\partial\x \partial\y_0}\approx
\frac{\frac{\partial f(\x,\y_0^{+})}{\partial\x}-\frac{\partial f(\x,\y_0^{-})}{\partial\x}}{2\epsilon} ,
\end{equation}
where $\y_0^{\pm}=\y_0\pm\epsilon\frac{\partial F(\x,\y^*)}{\partial\y^*}$, the $\epsilon$ is set to be a small scalar equal to the learning rate.
So that we may further reduce the computation complexity.

The overall iterative procedure is outlined in Alg.~\ref{alg:search}. Overall, we introduce an efficient and feasible solution strategy to solve bilevel optimization of hyper-parameters for registration networks.

\begin{table*}[t]
	\centering
	\renewcommand\tabcolsep{4.5pt} 
	\caption{Ablation analysis of the feature pyramid extraction module on five brain MRI datasets in terms of Dice score and NCC. Larger values indicate better performance. The Affine only model represents the results for affine alignment. }
	\vspace{-0.5em}
	\begin{tabular}{|m{1.0cm}<{\centering} |m{1.4cm}<{\centering} |m{1.8cm}<{\centering} |m{1.8cm}<{\centering} |m{1.8cm}<{\centering} |m{1.8cm}<{\centering} |m{1.8cm}<{\centering} |m{1.8cm}<{\centering} |m{1.8cm}<{\centering}|}
		\hline
		\multicolumn{2}{| c |}{\multirow{2}*{Model}}& \multirow{2}*{Affine only} & \multicolumn{2}{c| }{2-scale} &\multicolumn{2}{c| }{3-scale} &\multicolumn{2}{c| }{4-scale}\\ 
		\cline{4-9}
		\multicolumn{2}{ |c| }{~} &   & W/O FEN & W/ FEN & W/O FEN & W/ FEN & W/O FEN & W/ FEN  \\ 
		\hline
		\multirow{2}*{OASIS} & Dice score & 0.580 $\pm$ 0.028 & 0.724 $\pm$ 0.019 & 0.744 $\pm$ 0.016 & 0.764 $\pm$ 0.011 & \textbf{0.777 $\pm$ 0.006} & 0.770 $\pm$ 0.008 & 0.773 $\pm$ 0.007  \\  
		~ & NCC & 0.088 $\pm$ 0.004 & 0.217 $\pm$ 0.004 & 0.229 $\pm$ 0.003  & 0.233 $\pm$ 0.003  & \textbf{0.245 $\pm$ 0.002} & 0.236 $\pm$ 0.003 & 0.240 $\pm$ 0.003 \\ \hline
		\multirow{2}*{ABIDE} & Dice score & 0.624 $\pm$ 0.024 & 0.736 $\pm$ 0.016 & 0.740 $\pm$ 0.018 & 0.754 $\pm$ 0.016 & \textbf{0.764 $\pm$ 0.015} & 0.763 $\pm$ 0.014 & 0.761 $\pm$ 0.017  \\ 
		~ & NCC & 0.094 $\pm$ 0.005  & 0.214 $\pm$ 0.004 & 0.227 $\pm$ 0.004  & 0.229 $\pm$ 0.004  & \textbf{0.241 $\pm$ 0.004} & 0.231 $\pm$ 0.004 & 0.237 $\pm$ 0.004 \\  \hline
		\multirow{2}*{ADNI} & Dice score & 0.571 $\pm$ 0.049 & 0.702 $\pm$ 0.038 & 0.730 $\pm$ 0.033 & 0.752 $\pm$ 0.024 & \textbf{0.773 $\pm$ 0.017} & 0.763 $\pm$ 0.020 & 0.769 $\pm$ 0.018  \\ 
		~ & NCC & 0.086 $\pm$ 0.006 & 0.213 $\pm$ 0.007 & 0.227 $\pm$ 0.006  & 0.231 $\pm$ 0.005  & \textbf{0.244 $\pm$ 0.005} & 0.233 $\pm$ 0.005 & 0.239 $\pm$ 0.005 \\ \hline
		\multirow{2}*{PPMI} & Dice score & 0.610 $\pm$ 0.033 & 0.740 $\pm$ 0.023  & 0.758 $\pm$ 0.019 & 0.773 $\pm$ 0.013  & 0.785 $\pm$ 0.011  & 0.779 $\pm$ 0.011  & \textbf{0.789 $\pm$ 0.011}  \\ 
		~ & NCC & 0.088 $\pm$ 0.004 & 0.213 $\pm$ 0.005 & 0.225 $\pm$ 0.004  & 0.229 $\pm$ 0.004 & \textbf{0.240 $\pm$ 0.004}  & 0.230 $\pm$ 0.004 & 0.235 $\pm$ 0.004 \\ \hline
		\multirow{2}*{HCP} & Dice score & 0.666 $\pm$ 0.027 & 0.698 $\pm$ 0.027  & 0.759 $\pm$ 0.014 & 0.738 $\pm$ 0.018  & 0.776 $\pm$ 0.010  & 0.745 $\pm$ 0.021  & \textbf{0.777 $\pm$ 0.009}  \\ 
		~ & NCC & 0.098 $\pm$ 0.004 & 0.183 $\pm$ 0.011 & 0.220 $\pm$ 0.005  & 0.204 $\pm$ 0.010 & \textbf{0.240 $\pm$ 0.004}  & 0.204 $\pm$ 0.011 &  0.234 $\pm$ 0.004 \\ 
		\hline
	\end{tabular}
	\label{tab:ablation-feature}
\end{table*}

\section{Experiments}\label{sec:experiment}
In this section, we first introduce our experimental setup. 
Then we explore the impact of each component of our paradigm and the benefits of the proposed bilevel self-tuned training strategy. 
Next, to demonstrate the superiority of the proposed algorithm, we compare it with state-of-the-art deformable registration techniques on accuracy, robustness, diffeomorphism preservation as well as efficiency. 
We evaluate the registration algorithms mainly on 3D brain MRI scans. Evaluations are conducted in the way of one aligning all the source data to a common atlas, called image-to-atlas registration.
Then we extend our paradigm to address general registration between two arbitrary volumes on liver CT scans, called image-to-image registration.

We also explore the utility of our registration framework to support the down-streaming medical image analysis tasks.
Multi-modal image registration and fusion are two important research issues in medical image processing. 
To optimally fuse two medical images, one needs to first accurately align them by minimizing non-linear differences between them using registration techniques. 
In the following experiments,  we demonstrated how to apply our paradigm to solve the challenging multi-modal registration tasks and powerfully support the following medical multi-modal fusion.
Medical image segmentation is also crucial and highly relevant in medical image analysis. Manual segmentation of brain MR images requires expertise and is time-consuming, and thus efficient data augmentation methods should be explored. 
The proposed framework can be used to register labeled atlas images to produce more labels images for segmentation.
Moreover, the latent feature space in Section~\ref{sec:RFPE} helps to encode similar deformations close to each other and allows the generation of synthetic deformations for a single image, which is beneficial for the data augmentation.
We also demonstrated how to apply our paradigm to strongly support the medical image segmentation.

\subsection{Experimental Setup}
\textbf{Dataset and pre-processing.}~
We performed image-to-atlas registration on brain MR datasets, including 551 T1 weighted MR volumes from seven publicly available datasets: ADNI~\cite{MUELLER200555}, ABIDE~\cite{Martino15}, PPMI~\cite{MAREK2011629}, OASIS~\cite{jocn.2007.19.9.1498}, and HCP~\cite{EssenSBBYU13}.
These scans were splitted into 370, 40, and 141 for training, validation, and testing, respectively.
We used the publicly available atlas from~\cite{BalakrishnanZSG19} as the target.
Considering the large disparity among different datasets, all scans were preprocessed with motion correction, NU intensity correction, normalization, skull stripping, and affine registration.  We used FreeSurfer~\cite{Fischl12} software to perform skull stripping and used FSL~\cite{WoolrichJPCMBBJS09} software for affine registration. The images were cropped to $160 \times 192 \times 224$ with 1 mm isotropic resolution after cropping unnecessary areas. 
For evaluation, all test MRI scans were anatomically segmented with Freesurfer to extract 30 anatomical structures.
As for image-to-image registration on liver CT scans, we included four publicly available datasets: MSD~\cite{MSD}, BFH~\cite{ZhaoLLCX20},  SLIVER~\cite{HeimannGS09}, and LSPIG~\cite{ZhaoDCX19}. We used the MSD and BFH with 1025 scans in total, and splitter into 900 and 125 as training and validation data. We randomly selected 380 image pairs with segmentation ground truth in SLIVER~\cite{HeimannGS09} for the evaluation.  We also evaluated with 34 intrasubject image pairs with segmentation ground truth in LSPIG~\cite{ZhaoDCX19}. For liver CT scans, we carried out normalization preprocessing steps and resample to a size of $128 \times 128 \times 128$.
We embedded the affine network as an integrated part.
These experiments enable not only assessment of performance on multi-site datasets but also the evaluation of scans that were not observed by the deep networks during training.

In addition, we performed multi-modal registration on BraTS18 and ISeg19 datasets, which are obtained from the Brain Tumour Segmentation challenge 2018 and Infant Brain MRI Segmentation challenge 2019.
Overall, the available training set consists of 135 cases, and for each case, two image modalities were standardized into a 3D volume in size of $160 \times 160 \times 160$ with 1 mm isotropic resolution. Among them, 10 cases have segmentation ground truth. The set was splitted into 115, 10 and 10 for train, validation and test. We use segmentation accuracy as a proxy for evaluating image registration accuracy.
As most of the provided T1 and T2 weighted images were already aligned, we randomly chose one of the T2 scans as our atlas and modeled our task as trying to register T1 scans to this T2 atlas.

\begin{table}[t]
	\centering
	\caption{Ablation analysis of model configurations on five brain MRI datasets in terms of Dice score and NCC. }
	\vspace{-0.5em}
	\label{tab:ablation-configurations}
	\begin{tabular}{|m{1.0cm}<{\centering}| m{1.4cm}<{\centering}| m{0.9cm}<{\centering}| m{0.9cm}<{\centering}| m{0.9cm}<{\centering} |m{0.9cm}<{\centering}|}
		\hline
		\multicolumn{2}{|c| }{Methods}       &  RN-    & MN-  &  RN+     &  MN+  \\
		\hline
		\multirow{2}*{OASIS} & Dice score   & 0.771  & 0.751   &  0.774      & \textbf{0.781} \\
		~ & NCC   & 0.235  & 0.202    &  0.241      &  \textbf{0.250}\\  \hline
		\multirow{2}*{ABIDE} & Dice score   & 0.759  & 0.731   &  0.765      & \textbf{0.771} \\
		~ & NCC   & 0.230  & 0.197    &  0.236      &  \textbf{0.246}\\  \hline
		\multirow{2}*{ADNI} & Dice score   & 0.762  & 0.737   &  0.771      & \textbf{0.775} \\
		~ & NCC   & 0.236  & 0.197    &  0.239      &  \textbf{0.249}\\  \hline
		\multirow{2}*{PPMI} & Dice score   & 0.780  & 0.757   &  0.784      & \textbf{0.788} \\
		~ & NCC   & 0.229  & 0.196    &  0.236      &  \textbf{0.245}\\  \hline
		\multirow{2}*{HCP} & Dice score   & 0.763  & 0.741   &  0.771      & \textbf{0.777} \\
		~ & NCC   & 0.234  & 0.199    &  0.239      &  \textbf{0.249}\\
		\hline
	\end{tabular}
\end{table}

\begin{table*}[t]
	\centering
	\renewcommand\tabcolsep{4.5pt} 
	\caption{Ablation analysis of model components on five brain MRI datasets in terms of Dice score and NCC. RN represents the regularization network.  }
	\vspace{-0.5em}
	\label{tab:ablation-block}
	\begin{tabular}{|m{1.0cm}<{\centering} |m{1.4cm}<{\centering} |m{1.8cm}<{\centering} |m{1.8cm}<{\centering} |m{1.8cm}<{\centering} |m{1.8cm}<{\centering} |m{1.8cm}<{\centering} |m{1.8cm}<{\centering} |m{1.8cm}<{\centering}|}
		\hline
		\multicolumn{2}{| c |}{\multirow{2}*{Model}}& \multirow{2}*{Affine only} & \multicolumn{2}{c| }{2-scale} &\multicolumn{2}{c| }{3-scale} &\multicolumn{2}{c| }{4-scale}\\ 
		\cline{4-9}
		\multicolumn{2}{ |c| }{~} &   & W/O RN & W/ RN & W/O RN & W/ RN & W/O RN & W/ RN  \\ 
		\hline
		\multirow{2}*{OASIS} & Dice score & 0.580 $\pm$ 0.028 & 0.746 $\pm$ 0.015 & 0.754 $\pm$ 0.011 & 0.771 $\pm$ 0.009 & \textbf{0.777 $\pm$ 0.006} & 0.773 $\pm$ 0.007 & 0.774 $\pm$ 0.007  \\ 
		~ & NCC & 0.088 $\pm$ 0.004  & 0.227 $\pm$ 0.003 & 0.228 $\pm$ 0.003 & 0.239 $\pm$ 0.003 & \textbf{0.245 $\pm$ 0.002} & 0.236 $\pm$ 0.003 & 0.245 $\pm$ 0.003 \\  \hline
		
		\multirow{2}*{ABIDE} & Dice score & 0.624 $\pm$ 0.024 & 0.746 $\pm$ 0.016 & 0.745 $\pm$ 0.016 & 0.759 $\pm$ 0.014 & 0.764 $\pm$ 0.015 & 0.762 $\pm$ 0.014 & \textbf{0.768 $\pm$ 0.015}  \\ 
		~ & NCC  & 0.094 $\pm$ 0.005  & 0.223 $\pm$ 0.003 & 0.225 $\pm$ 0.004 & 0.235 $\pm$ 0.004 & 0.241 $\pm$ 0.004 & 0.232 $\pm$ 0.004 & \textbf{0.242 $\pm$ 0.004} \\  \hline
		
		\multirow{2}*{ADNI} & Dice score  & 0.571 $\pm$ 0.049 & 0.731 $\pm$ 0.033 & 0.739 $\pm$ 0.029 & 0.765 $\pm$ 0.020 & \textbf{0.773 $\pm$ 0.017} & 0.767 $\pm$ 0.019 & 0.768 $\pm$ 0.020  \\ 
		~ & NCC & 0.086 $\pm$ 0.006  & 0.224 $\pm$ 0.006 & 0.225 $\pm$ 0.006 & 0.237 $\pm$ 0.005 & \textbf{0.244 $\pm$ 0.005} & 0.234 $\pm$ 0.005 & 0.244 $\pm$ 0.005 \\  \hline
		
		\multirow{2}*{PPMI} & Dice score & 0.610 $\pm$ 0.033 & 0.758 $\pm$ 0.018 & 0.762 $\pm$ 0.016 & 0.779 $\pm$ 0.013 & \textbf{0.785 $\pm$ 0.011} & 0.780 $\pm$ 0.012 & 0.782 $\pm$ 0.012 \\  
		~ & NCC  & 0.088 $\pm$ 0.004  & 0.222 $\pm$ 0.004 & 0.224 $\pm$ 0.004 & 0.234 $\pm$ 0.004 & 0.240 $\pm$ 0.004 & 0.231 $\pm$ 0.004 & \textbf{0.241 $\pm$ 0.004} \\  \hline
		
		\multirow{2}*{HCP} & Dice score & 0.666 $\pm$ 0.027 & 0.738 $\pm$ 0.028 & 0.745 $\pm$ 0.024 & 0.767 $\pm$ 0.017 & \textbf{0.776 $\pm$0.010} & 0.769 $\pm$ 0.015 & 0.770 $\pm$ 0.016  \\ 
		~ & NCC & 0.098 $\pm$ 0.004  & 0.225 $\pm$ 0.005 & 0.226 $\pm$ 0.005 & 0.238 $\pm$ 0.004 & 0.240 $\pm$ 0.004 & 0.235 $\pm$ 0.005 & \textbf{0.244 $\pm$ 0.004} \\ 
		\hline
	\end{tabular}
\end{table*}

\begin{figure*}
	\centering
	\begin{tabular}{c@{\extracolsep{0.35em}}c@{\extracolsep{0.35em}}c@{\extracolsep{0.35em}}c}
		\includegraphics[width=0.133\textwidth]{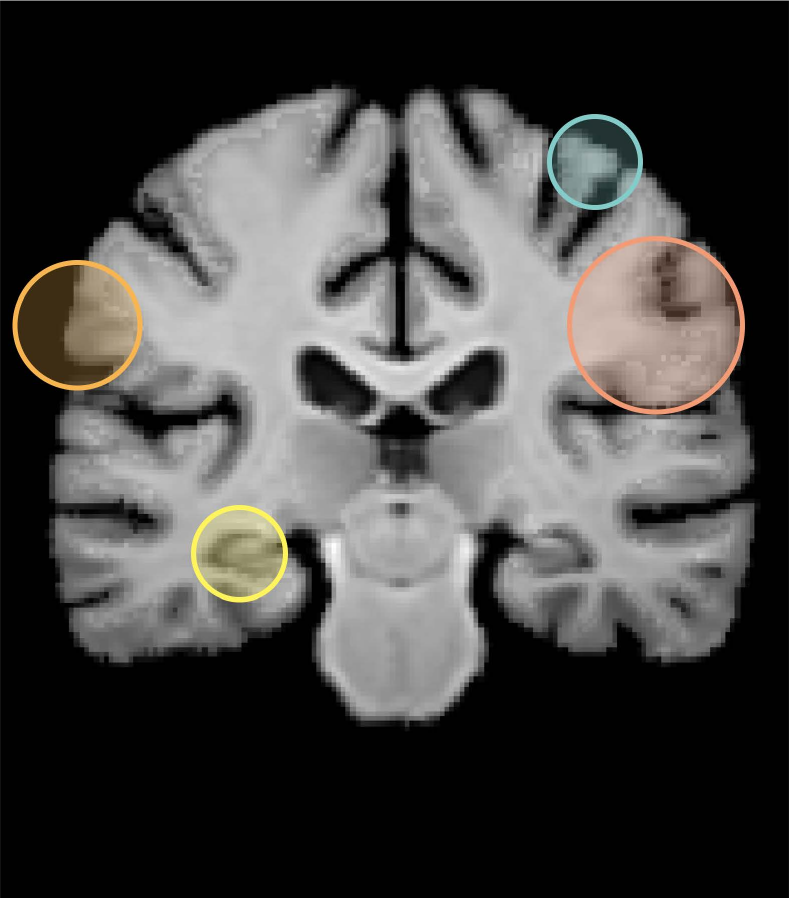}
		&\includegraphics[width=0.266\textwidth]{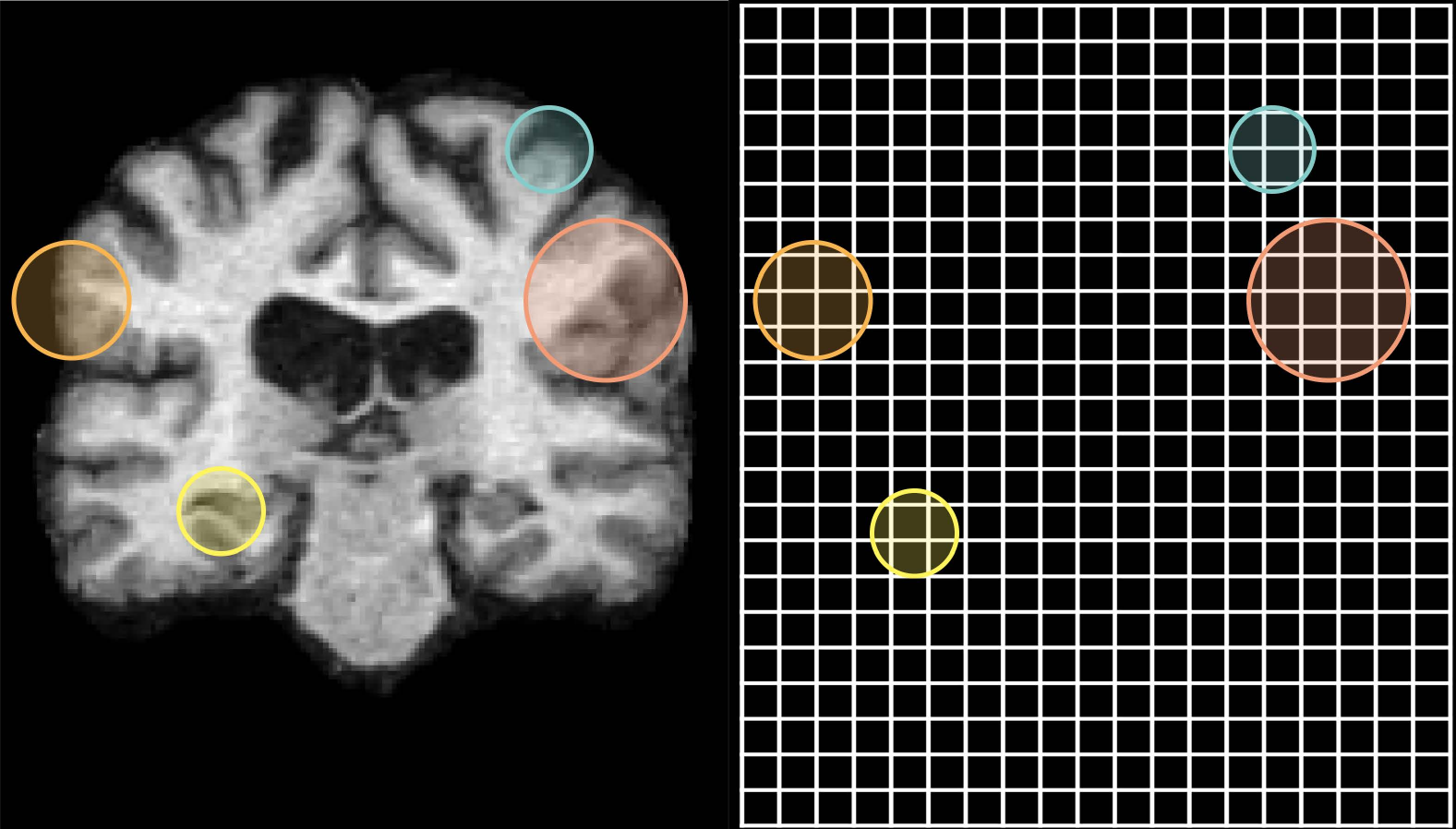}
		&\includegraphics[width=0.266\textwidth]{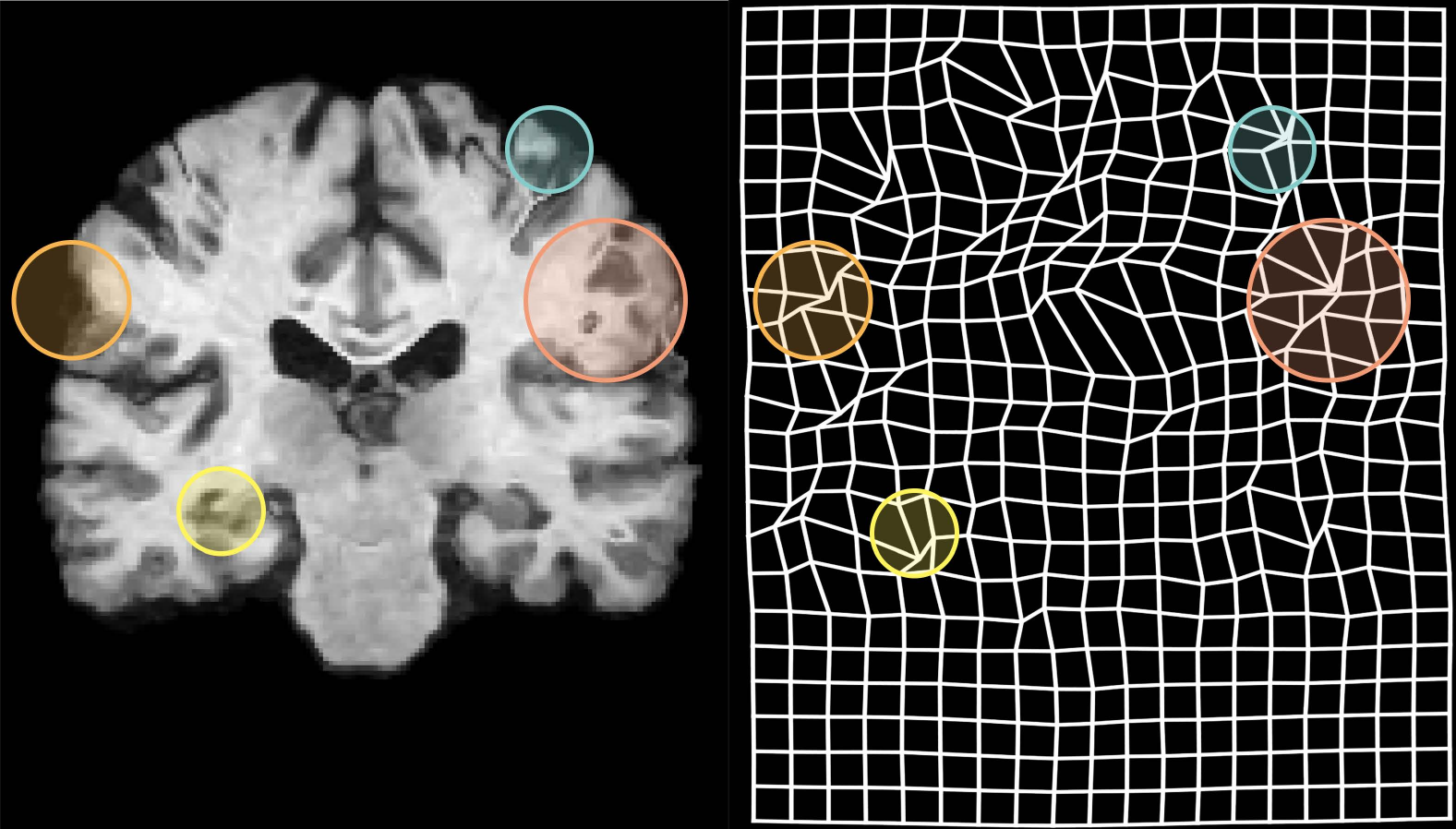}			
		&\includegraphics[width=0.266\textwidth]{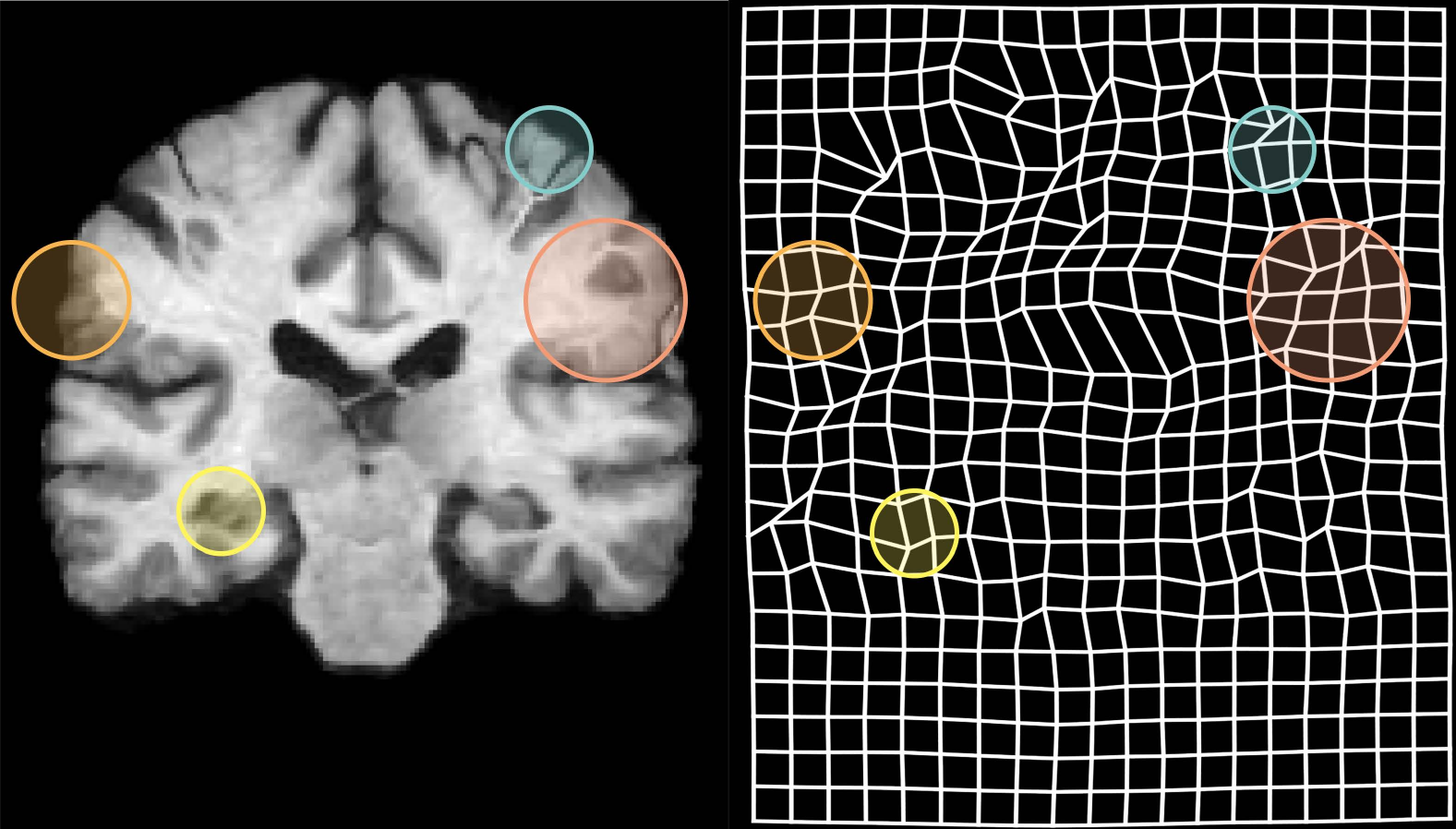}\\
		
		Target & Source& 	W/O Constraint &  W/ Constraint\\
	\end{tabular}
	\caption{Comparisons on deformation fields by warping the source image to target without and with the constraint module. Singularities emerge in the circled fields when applying no constraint.}
	\label{fig:integration} 
\end{figure*}

\begin{figure*}[!htp]
	\centering
	\begin{tabular}{c@{\extracolsep{0.2em}}c@{\extracolsep{0.2em}}c@{\extracolsep{0.2em}}c@{\extracolsep{0.2em}}c@{\extracolsep{0.2em}}c@{\extracolsep{0.2em}}c@{\extracolsep{0.2em}}c}		
		\includegraphics[width=0.1152\textwidth]{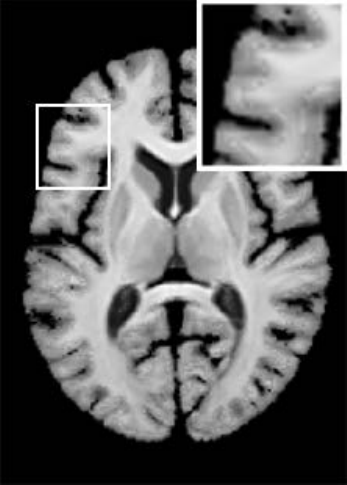}
		&\includegraphics[width=0.1152\textwidth]{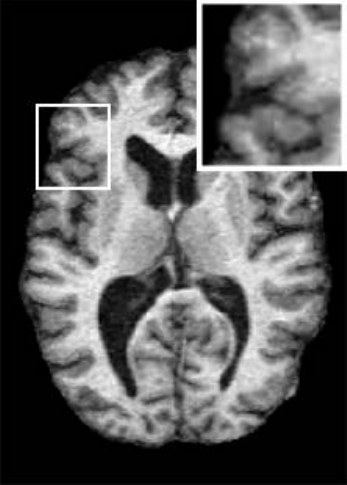}
		&\includegraphics[width=0.1152\textwidth]{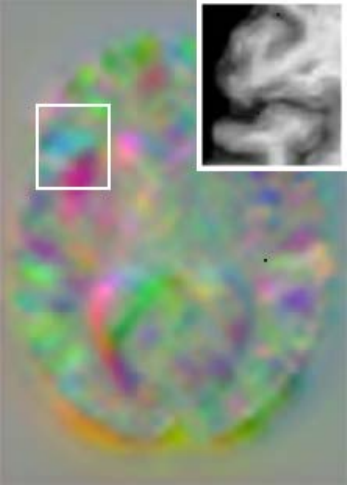}
		&\includegraphics[width=0.1152\textwidth]{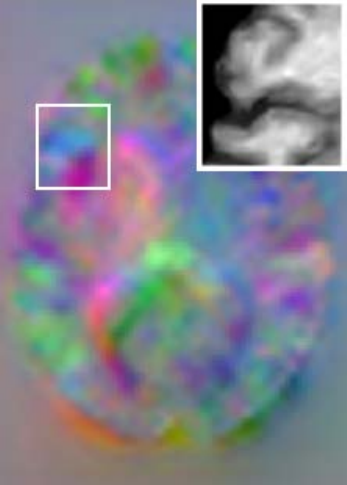}			
		&\includegraphics[width=0.1152\textwidth]{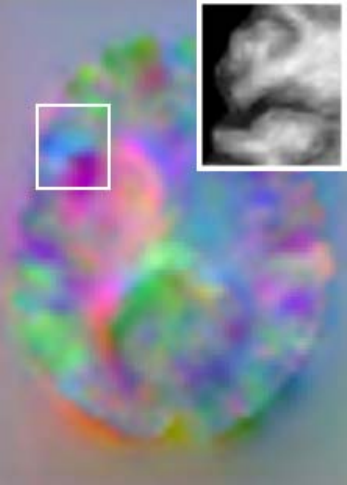}
		&\includegraphics[width=0.1152\textwidth]{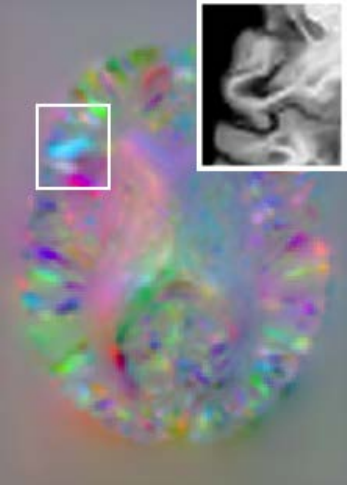}
		&\includegraphics[width=0.1152\textwidth]{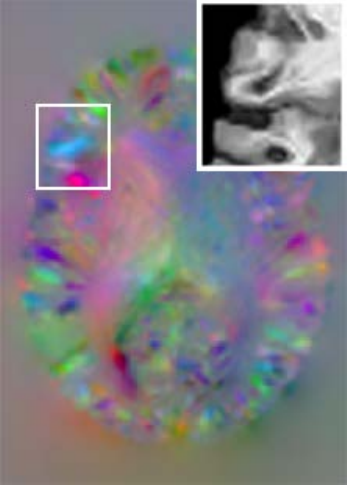}
		&\includegraphics[width=0.1152\textwidth]{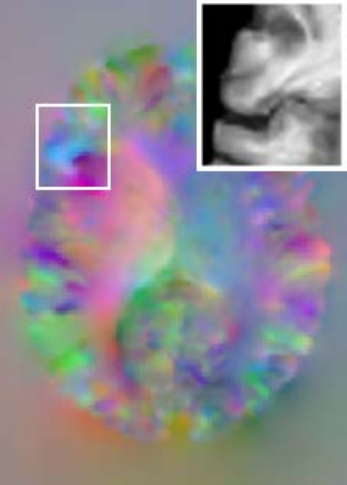}\\
		
		Target & Source &$\bm{u}^{1}$ & $\bm{v}^{1}$ &  $\bm{\varphi}^{1}$ & $\bm{u}^{2}$ & $\bm{v}^{2}$ & $\bm{\varphi}^{2}$\\
	\end{tabular}
	\caption{ The evolution of deformation color maps and registered images with the propagation of registration fields in the first two scales.} 
	\label{fig:multi-step}
\end{figure*}

\textbf{Implementation.}~
The affine network progressively downsamples the input with 9 convolutions and employs a fully-connected layer to produce 12 numeric parameters, composed of a $3 \times 3$ transform matrix $A$ and a 3D vector $b$. 
As for the propagation network, we use filters of size $3 \times 3 \times 3$ for all the convolutional layers. All convolutions are followed by a leaky ReLU function except the one that outputs the registration field.
To generate feature representations, we use layers of convolutional filters to downsample the features at the previous pyramid level by a factor of 2.
Computation on the full resolution may easily exhaust the memory, thus we choose to output a half-resolution smooth enough deformation field and up-sample it via interpolation to obtain the full-resolution deformation field. 
The proposed method~\footnote{Our codes are available at https://github.com/dut-media-lab/MultiPropReg} is implemented with Pytorch~\cite{PaszkeGMLBCKLGA19} package. 

\textbf{Evaluation metrics.}~
To achieve comprehensive evaluation, both the average Dice score~\cite{Dice} over registered testing pairs and the Jacobian matrix over the computed deformation are considered as evaluation metrics, evaluating the anatomical overlap correspondences of the registered volume pairs and the smoothness of the deformation fields, respectively.
The Dice score of two regions $A, B$ is formulated as:
\begin{equation}
Dice(A, B) = 2 \cdot \frac{|A \cap B |}{|A| + |B|},
\end{equation}
where a Dice score of 1 means the most perfectly overlap.
The Jacobian matrix $\mathbf{J_{\phi}(x) = \nabla \phi(x) }$ captures the local properties of $\phi$ around voxel $\mathbf{x}$. According to~\cite{Ashburner07}, the deformation is diffeomorphic at the locations where $\mathbf{ J_{\phi}(x) > 0}$. 
We count all the folds, where $\mathbf{J_{\phi}(x) \le 0}$.
In addition, we compute the average Normalized Correlation Coefficient (NCC) between image pairs as an auxiliary evaluation metric.

\begin{table}[!t]
	\centering
	\caption{ Ablation analysis of the bilevel self-tuned training strategy on five brain MRI datasets and two liver CT datasets in terms of Dice score and NCC. The default model represents the model that uses traditional training with manually selected hyper-parameters. }
	\vspace{-0.5em}
	\begin{tabular}{|m{1.0cm}<{\centering} |m{1.4cm}<{\centering} |m{1.9cm}<{\centering} |m{1.9cm}<{\centering} |}
		\hline
		\multicolumn{2}{|c |}{Methods}       &  Default      &  Bilevel  \\
		\hline
		\multirow{2}*{OASIS} & Dice score   & \textbf{0.777 $\pm$ 0.006}  & 0.776 $\pm$ 0.007   \\
		~ & NCC   & 0.245 $\pm$ 0.002  & \textbf{0.257 $\pm$ 0.002}   \\ \hline
		\multirow{2}*{ABIDE} & Dice score    & 0.764 $\pm$ 0.015  & \textbf{0.770 $\pm$ 0.013}   \\
		~ & NCC   & 0.241 $\pm$ 0.004  & \textbf{0.251 $\pm$ 0.003}   \\ \hline
		\multirow{2}*{ADNI} & Dice score     & 0.773 $\pm$ 0.017  & \textbf{0.775 $\pm$ 0.016}   \\
		~ & NCC     & 0.244 $\pm$ 0.005  & \textbf{0.255 $\pm$ 0.004}   \\ \hline
		\multirow{2}*{PPMI} & Dice score     & 0.785 $\pm$ 0.011  & \textbf{0.785 $\pm$ 0.011}   \\
		~ & NCC    & 0.240 $\pm$ 0.004  & \textbf{0.254 $\pm$ 0.003}   \\ \hline
		\multirow{2}*{HCP} & Dice score     & 0.776 $\pm$ 0.010  & \textbf{0.776 $\pm$ 0.010}   \\
		~ & NCC   & 0.240 $\pm$ 0.004  & \textbf{0.250 $\pm$ 0.003}   \\ \hline	
		\multirow{2}*{SLIVER} & Dice score    & 0.883 $\pm$ 0.042  & \textbf{0.910 $\pm$ 0.027}   \\
		~ & NCC    & 0.228 $\pm$ 0.153  & \textbf{0.374 $\pm$ 0.035}   \\
		\hline
		\multirow{2}*{LSPIG} & Dice score      & 0.822 $\pm$ 0.061  & \textbf{0.855 $\pm$  0.045}   \\
		~ & NCC     & 0.144 $\pm$ 0.124  & \textbf{0.348 $\pm$ 0.056}   \\ \hline
	\end{tabular}
	\label{tab:ablation_HO}
\end{table}

\begin{figure*}[t]
	\centering
	\begin{tabular}{c@{\extracolsep{0.22em}}c@{\extracolsep{0.22em}}c@{\extracolsep{0.22em}}c@{\extracolsep{0.22em}}c@{\extracolsep{0.22em}}c@{\extracolsep{0.22em}}c}
		
		\includegraphics[width=0.135\textwidth]{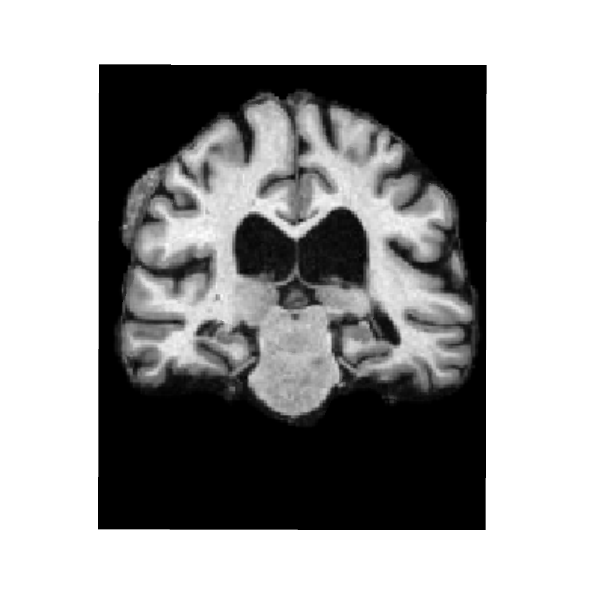}
		&\includegraphics[width=0.135\textwidth]{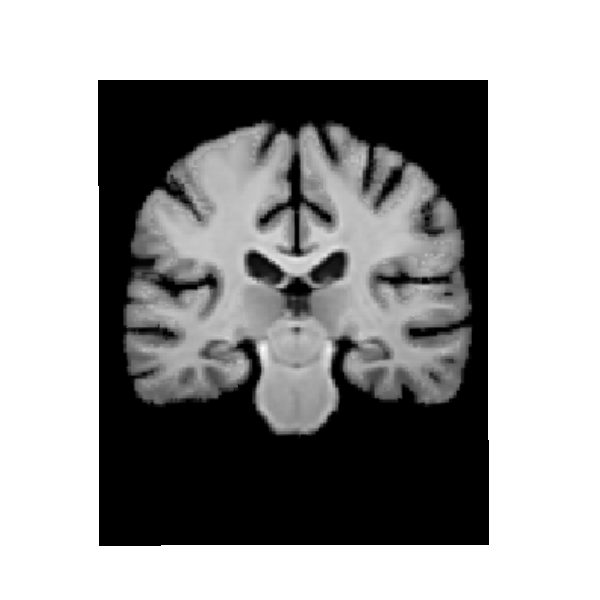}
		&\includegraphics[width=0.135\textwidth]{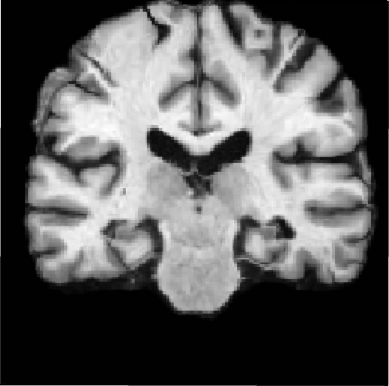}
		&\includegraphics[width=0.135\textwidth]{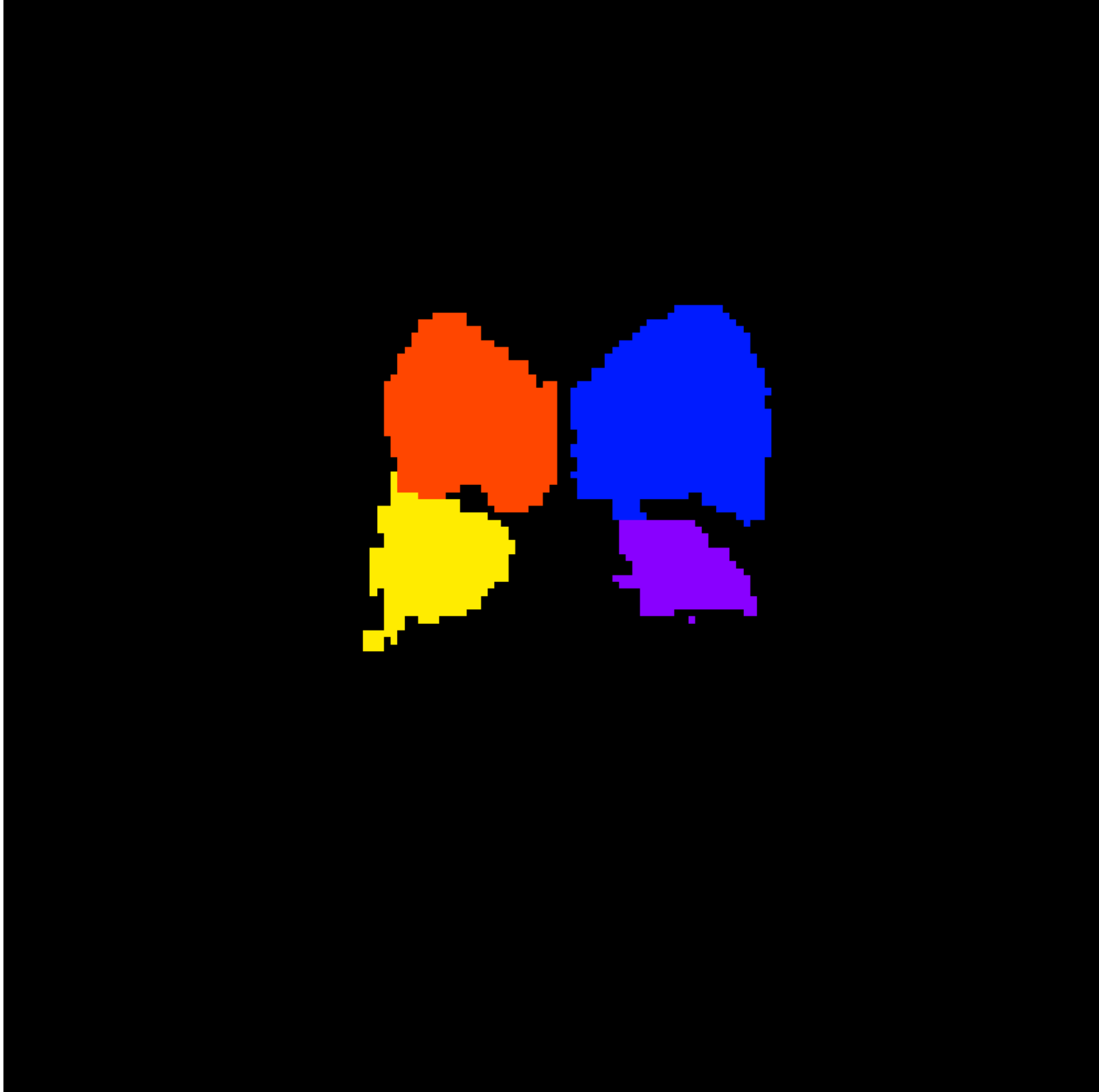}
		&\includegraphics[width=0.135\textwidth]{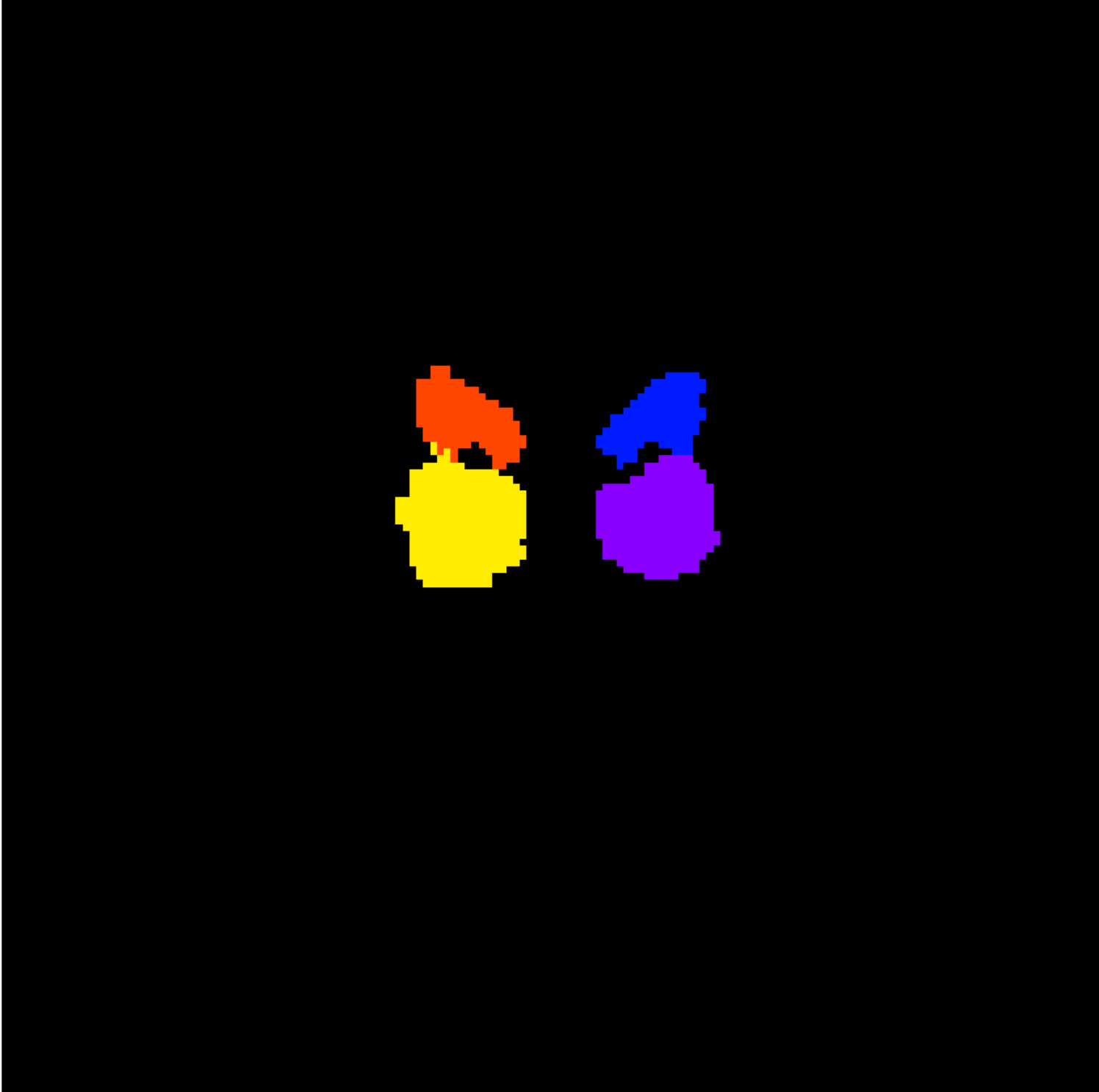}
		&\includegraphics[width=0.135\textwidth]{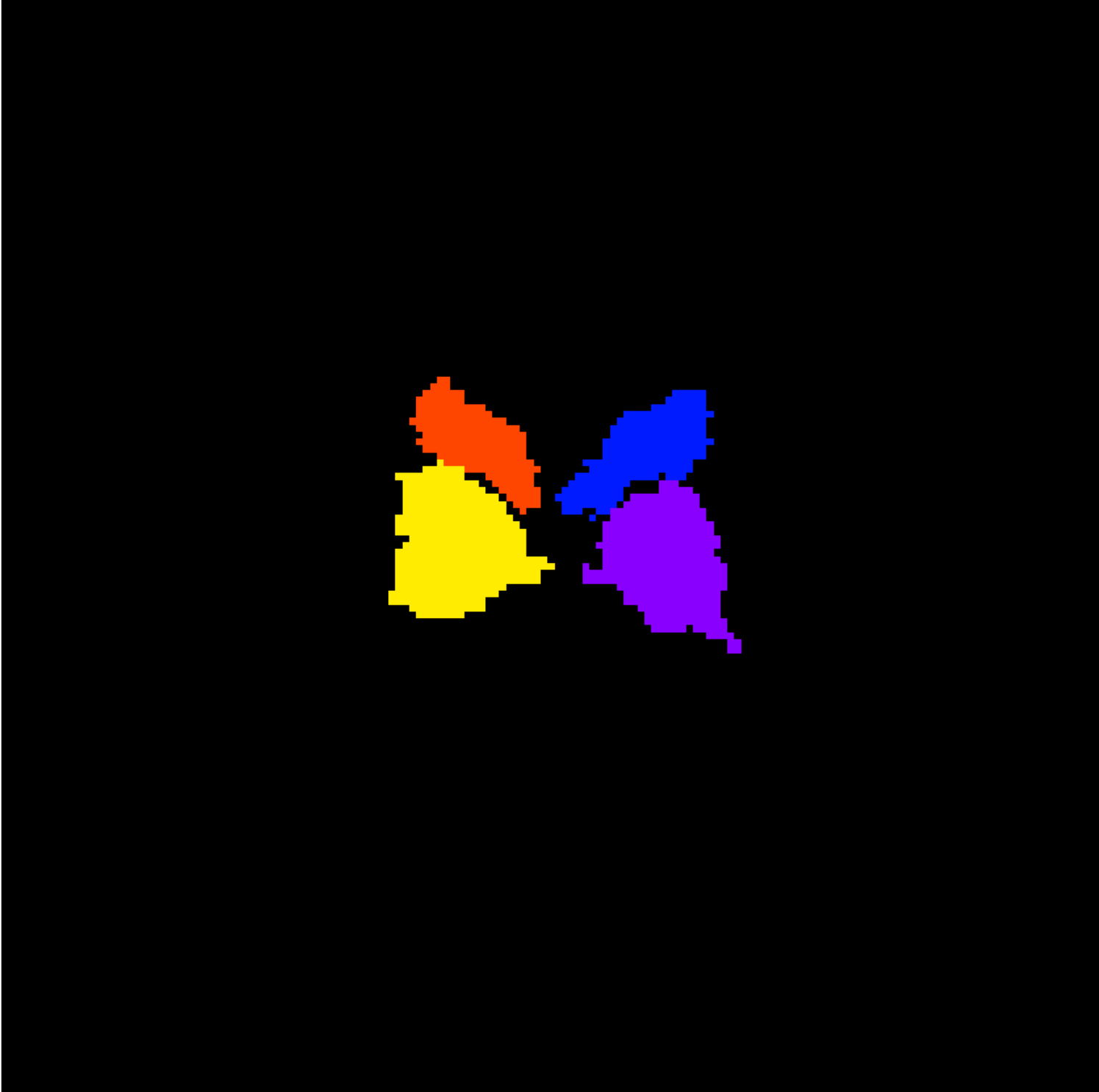}
		&\includegraphics[width=0.135\textwidth]{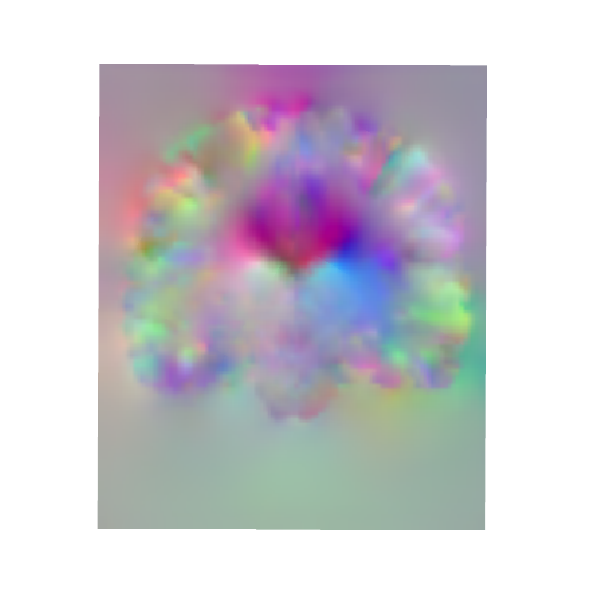}\\

		\includegraphics[width=0.135\textwidth]{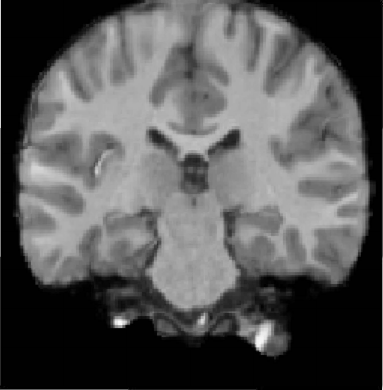}
		&\includegraphics[width=0.135\textwidth]{figures/Supplementary_material/crop/fixed}
		&\includegraphics[width=0.135\textwidth]{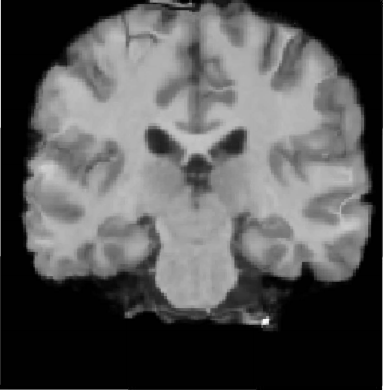}
		&\includegraphics[width=0.135\textwidth]{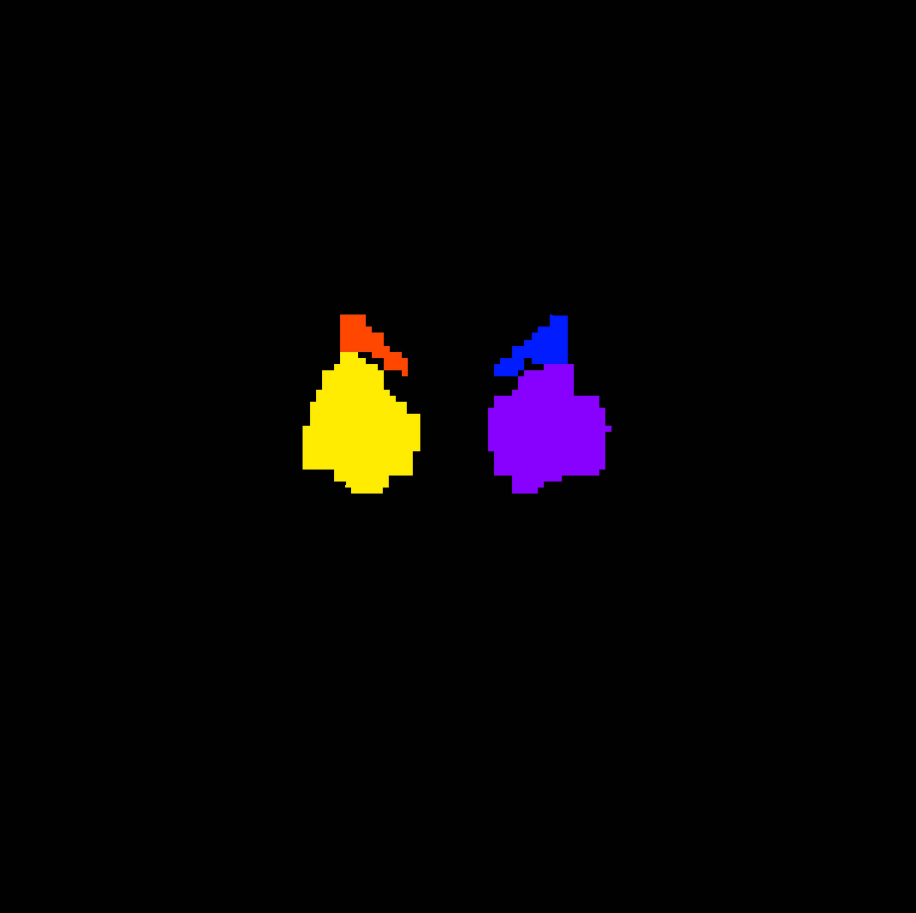}
		&\includegraphics[width=0.135\textwidth]{figures/Supplementary_material/crop/fixed_seg}
		&\includegraphics[width=0.135\textwidth]{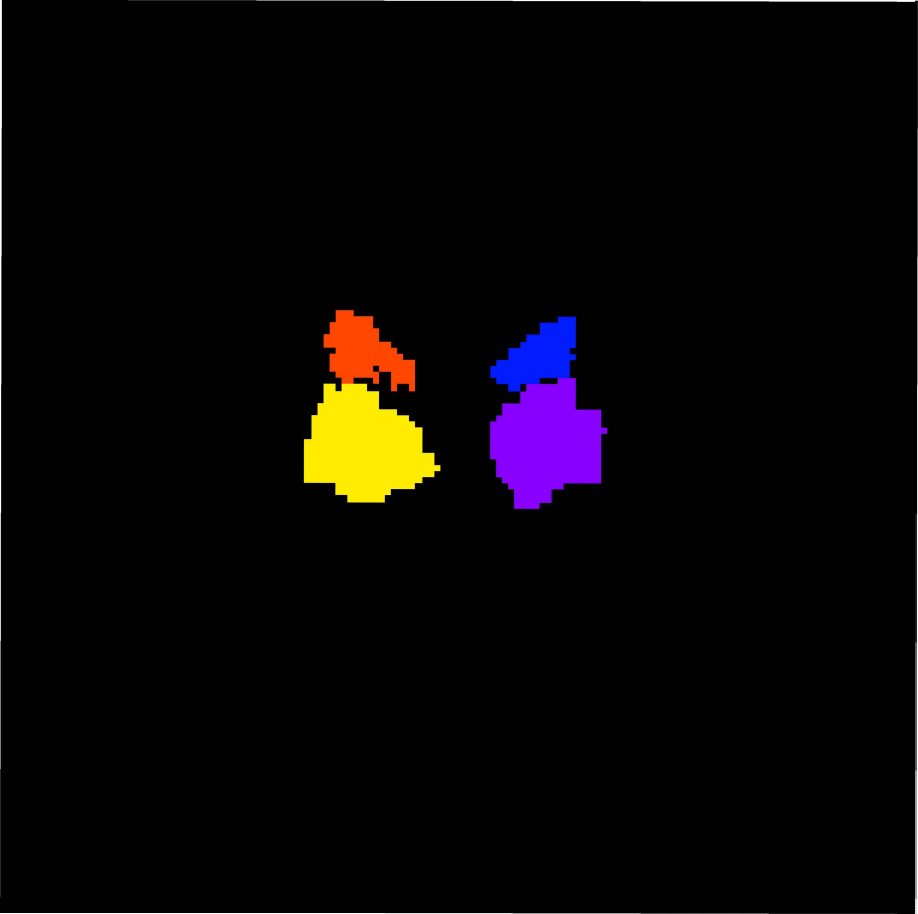}
		&\includegraphics[width=0.135\textwidth]{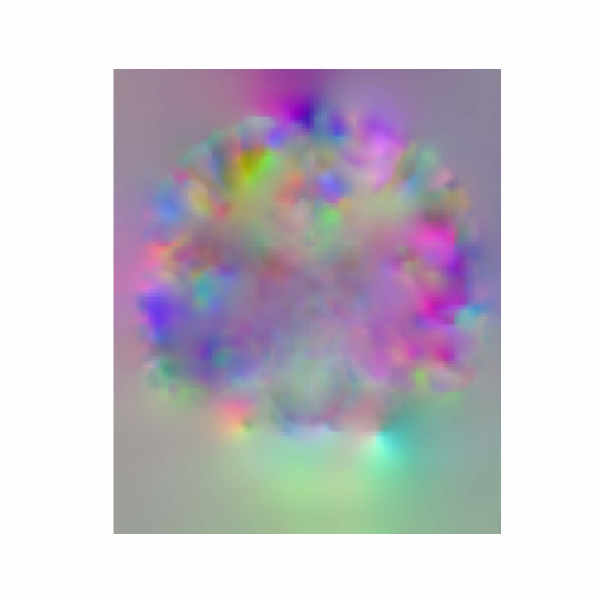}\\
		
		Source  & Target  &  Warped  & Source-label &  Target-label & Warped-label & Flow field \\
		
	\end{tabular}
	\caption{ Example MR coronal slices of input target, source and warped image for our method with corresponding labels of ventricles, thalami, and hippocampi. The last column shows the RGB image of the registration field. Each row refers to an example registration case of brain MR data. }
	\label{fig:example_mri} 
\end{figure*}

\begin{figure*}[!t]
	\centering 
	\includegraphics[width=1.0\textwidth]{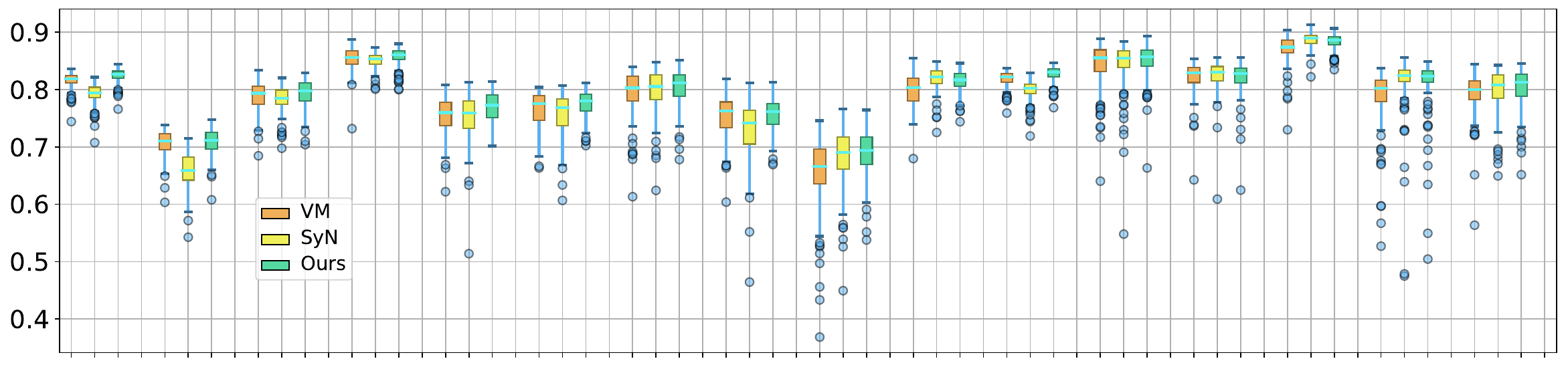} 
	\caption{ Boxplot indicates the Dice scores for SyN, VM and our algorithm over sixteen anatomical structures including Cerebral White Matter (CblmWM), Cerebral Cortex (CblmC), Lateral Ventricle (LV), Inferior Lateral Ventricle (ILV), Cerebellum White Matter (CeblWM), Cerebellum Cortex (CereC), Thalamus (Tha), Caudate (Cau), Putamen (Pu), Pallidum (Pa), Hippocampus (Hi), Accumbens area (Am), Vessel, Third Ventricle (3V), Fourth Ventricle (4V), and Brain Stem (BS).} 
	\label{fig:boxplot2}
\end{figure*}

\begin{table*}[!htb]
	\centering
	\caption{ Qualitative comparison results on brain MR registration tasks. The mean and standard deviations of the Dice score on five different datasets are listed. The average Dice score is computed over all the structures and subjects.   }
	\vspace{-0.5em}
	\begin{tabular}{|m{1.0cm}<{\centering}| m{1.8cm}<{\centering}|m{1.8cm}<{\centering}|m{1.8cm}<{\centering}| m{1.8cm}<{\centering}| m{1.8cm}<{\centering}| m{1.8cm}<{\centering}| m{1.8cm}<{\centering}|}
		\hline
		Methods  & Affine only         & Elastix        &  NiftyReg      &  SyN      & VM    & VM-diff    & Ours  \\
		\hline
		OASIS   & 0.580 $\pm$ 0.028  & 0.709 $\pm$ 0.023  & 0.748 $\pm$ 0.017 & 0.765 $\pm$ 0.010  & 0.765 $\pm$ 0.010 & 0.757 $\pm$ 0.011  & \textbf{0.777 $\pm$ 0.006}   \\
		ABIDE   & 0.624 $\pm$ 0.024  & 0.699 $\pm$ 0.025  & 0.747 $\pm$ 0.026  & 0.728  $\pm$ 0.029  & 0.754 $\pm$ 0.016 & \textbf{0.773 $\pm$ 0.009}  & 0.764 $\pm$ 0.016   \\
		ADNI   & 0.571 $\pm$ 0.049  & 0.697 $\pm$ 0.039  & 0.737 $\pm$ 0.035  & 0.761 $\pm$ 0.021  & 0.761 $\pm$ 0.024 & 0.768 $\pm$ 0.020  & \textbf{0.773 $\pm$ 0.017}   \\
		PPMI   & 0.610 $\pm$ 0.033  & 0.730 $\pm$ 0.021  & 0.765 $\pm$ 0.015  & 0.778 $\pm$ 0.013  & 0.775 $\pm$ 0.013 & 0.781 $\pm$ 0.011  & \textbf{0.787 $\pm$ 0.010}   \\
		HCP   & 0.666 $\pm$ 0.027  & 0.729 $\pm$ 0.017  & 0.768 $\pm$ 0.013  & 0.767 $\pm$ 0.016  & 0.768 $\pm$ 0.013 & 0.413 $\pm$ 0.111  & \textbf{0.776 $\pm$ 0.010}   \\
		\hline
	\end{tabular}
	\label{tab:compare_mri}
\end{table*}

\textbf{State-of-the-art methods.}~
We compare the proposed method with state-of-the-art registration techniques, including  three optimization-based tools: Elastix~\cite{5338015}, Symmetric Normalization (SyN)~\cite{SyN}, NiftyReg~\cite{SunNK14}, and two learning-based methods: VoxelMorph~\cite{BalakrishnanZSG19} and its diffeomorphic variant~\cite{DalcaBGS19} (referred as VM and VM-diff, respectively). 
The parameter settings of the conventional methods are as follows.
For Elastix, we run B-spline registration with Mattes Mutual Information as a cost function and set the control point spacing to 16 voxels. Four scales are used with 500 iterations per scale.
For the SyN algorithm, we use the version implemented in the ANTs~\cite{AvantsTSCKG11} package and take Cross-Correlation as the similarity measure metric and use the SyN step size of 0.25, Gaussian parameters (9, 0.2), at three scales with 201 iterations each.
As for NiftyReg, we use the Normalized Mutual Information cost function. We run it with 12 threads using $1500$ iterations.
We run Elastix, SyN, and NiftyReg on a PC with i7-8700 (@3.20GHz, 32G RAM), while learning-based methods on NVIDIA TITAN XP.

\begin{figure*}[t]
	\centering
	\begin{tabular}{@{\extracolsep{0.2em}}c@{\extracolsep{0.2em}}c@{\extracolsep{0.2em}}c@{\extracolsep{0.2em}}c@{\extracolsep{0.2em}}c@{\extracolsep{0.2em}}c@{\extracolsep{0.2em}}c@{\extracolsep{0.2em}}c}
		
		\includegraphics[width=0.121\textwidth]{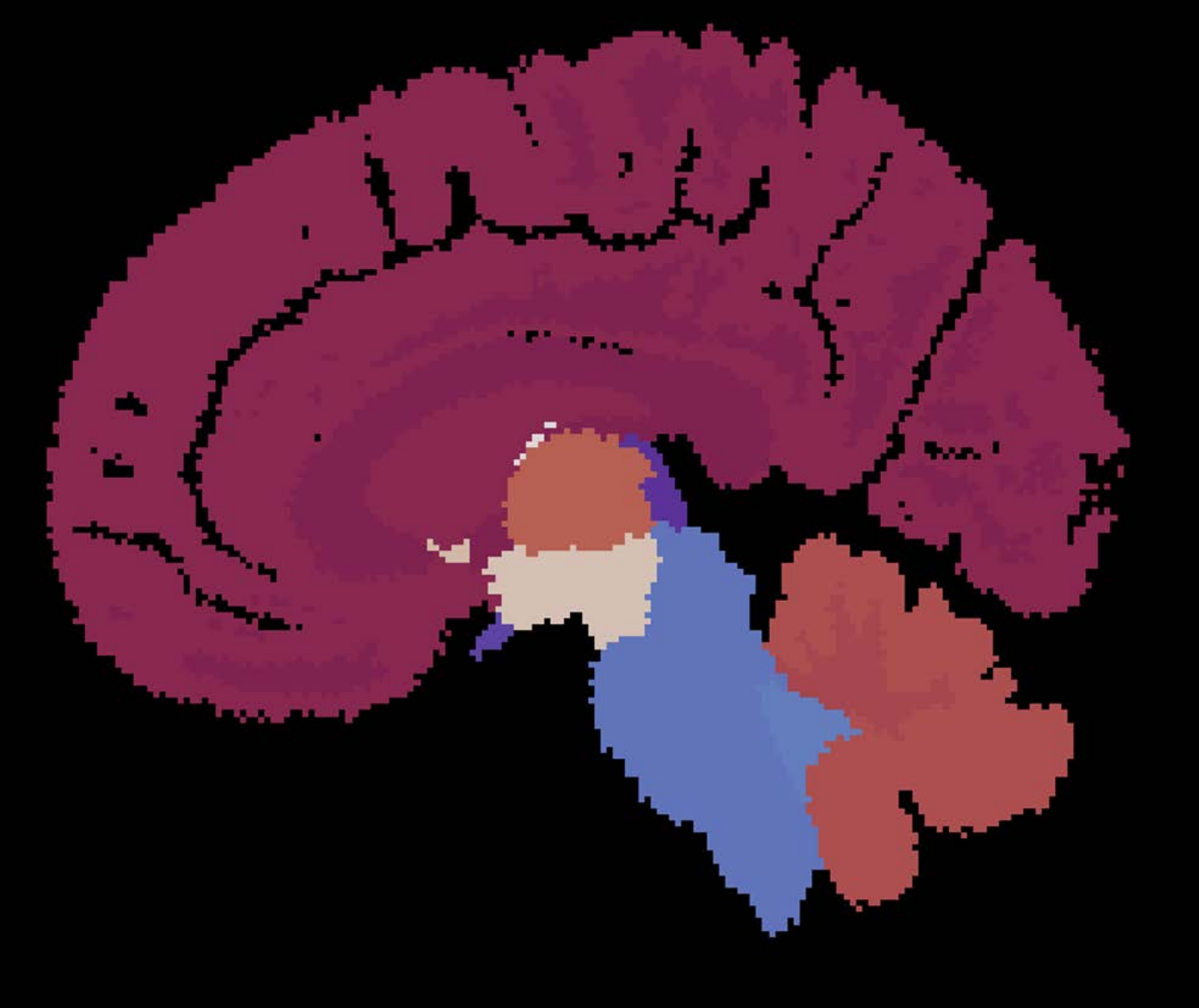}
		&\includegraphics[width=0.121\textwidth]{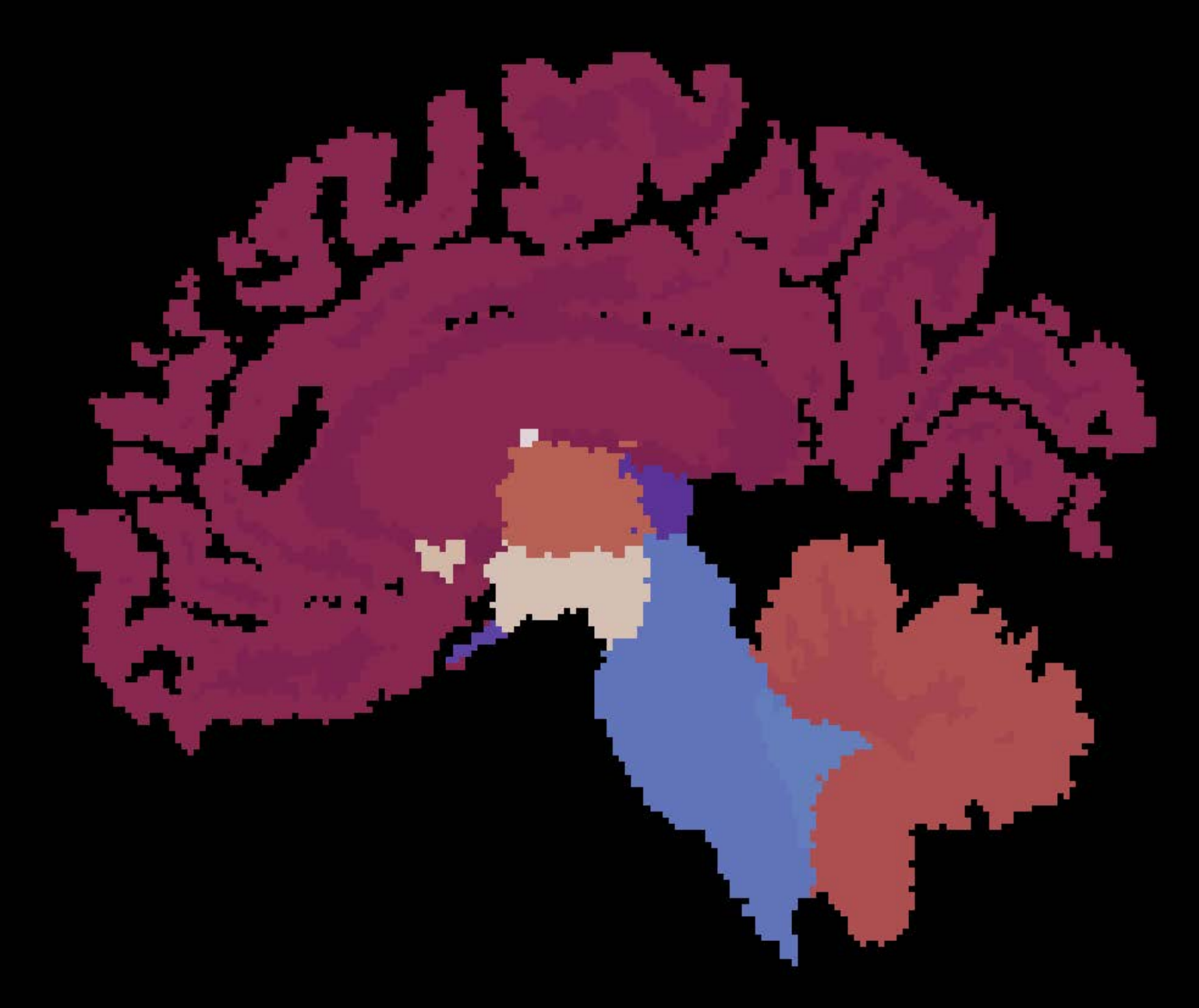}
		&\includegraphics[width=0.121\textwidth]{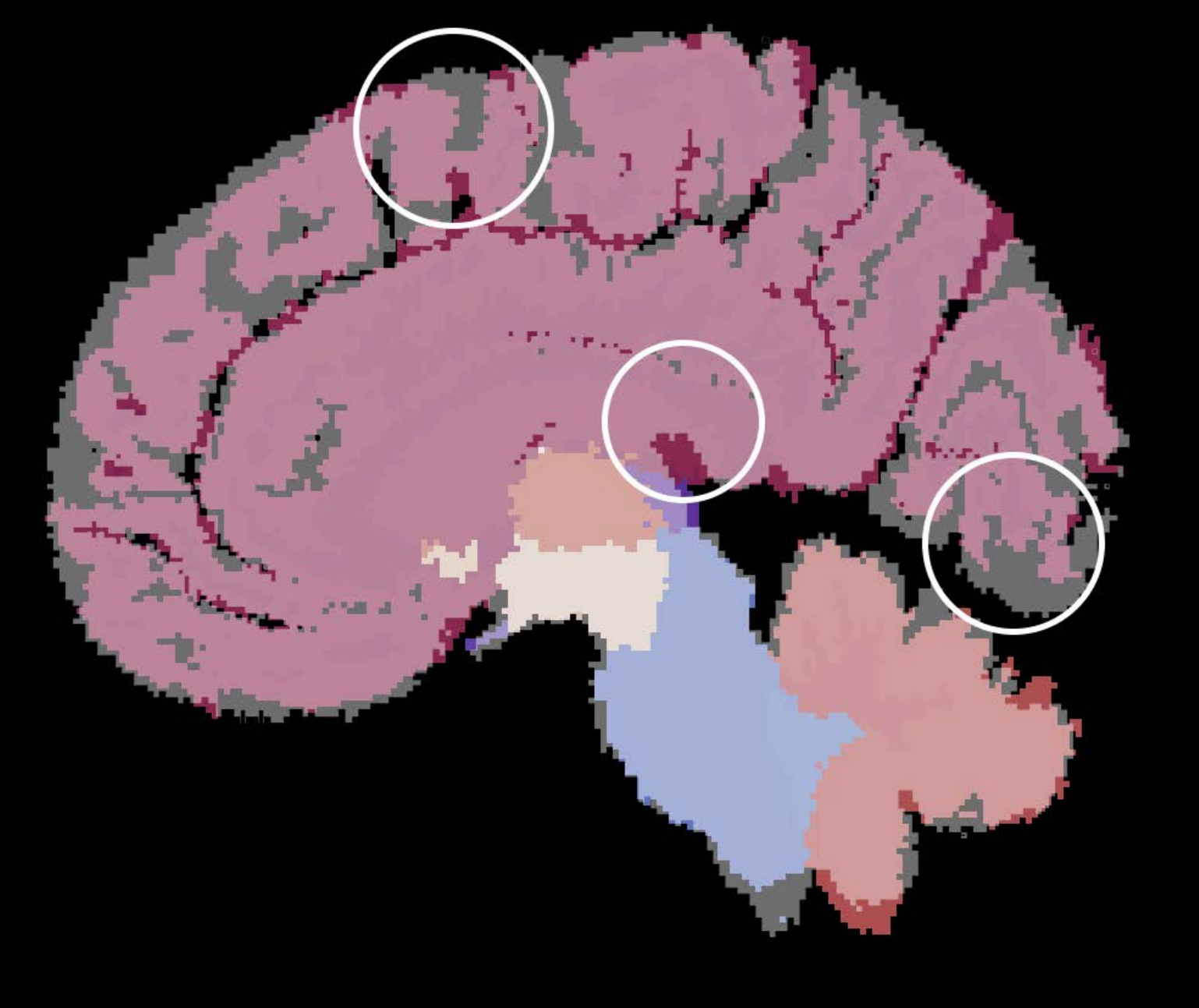}
		&\includegraphics[width=0.121\textwidth]{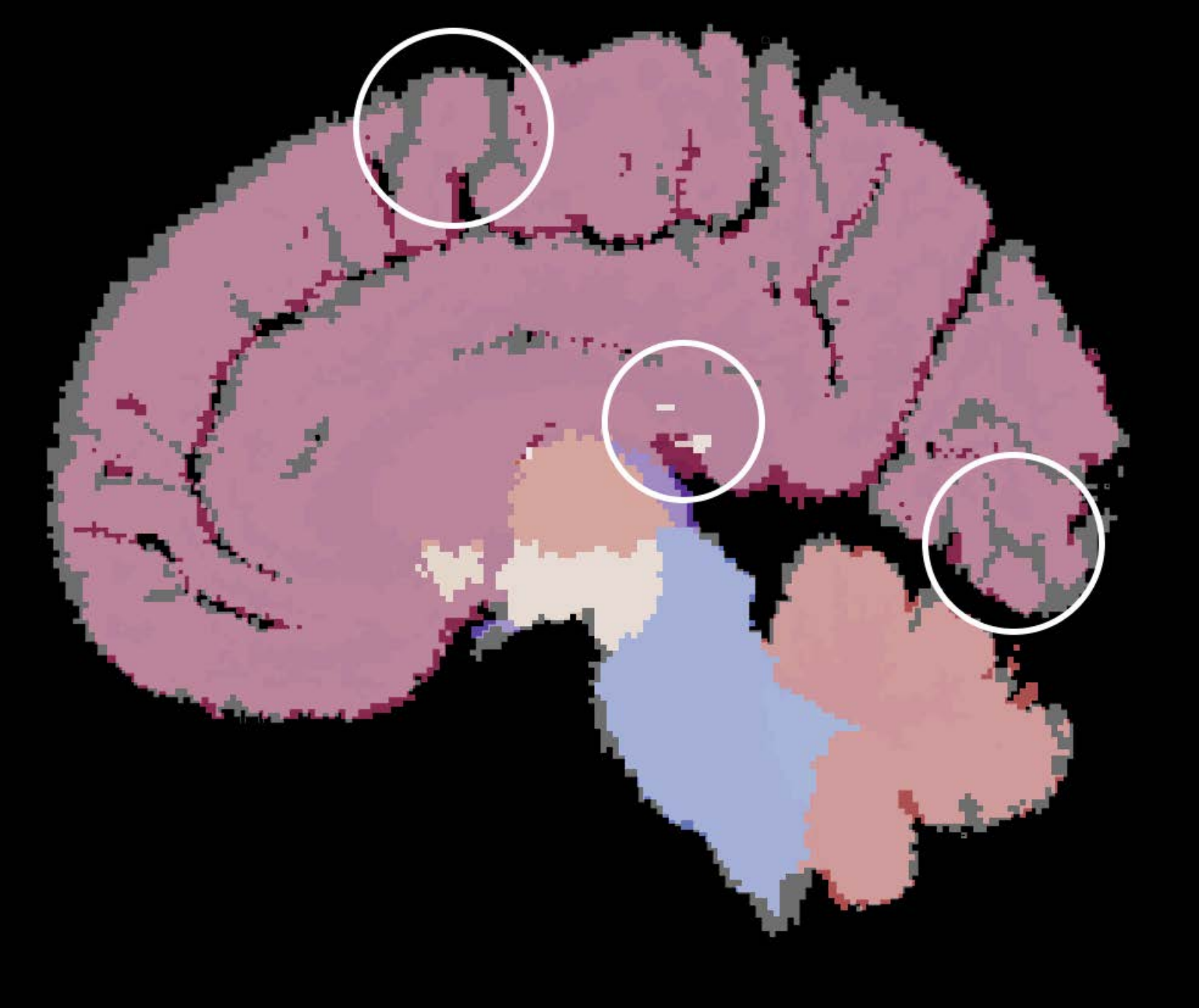}			
		&\includegraphics[width=0.121\textwidth]{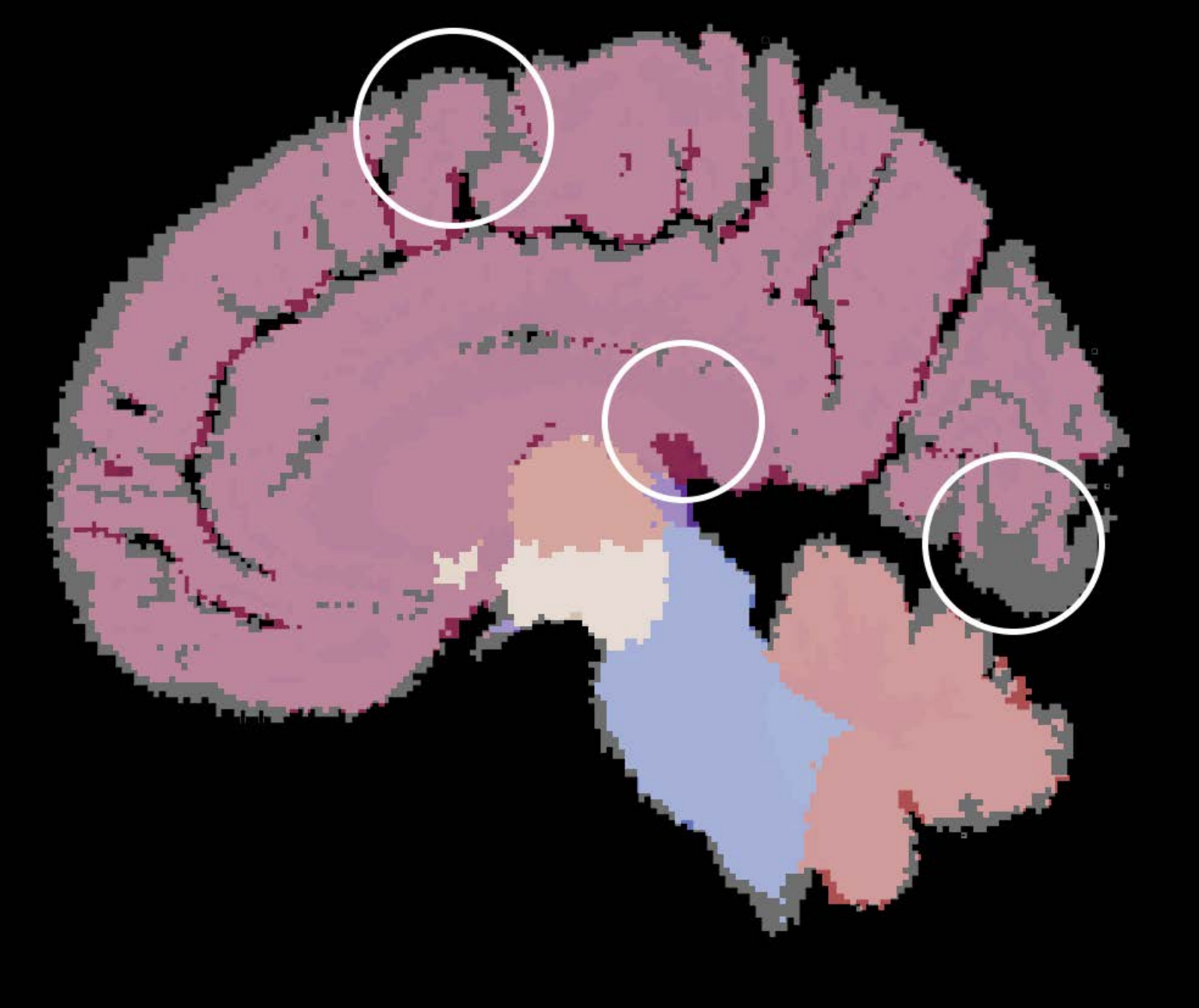}
		&\includegraphics[width=0.121\textwidth]{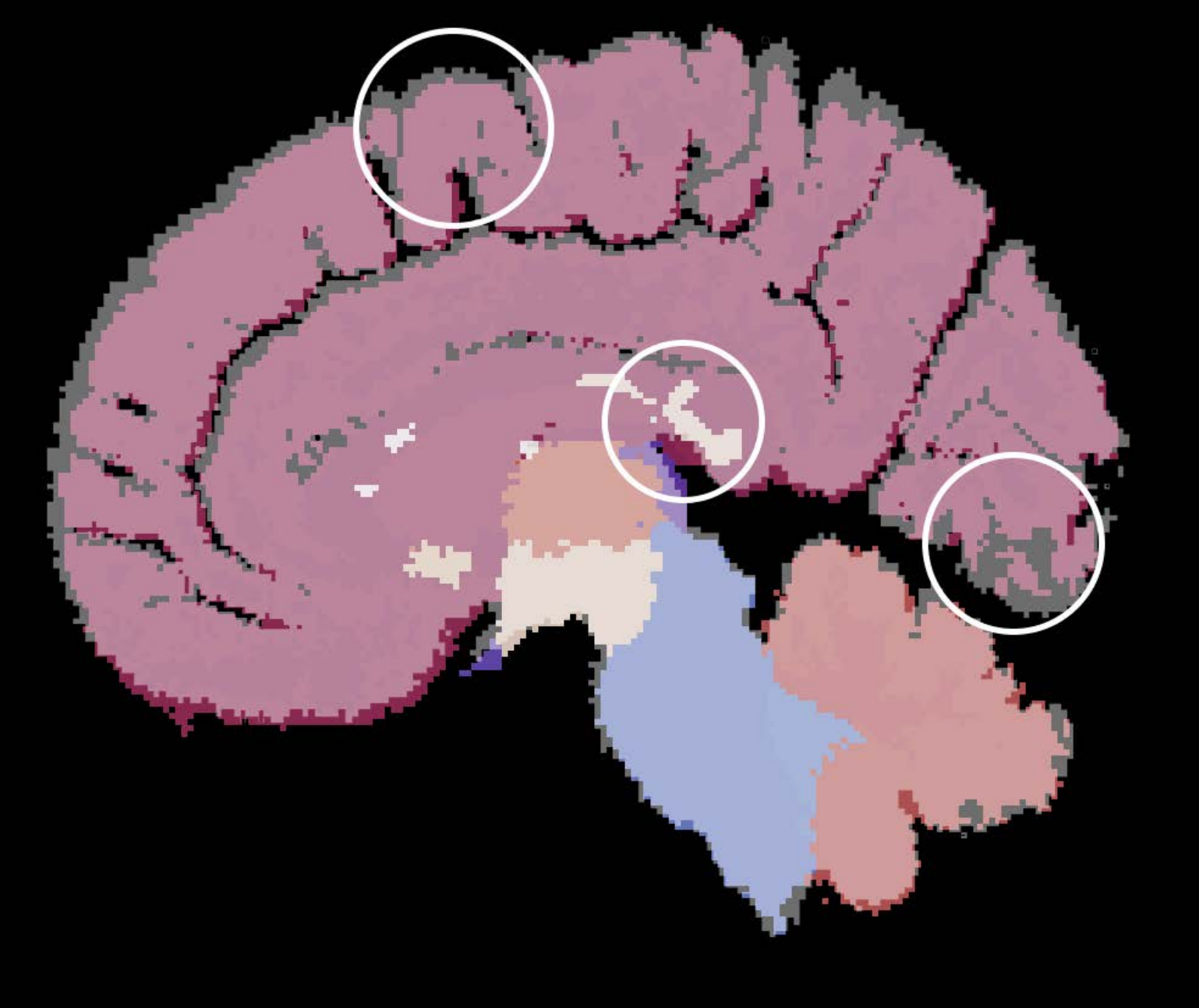}
		&\includegraphics[width=0.121\textwidth]{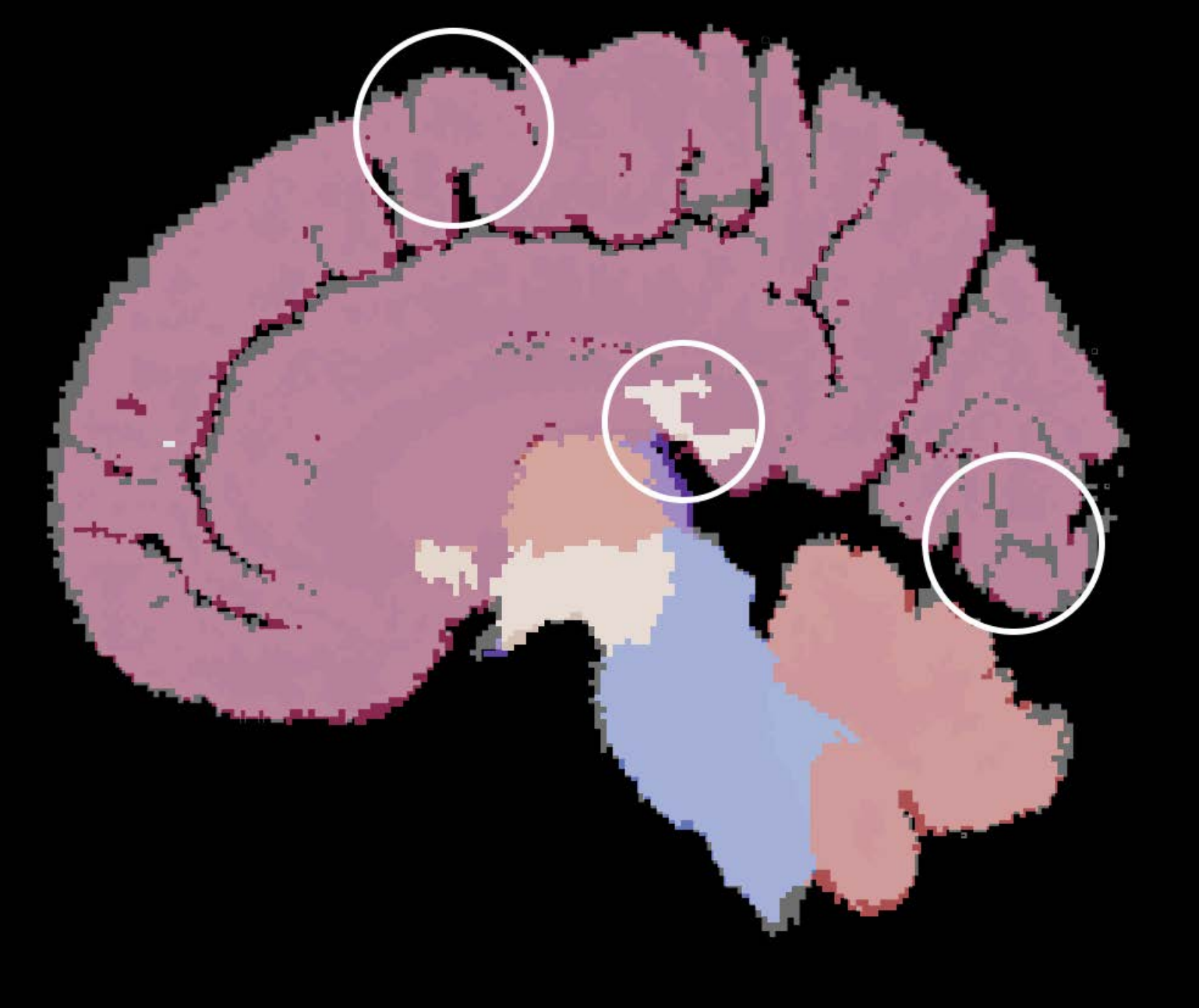}
		&\includegraphics[width=0.121\textwidth]{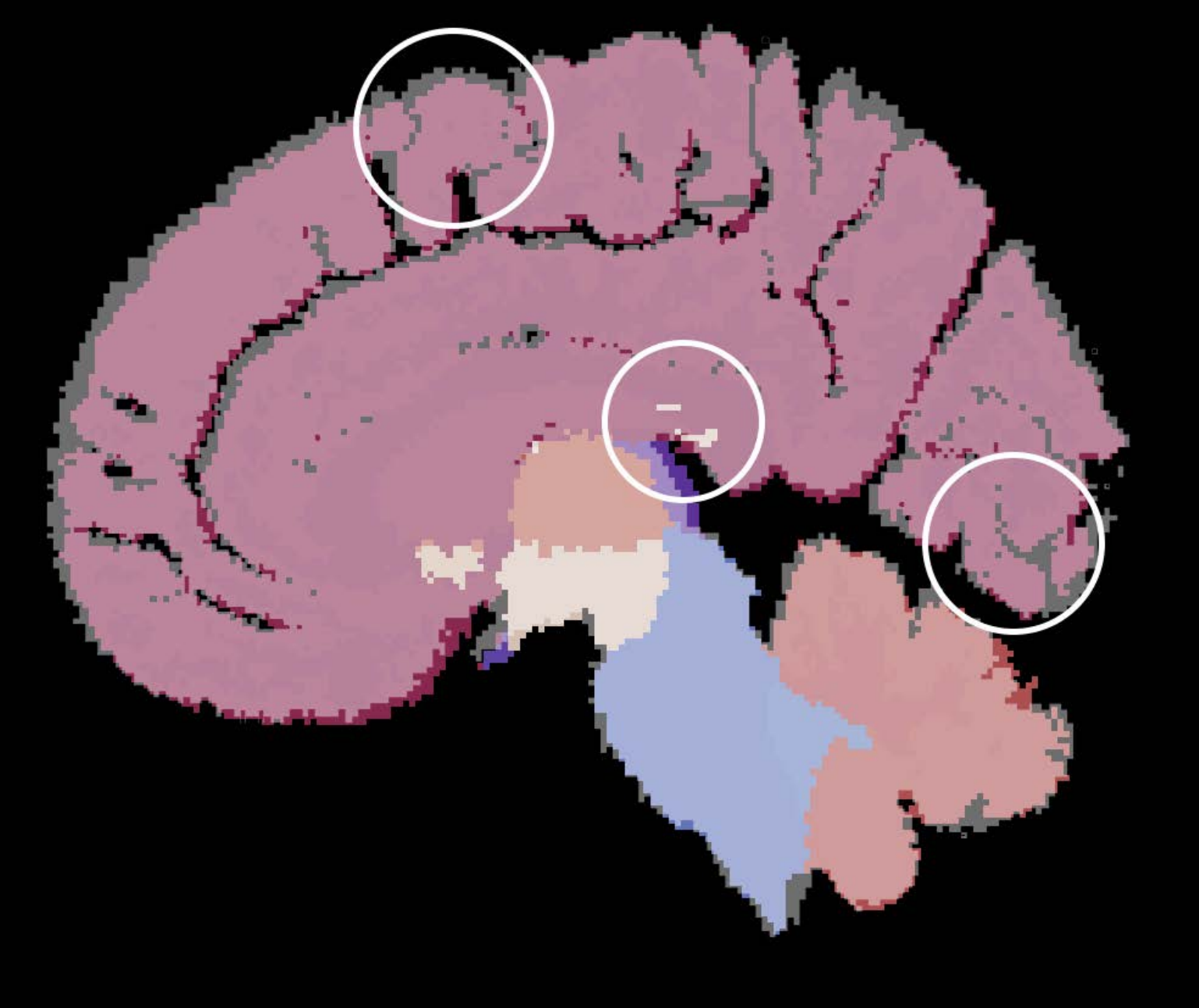}\\
		Target &  Source &  Elastix & SyN & NiftyReg & VM & VM-diff & Ours\\
		(1.0)~ & (0.570)~ & (0.709)~ & (0.763)~ & (0.741)~ & (0.761)~ & (0.755)~ & (\textbf{0.772})~\\
	\end{tabular}
	\caption{ Registered MR slices overlaid with atlas using different methods. The Dice scores are given in the bottom parentheses. Circles indicate several evident inconsistencies.} 
	\label{fig:compare_seg}
\end{figure*}

\begin{figure*}[!htp]
	\centering
	\begin{tabular}{@{\extracolsep{0.25em}}c@{\extracolsep{0.25em}}c@{\extracolsep{0.25em}}c@{\extracolsep{0.25em}}c@{\extracolsep{0.25em}}c}
		
		\includegraphics[width=0.194\textwidth]{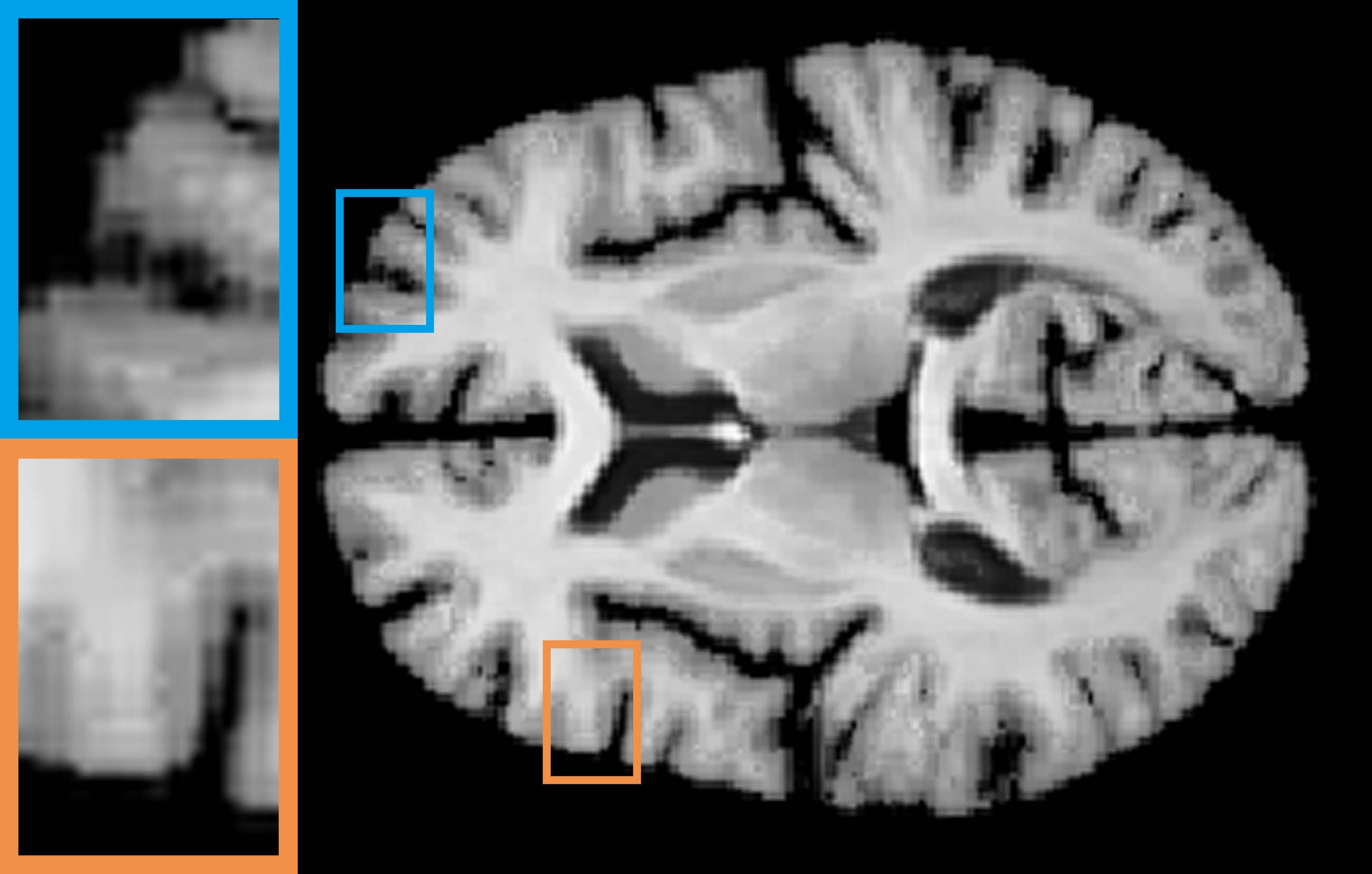}
		&\includegraphics[width=0.194\textwidth]{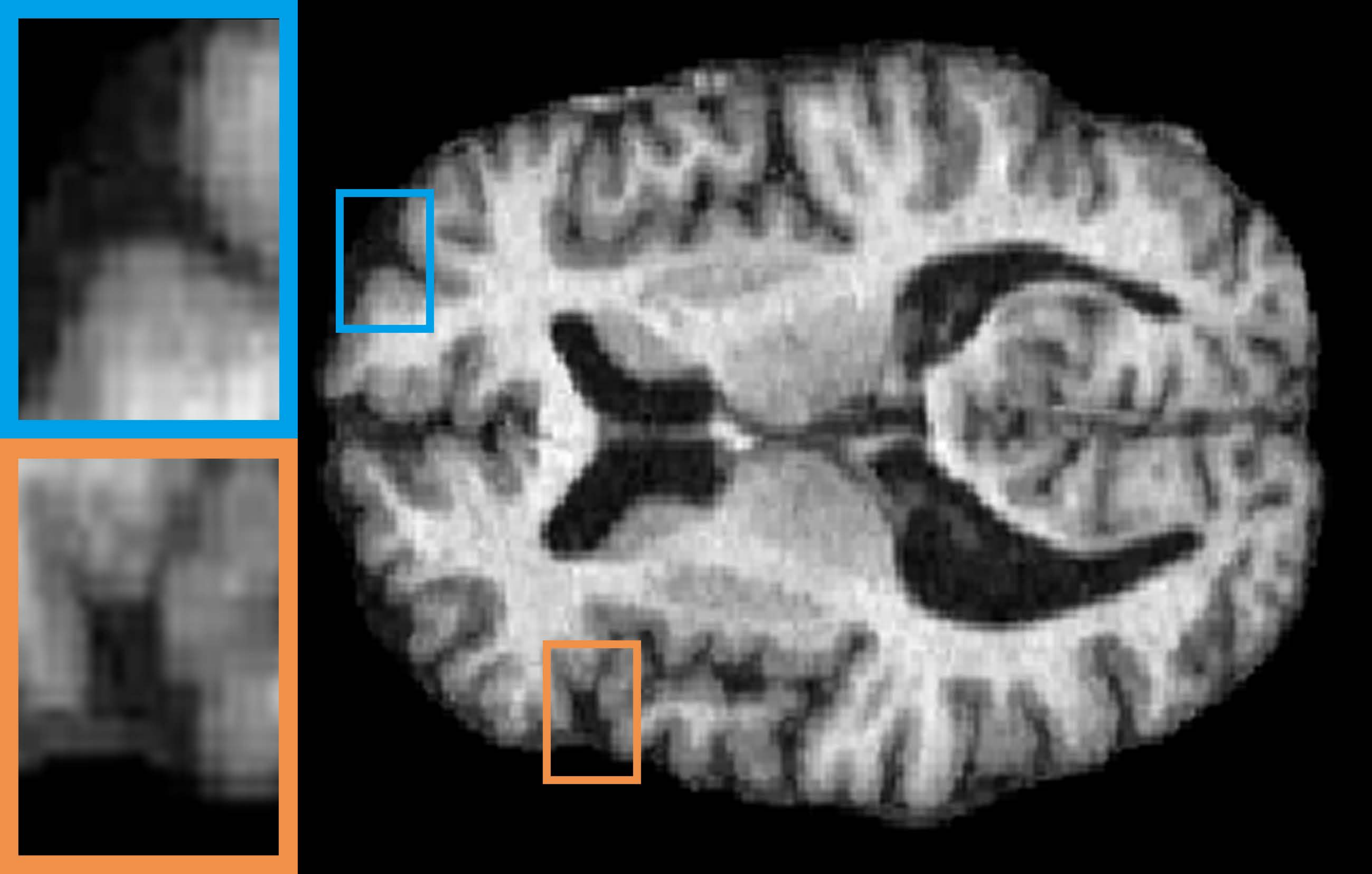}
		&\includegraphics[width=0.194\textwidth]{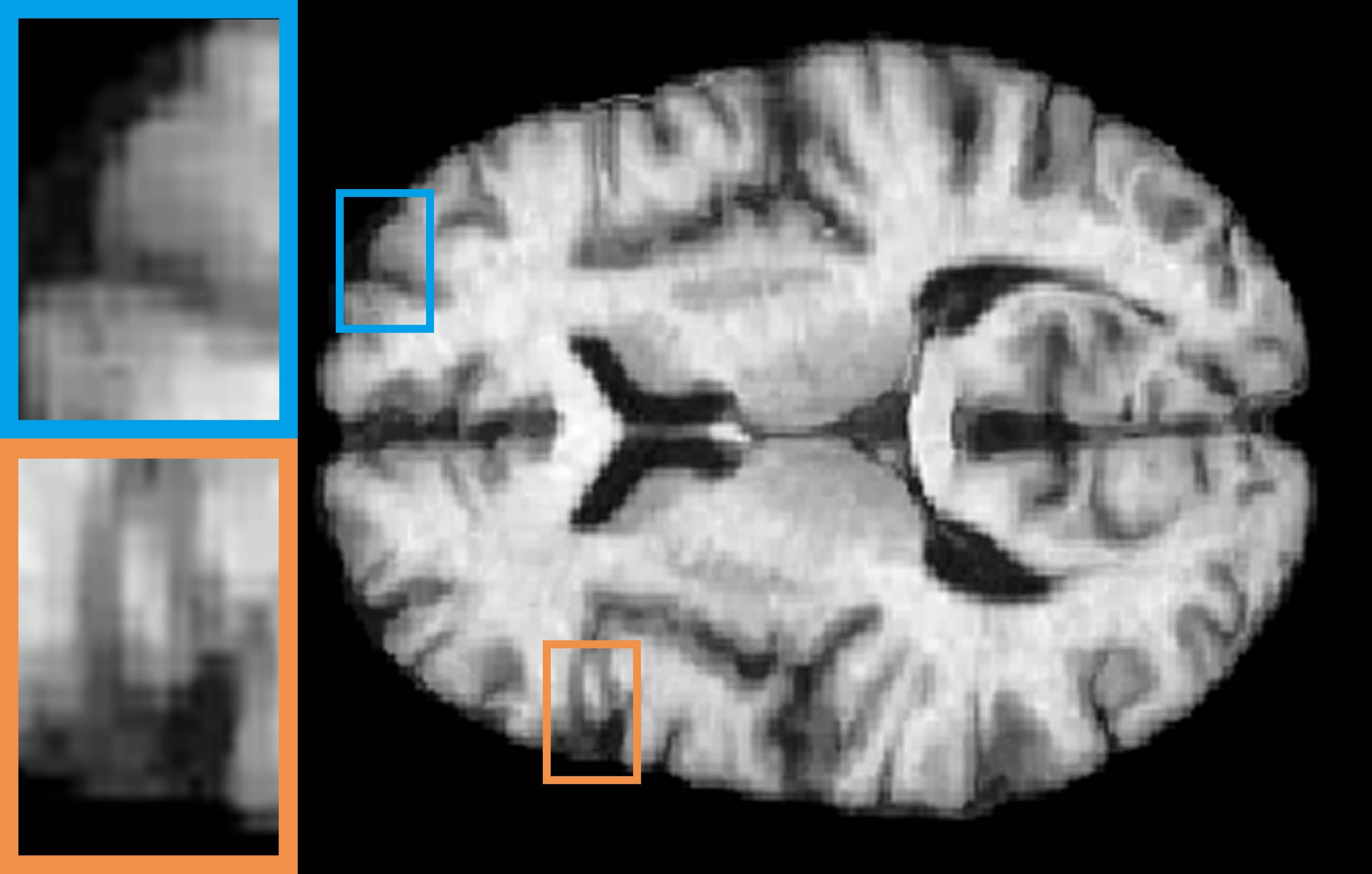}
		&\includegraphics[width=0.194\textwidth]{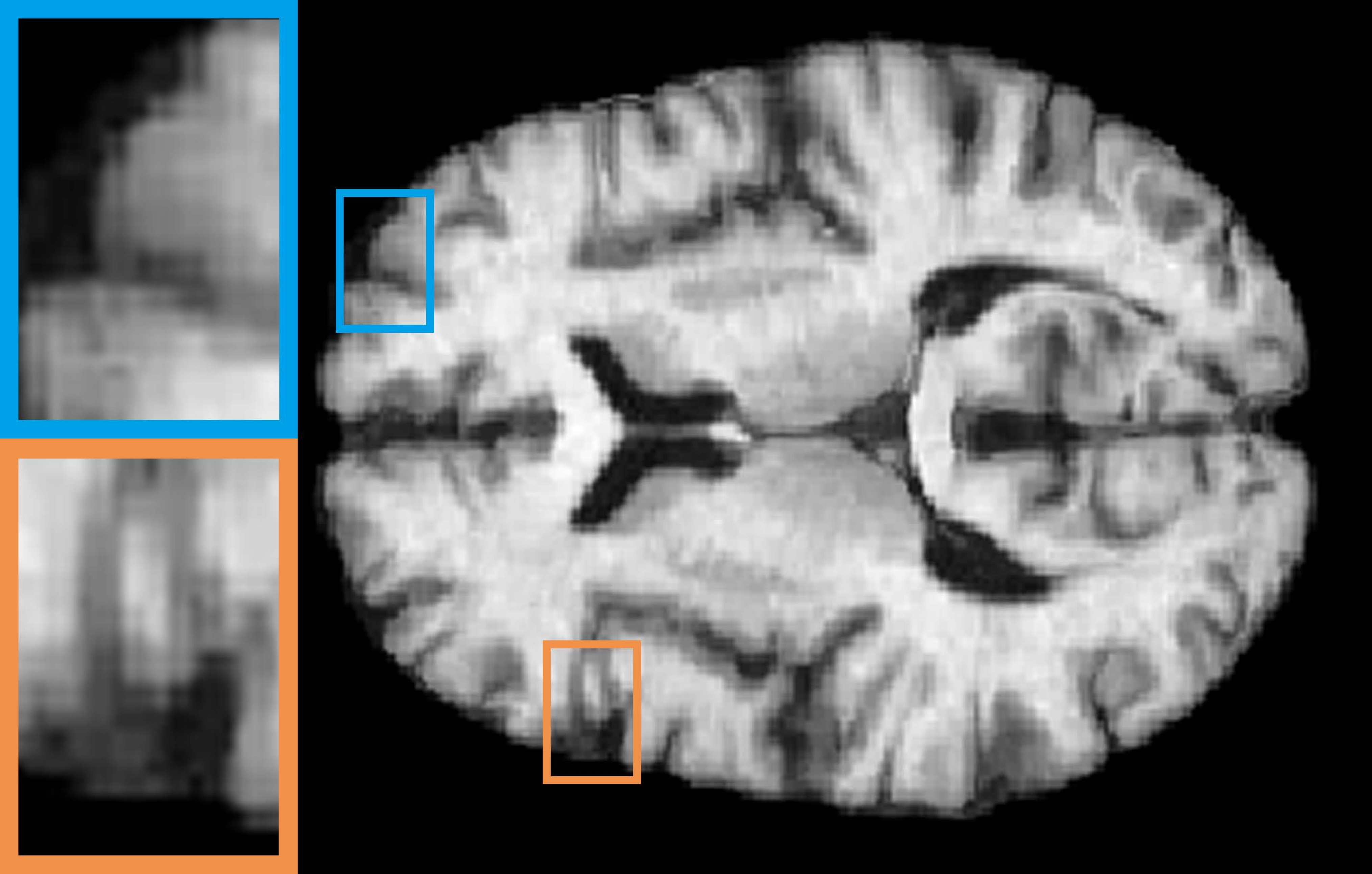}
		&\includegraphics[width=0.194\textwidth]{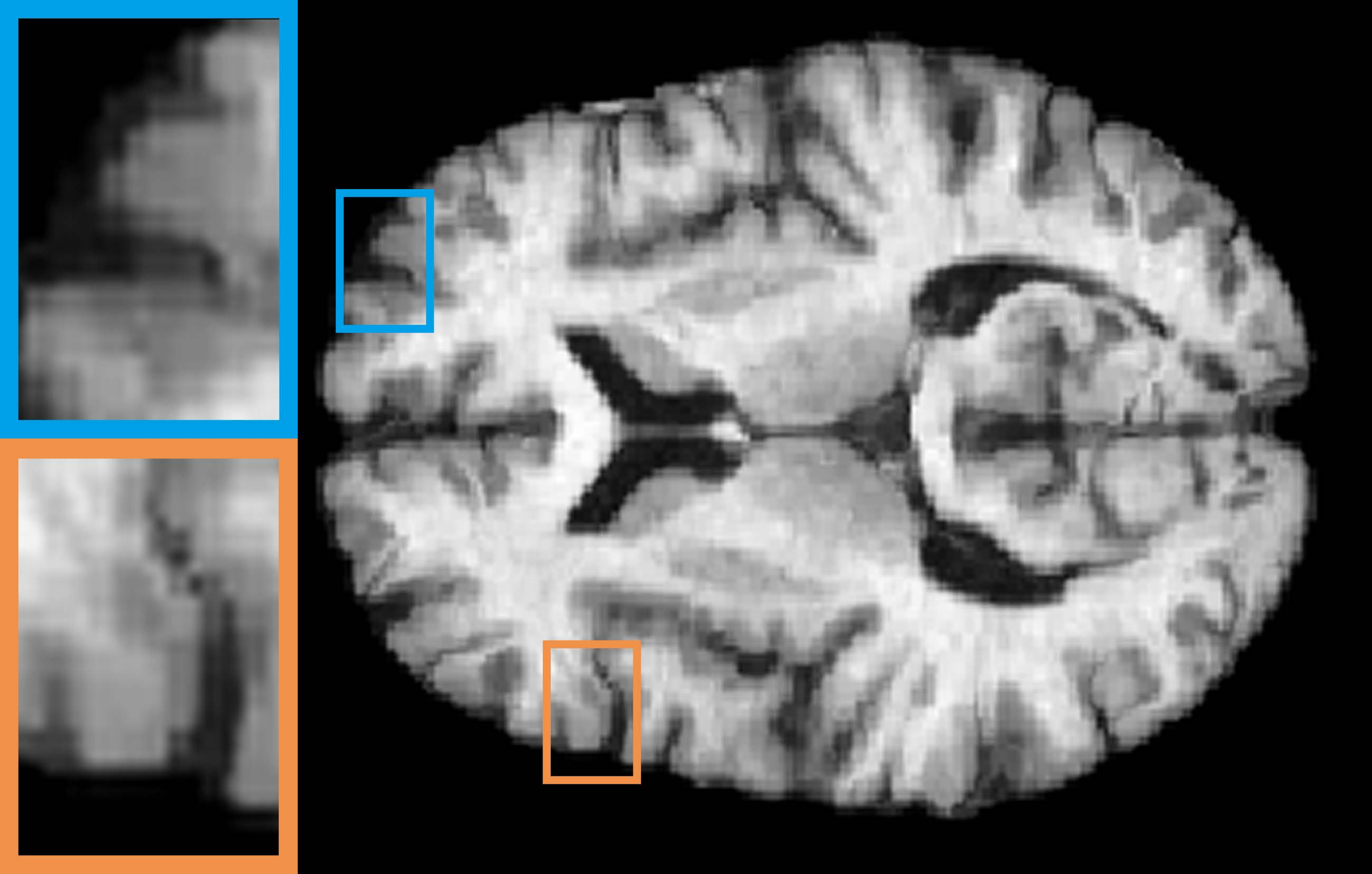}	\\
		\includegraphics[width=0.194\textwidth]{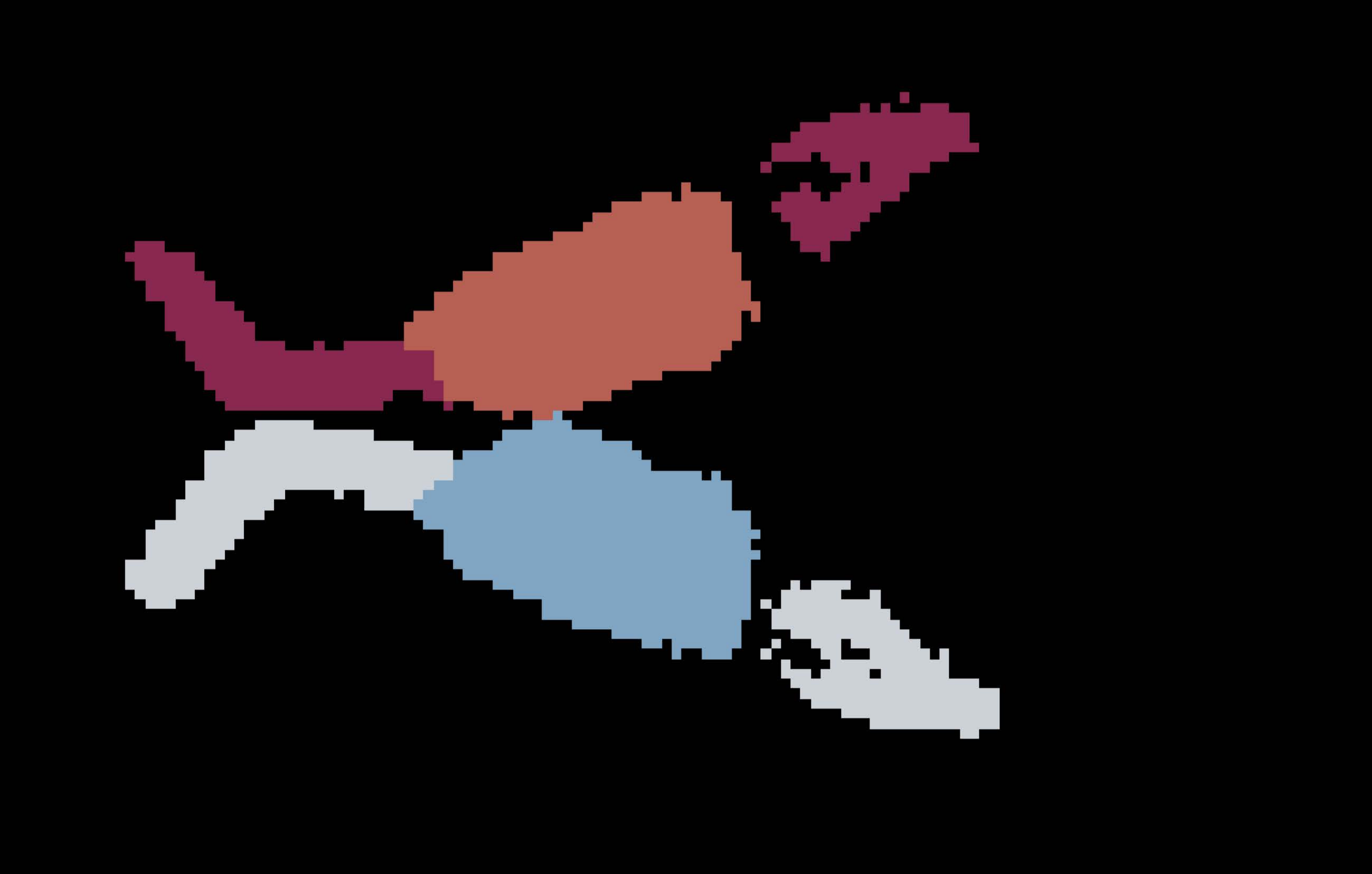}
		&\includegraphics[width=0.194\textwidth]{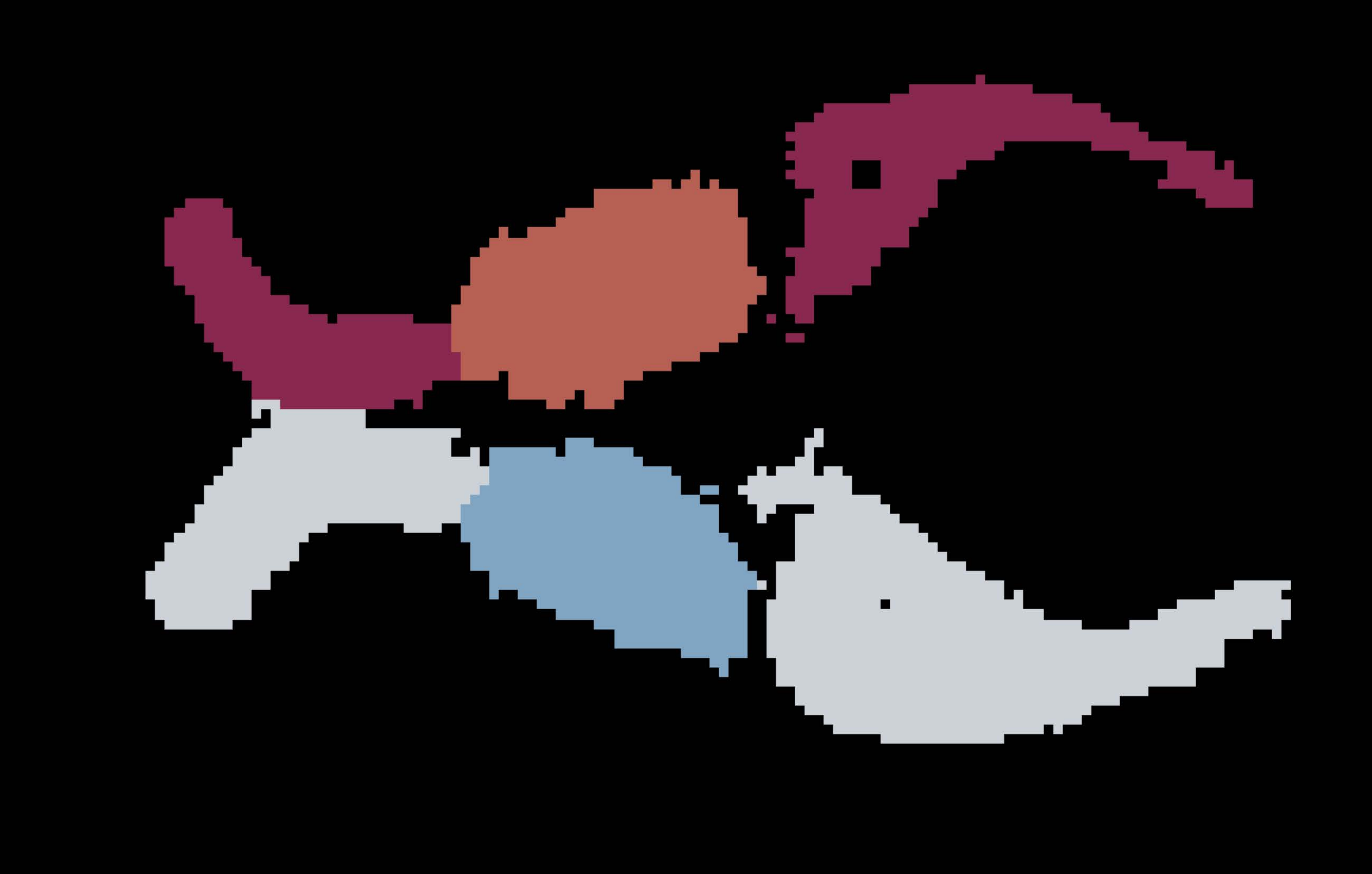}
		&\includegraphics[width=0.194\textwidth]{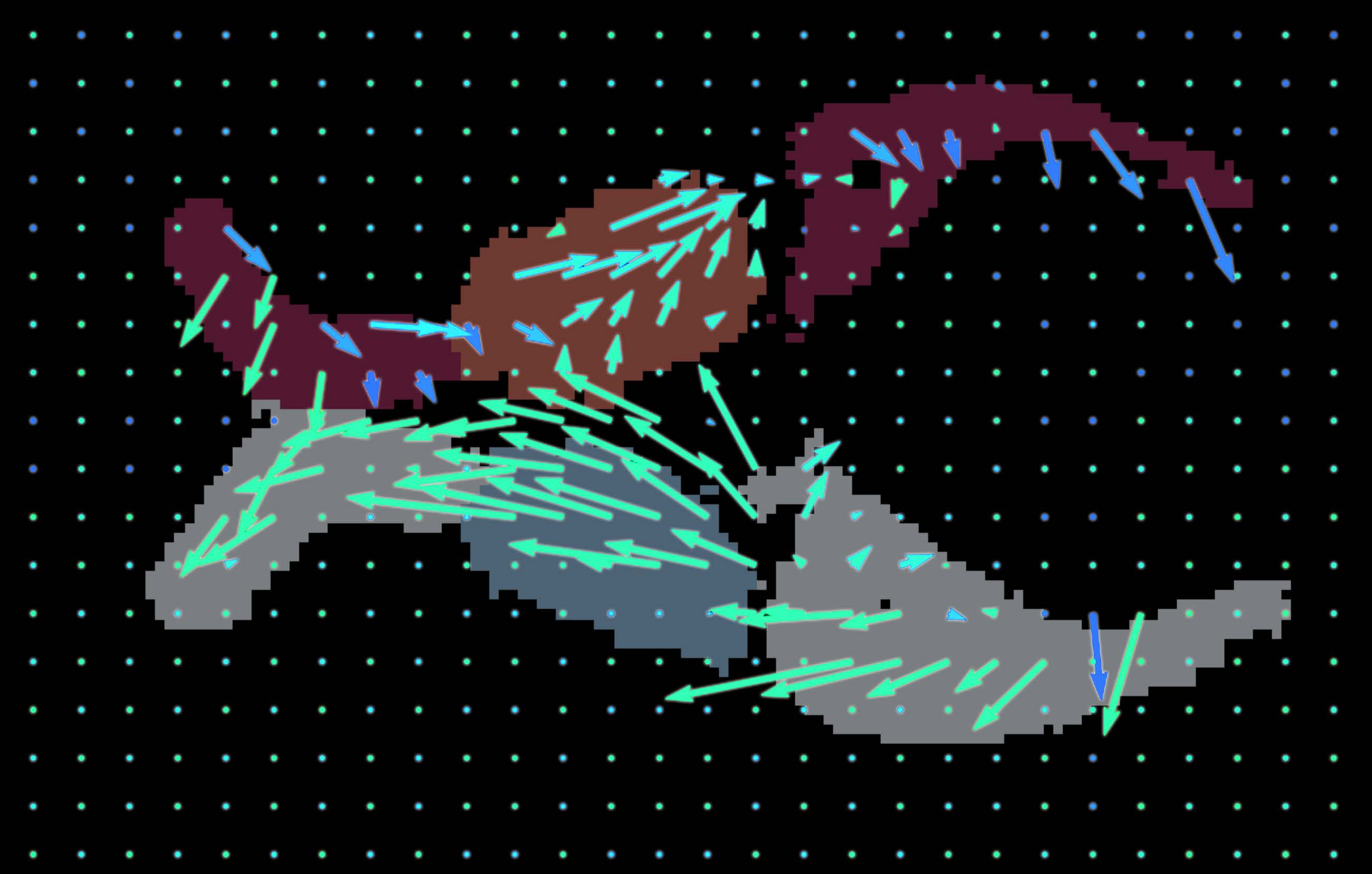}
		&\includegraphics[width=0.194\textwidth]{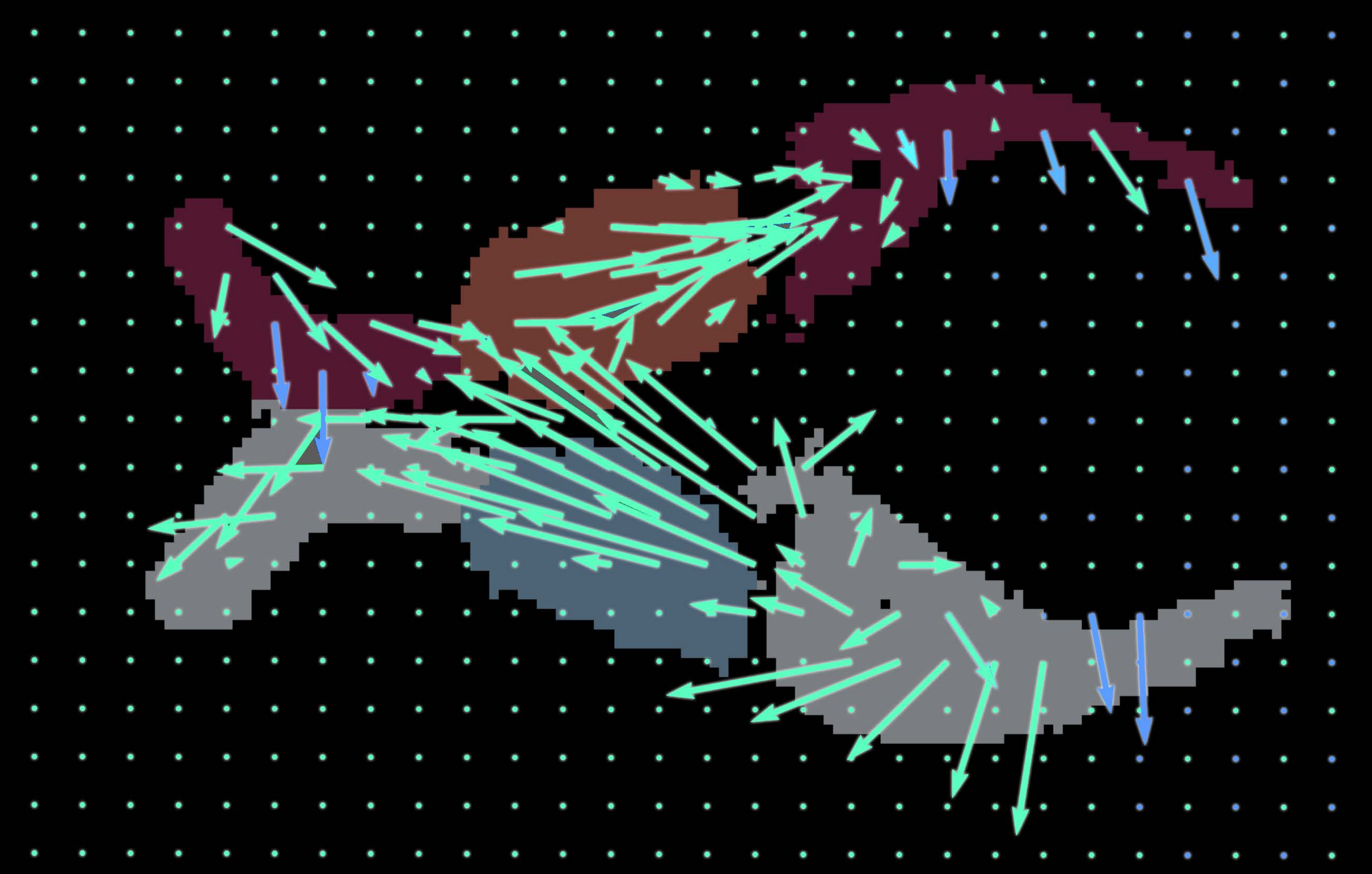}
		&\includegraphics[width=0.194\textwidth]{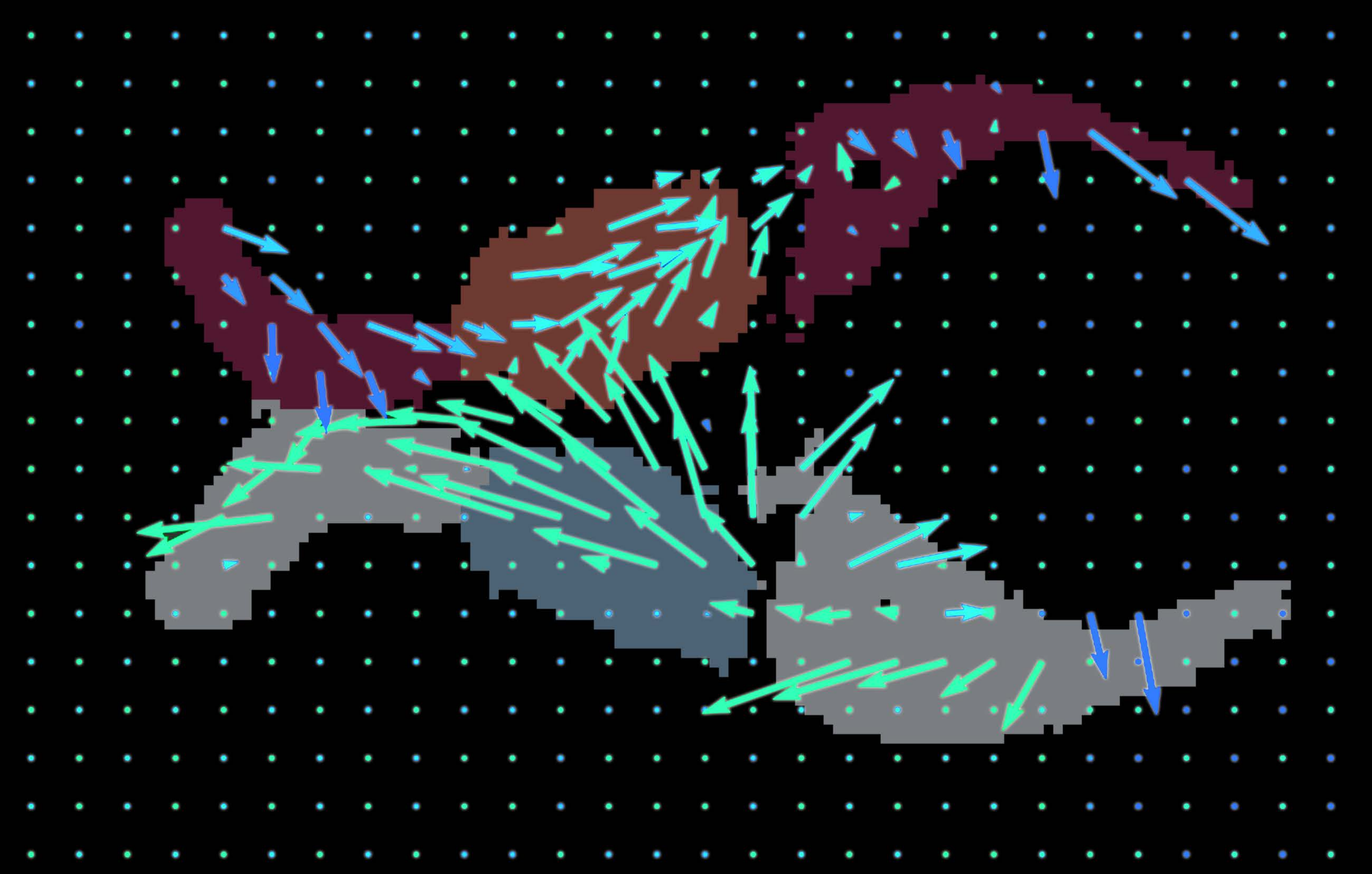}\\
		
		Target  & Source  & VM & VM-diff  & Ours   \\
		(1.0)~  & (0.596)~   & (0.850)~ &  (0.848)~  &  (\textbf{0.856})   \\
		
	\end{tabular}
	\caption{ Registered MR slices and segmented anatomical structures using VM, VM-diff, and our method. The Dice scores are parenthesized. } 
	\label{fig:arrow}
\end{figure*}

\begin{table*}[t]
	\centering
	\caption{Qualitative comparison results on brain MR registration tasks. The means and standard deviations of the number of occurred folds are listed. }
	\label{tab:compare_jac}
	\vspace{-0.5em}
	\begin{tabular}{|m{1.0cm}<{\centering}|m{1.8cm}<{\centering}| m{1.8cm}<{\centering} |m{1.8cm}<{\centering} |m{1.8cm}<{\centering} |m{1.8cm}<{\centering} |m{1.8cm}<{\centering}|}
		\hline
		Methods    & Elastix     & NiftyReg    &  SyN    & VM    & VM-diff   & Ours \\
		\hline
		OASIS  & \textbf{0}    & 416.2 $\pm$ 416.0  & 29094 $\pm$ 8772   & 32029 $\pm$ 3498 &  35.7 $\pm$ 13.3  & 6.2 $\pm$ 4.6  \\
		ABIDE  & \textbf{0}    & 11.4 $\pm$ 13.2  & 27288 $\pm$ 3411   & 28861 $\pm$ 1616 &  25.4 $\pm$ 13.1  & 1.0 $\pm$ 0.9 \\
		ADNI  & 307.5 $\pm$ 1068    & 572.2 $\pm$ 878.9  & 30737 $\pm$ 9537   & 33047 $\pm$ 4667 &  43.4 $\pm$ 33.1  & \textbf{5.3 $\pm$ 6.0} \\
		PPMI  & 2.0 $\pm$ 15.6    & 314.3 $\pm$ 353.6  & 25452 $\pm$ 6490   & 30192 $\pm$ 3375 &  29.7 $\pm$ 24.8  & \textbf{0.1 $\pm$ 0.7}  \\
		HCP  & \textbf{0}    & 9576 $\pm$ 2287  & 28379 $\pm$ 4411   & 30716 $\pm$ 2086 &  3945 $\pm$ 3854  & \textbf{0}  \\
		\hline
	\end{tabular}
	
\end{table*}

\subsection{Ablation Analysis}
We investigate the role of different propagation components in our model, including feature extraction, context information, multi-scale degree as well as the architecture of the matching network and regularization network. 
We also discuss the benefits of the bilevel self-tuned training. 

\textbf{Feature extraction network evaluation.}~
We substitute our Feature Extraction Network (FEN) with handcrafted image pyramids as the case of without FEN and compare the performance gap between these two cases. Tab.~\ref{tab:ablation-feature} lists the registration accuracy in terms of both the Dice score and NCC at the 2, 3, and 4-scale. As shown, the importance of this module is clear given the inferior quality of the models without FEN under different scales.
Moreover, as the table shows, the FEN may evidently increase the accuracy in terms of both evaluation metrics and helps to achieve a more robust alignment quality (lower standard deviation).
In principle, the number of the scale of the model shows the trade-off between accuracy and model size. On the whole, the 3-scale model outputs superior performance over the other scale.

\textbf{Model configurations evaluation.}~
Firstly, we illustrate the effect of network architecture by changing the number of convolutional layers of either Matching Network (MN) or Regularization Network (RN). Our deep framework uses a three-layer matching network and three-layer regularization network at each level. Tab.~\ref{tab:ablation-configurations} shows the results by four variants that use one layer ($-$) and six layers ($+$) and keeping the rest the same.
As shown, the larger-capacity matching network leads to better results, resembling the critical role of data matching term for accurate spatial correspondence. 
Further, we consider the case where the regularization network is eliminated and only the matching network is engaged for the registration process to figure out the significance of regularization on the deformation field. From Tab.~\ref{tab:ablation-block} we can see that the participation of regularization can ideally exploit the context information to refine the predicted field and obviously promote the registration performance. Besides, we observe in experiments that a deeper network architecture could more easily get stuck at overfitting, which can be solved by employing more training data. Therefore, we adopt the 3-scale model to perform the following experiments.

\begin{figure*}[t]
	\centering
	\begin{tabular}{@{\extracolsep{0.23em}}c@{\extracolsep{0.23em}}c@{\extracolsep{0.23em}}c@{\extracolsep{0.23em}}c@{\extracolsep{0.23em}}c@{\extracolsep{0.23em}}c@{\extracolsep{0.23em}}c@{\extracolsep{0.23em}}c}
		
		\includegraphics[width=0.122\textwidth]{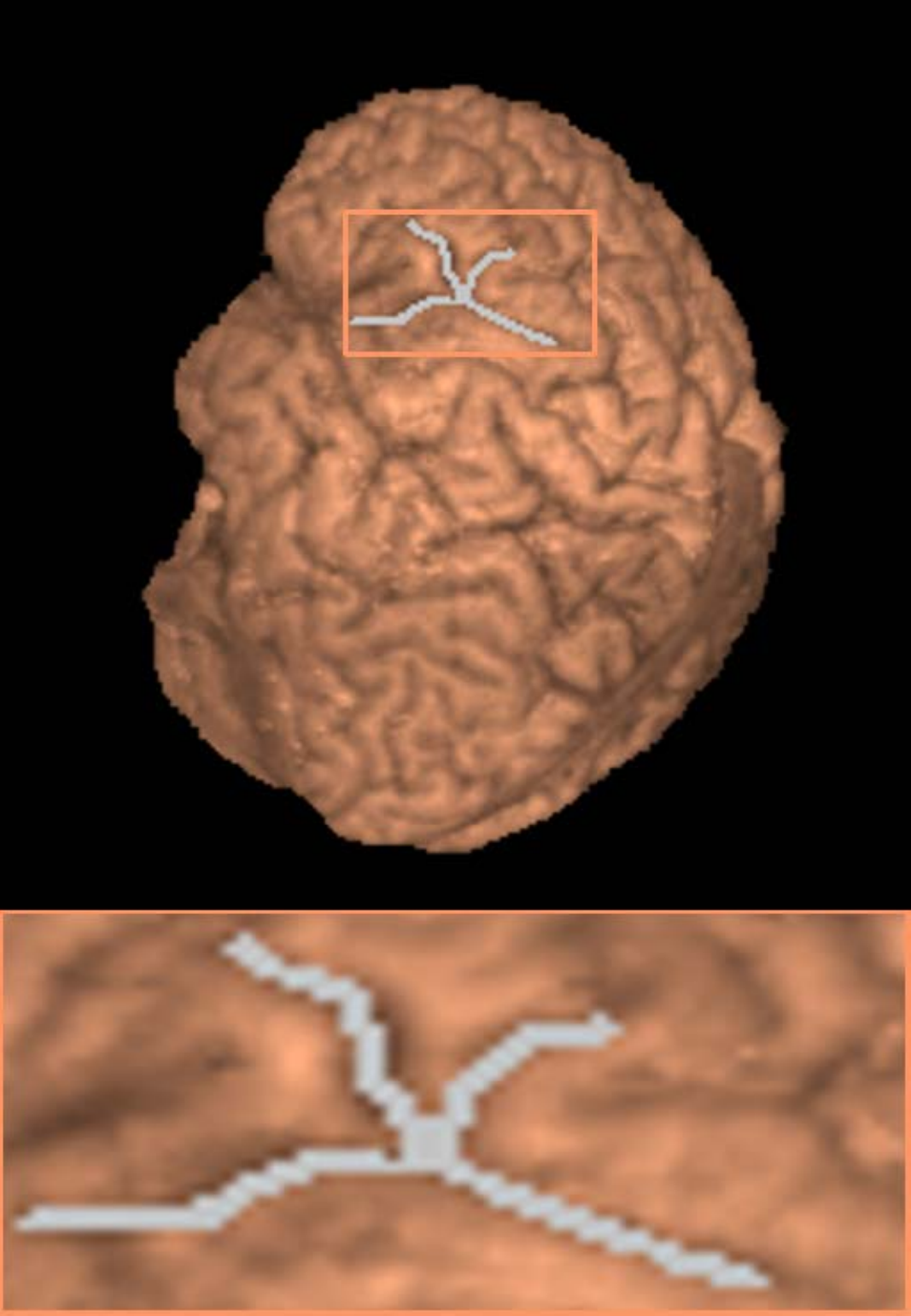}
		&\includegraphics[width=0.122\textwidth]{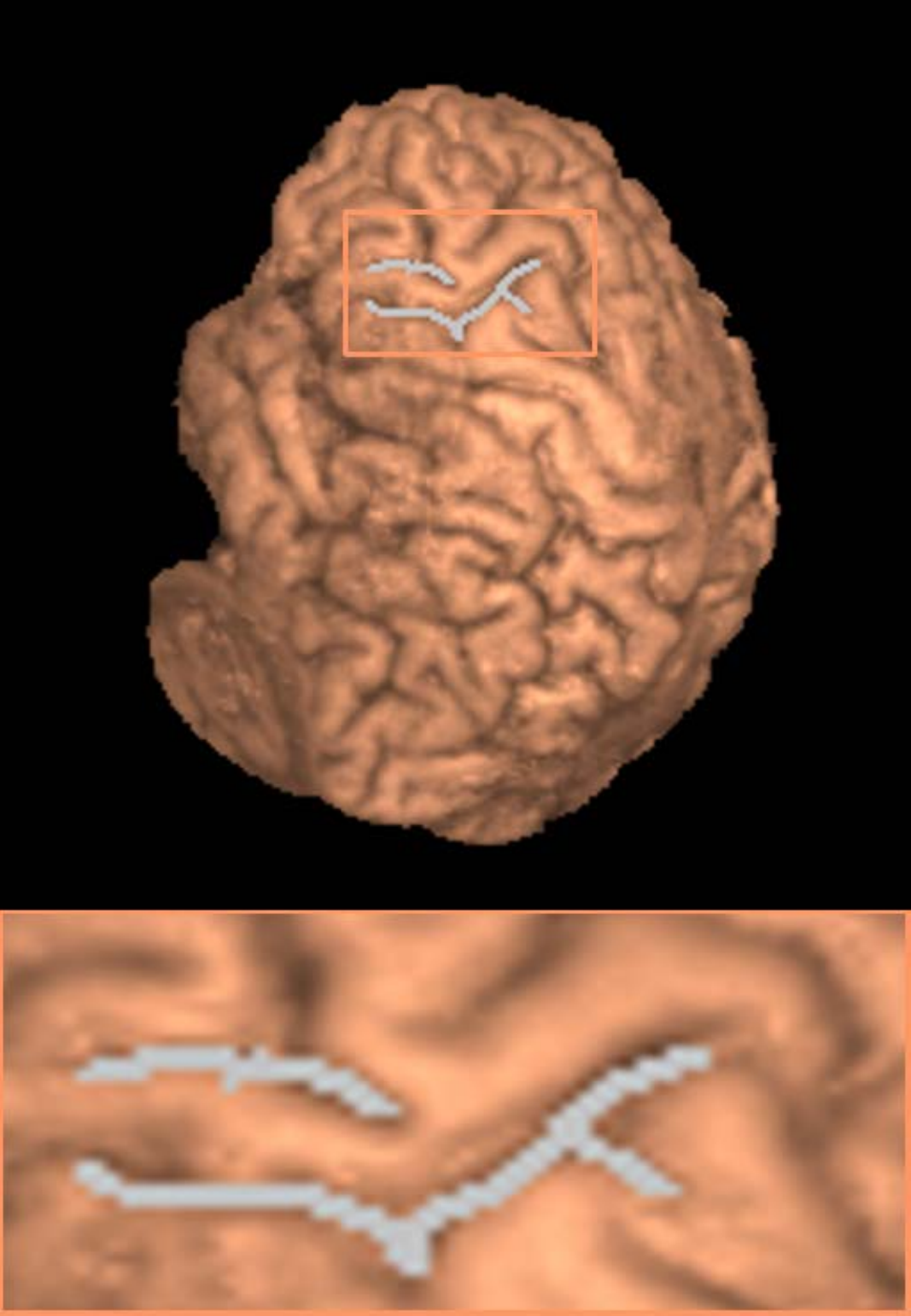}
		&\includegraphics[width=0.122\textwidth]{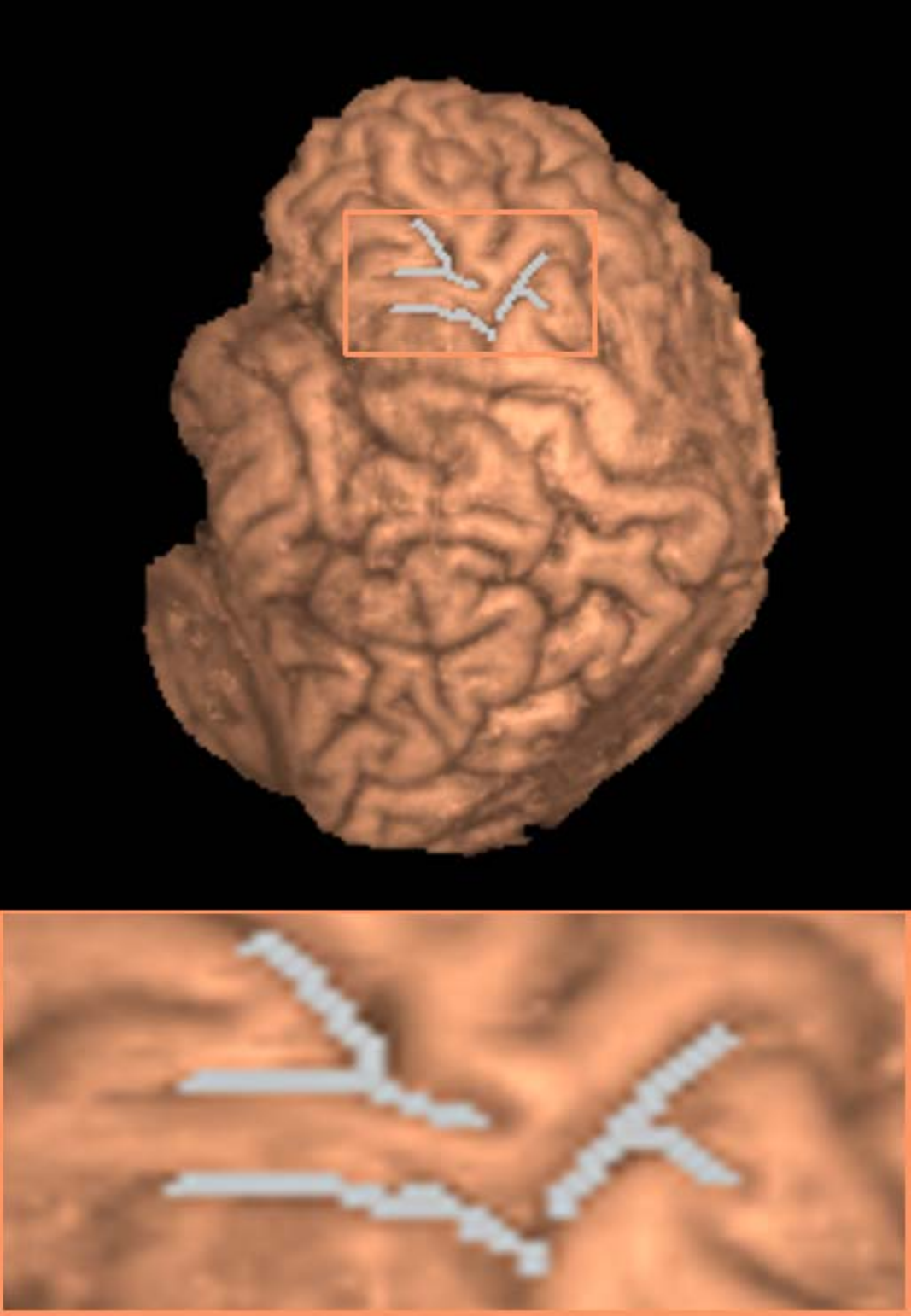}
		&\includegraphics[width=0.122\textwidth]{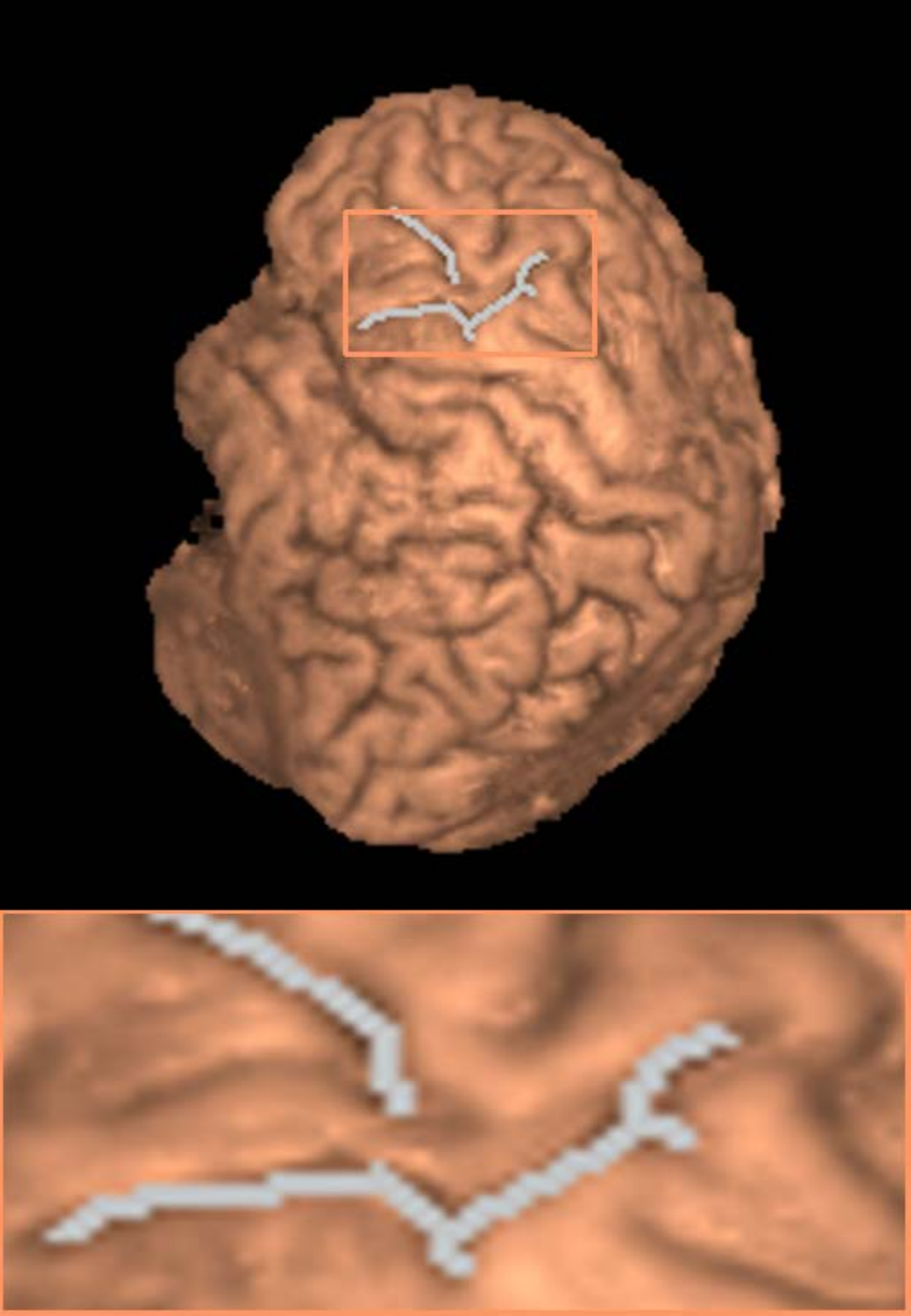}			
		&\includegraphics[width=0.122\textwidth]{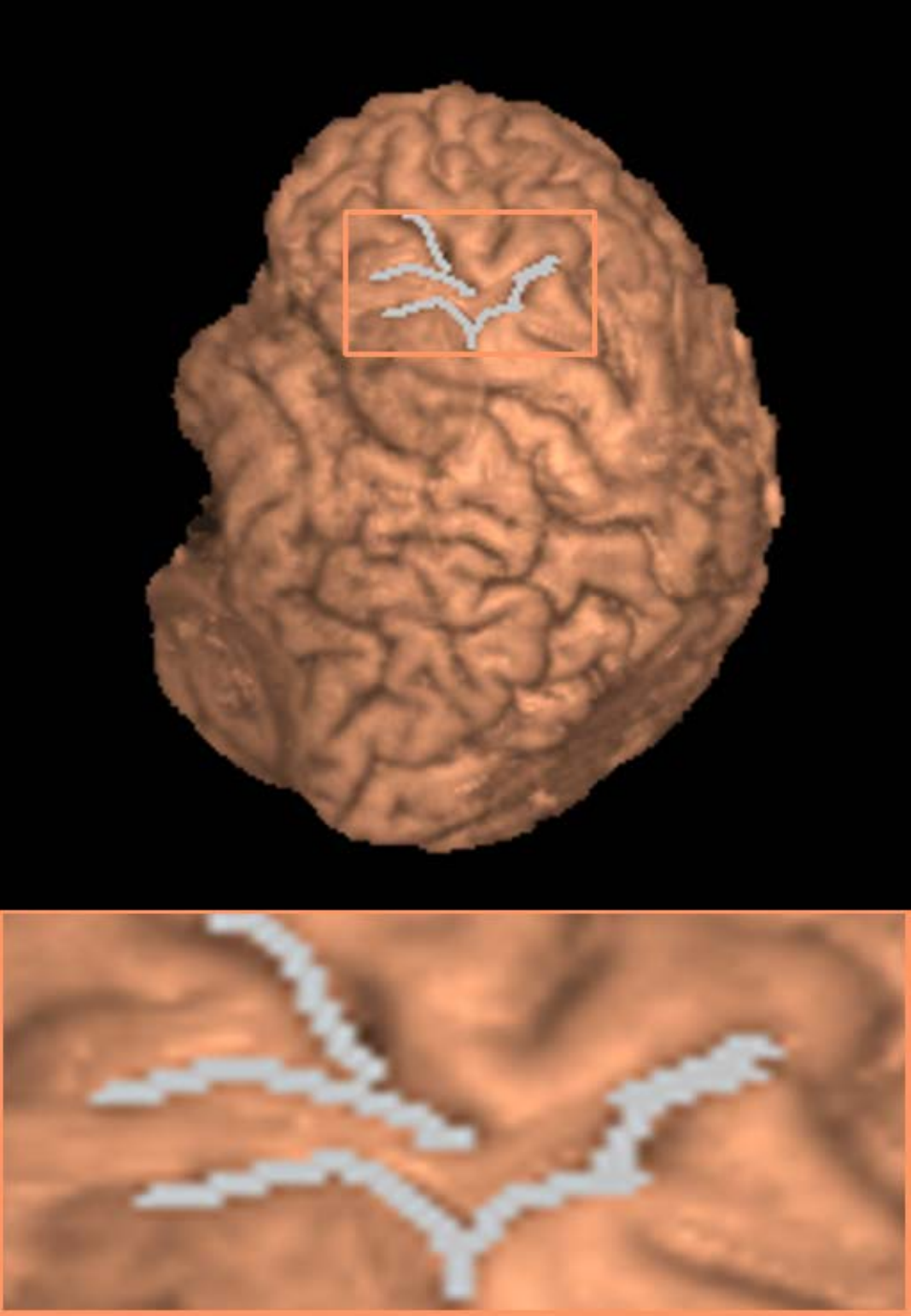}
		&\includegraphics[width=0.122\textwidth]{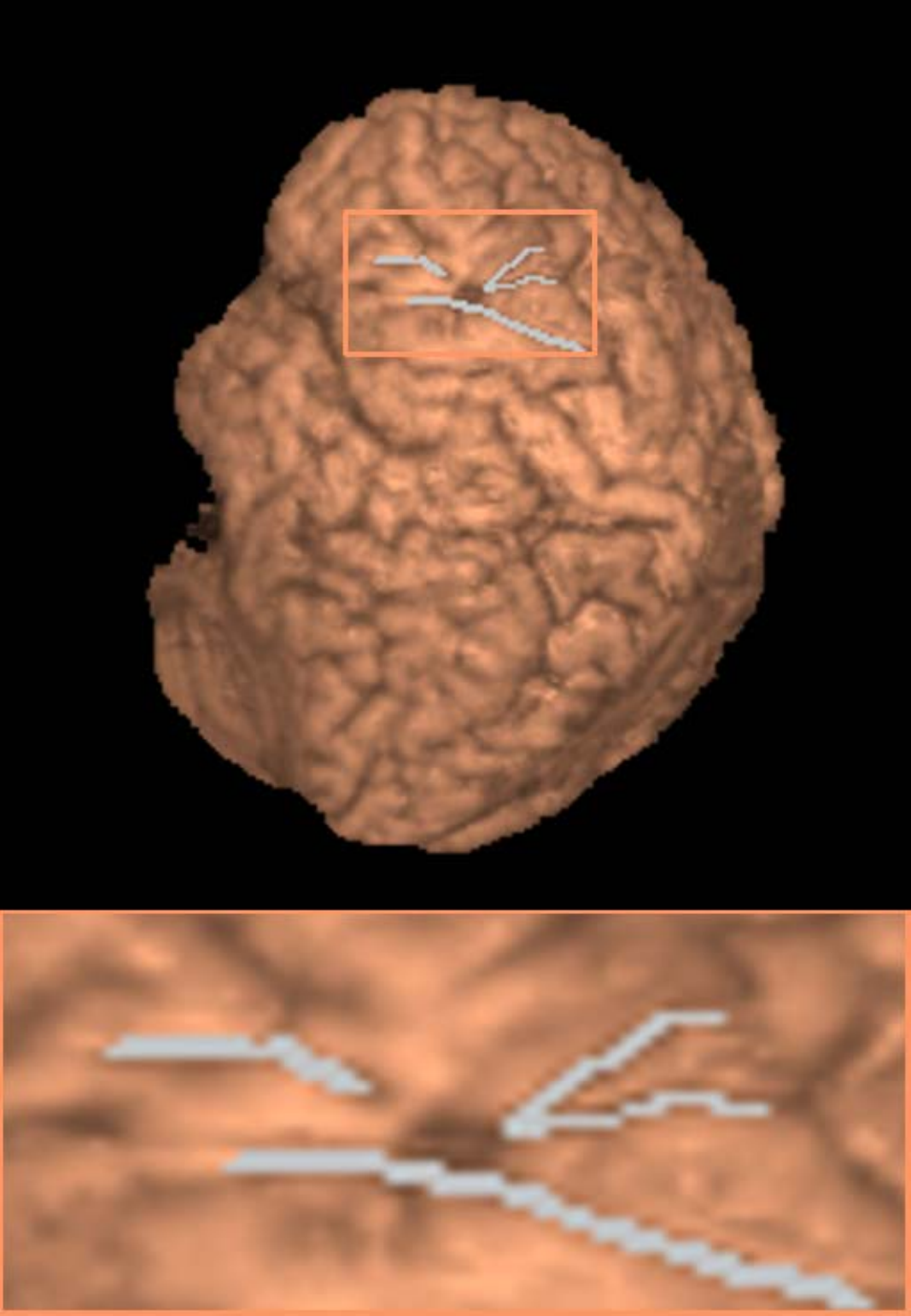}
		&\includegraphics[width=0.122\textwidth]{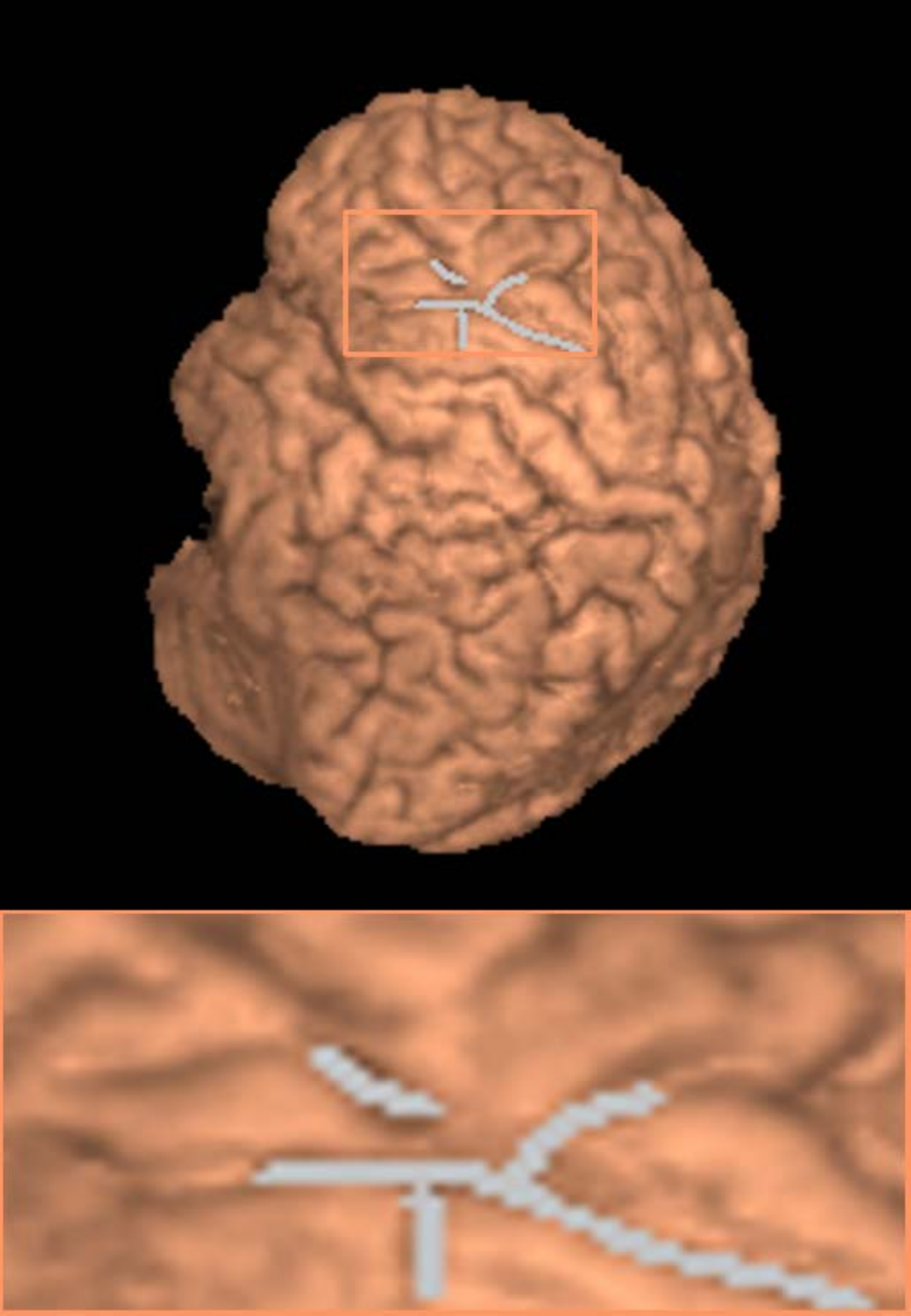}
		&\includegraphics[width=0.122\textwidth]{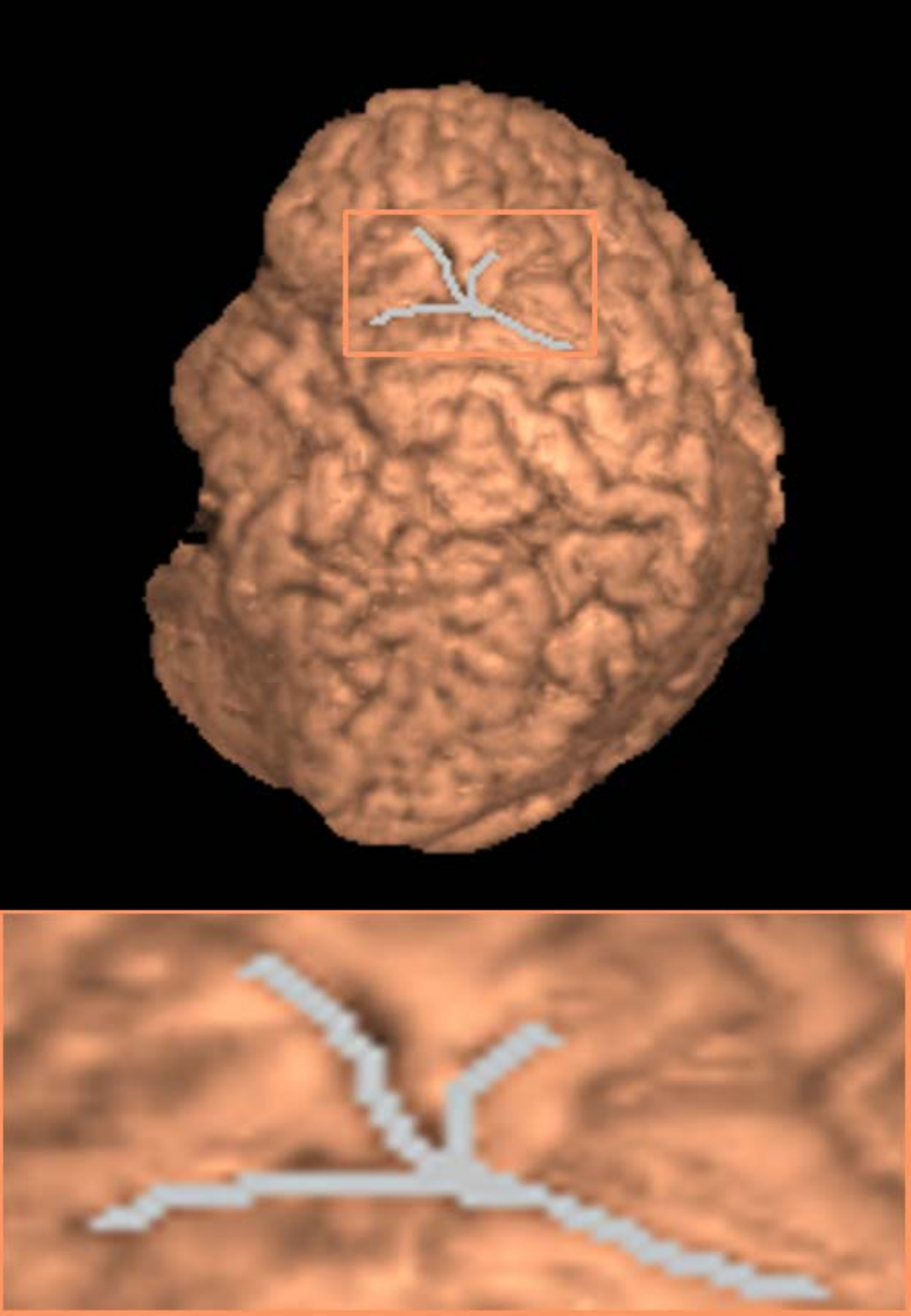}\\
		
		\includegraphics[width=0.122\textwidth]{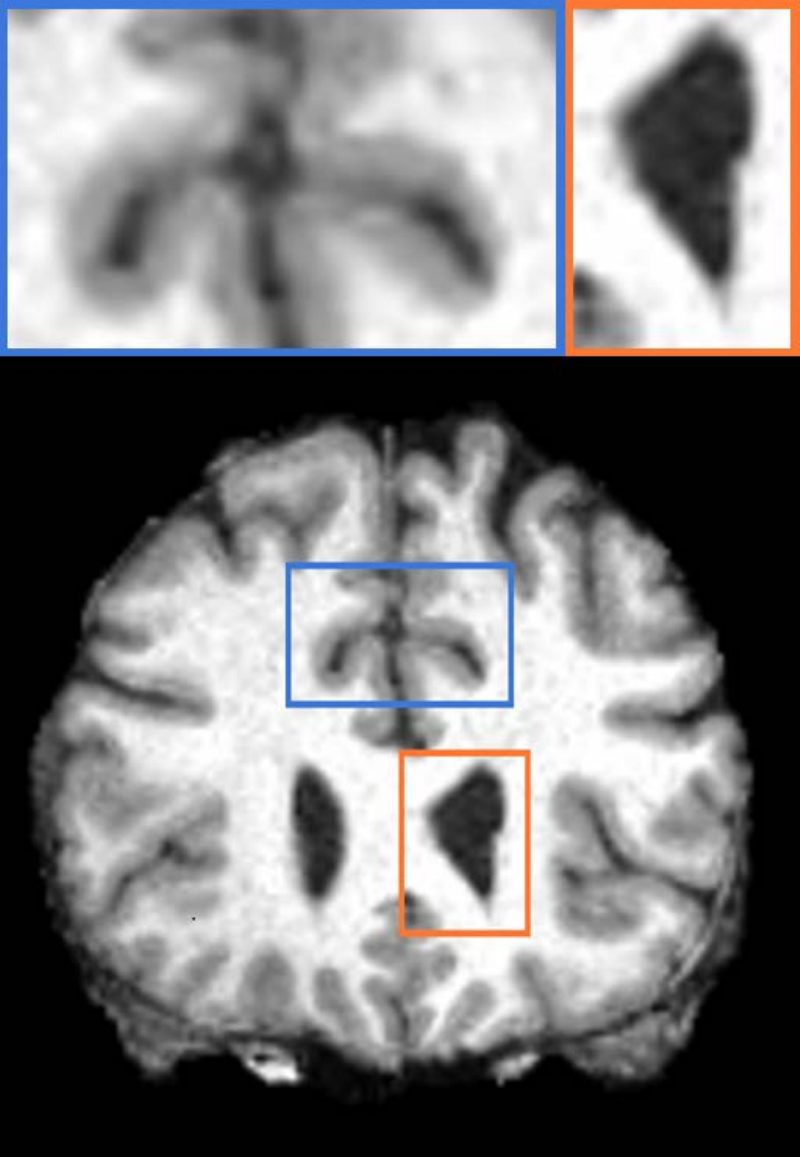}
		&\includegraphics[width=0.122\textwidth]{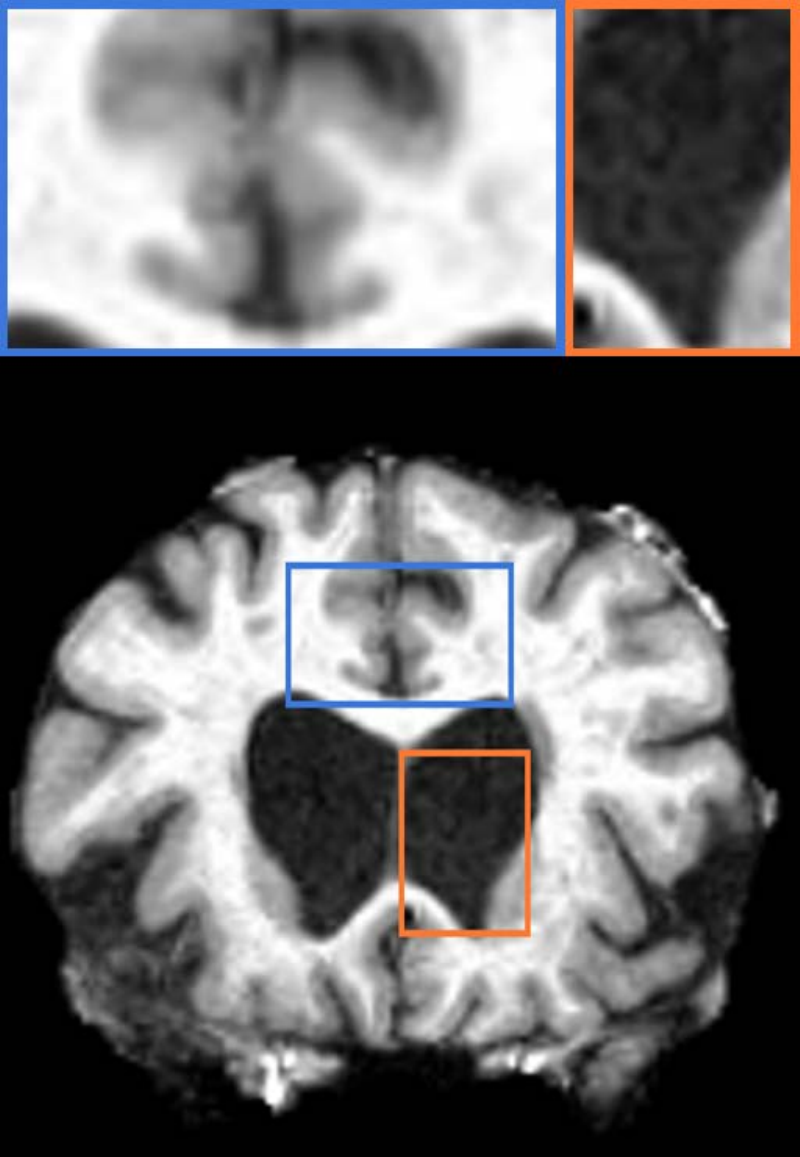}
		&\includegraphics[width=0.122\textwidth]{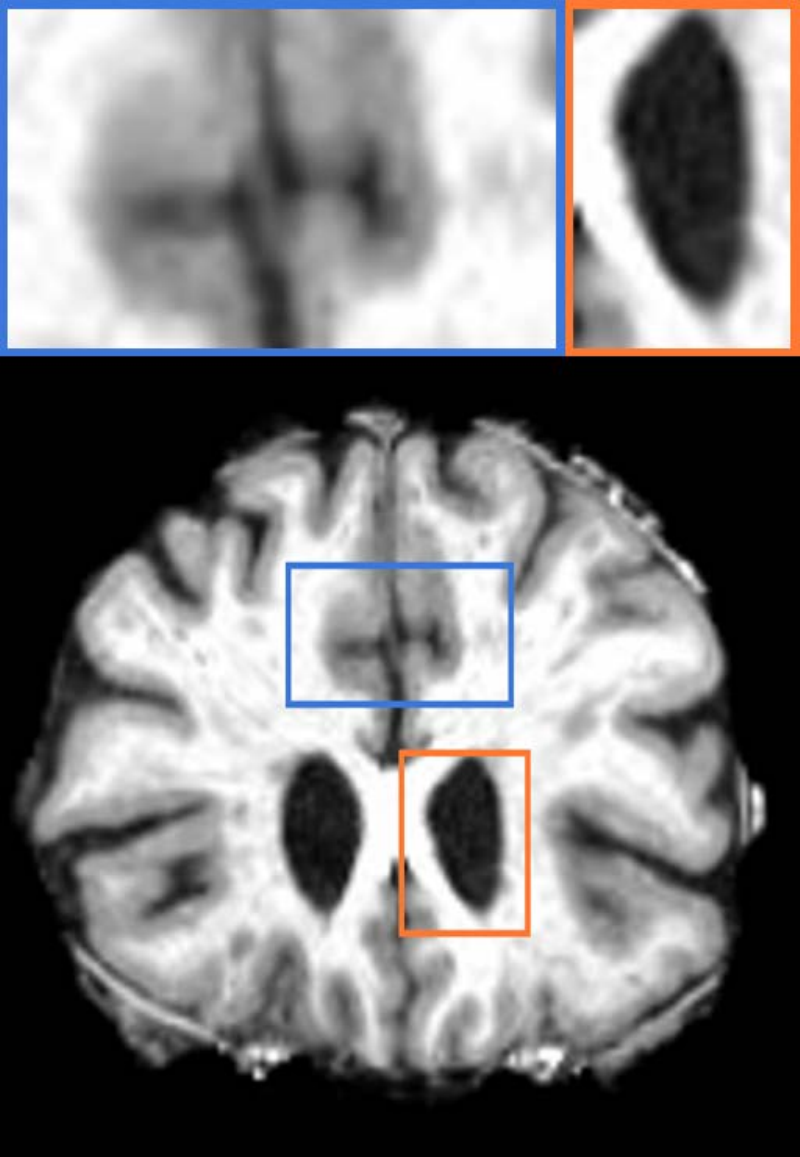}
		&\includegraphics[width=0.122\textwidth]{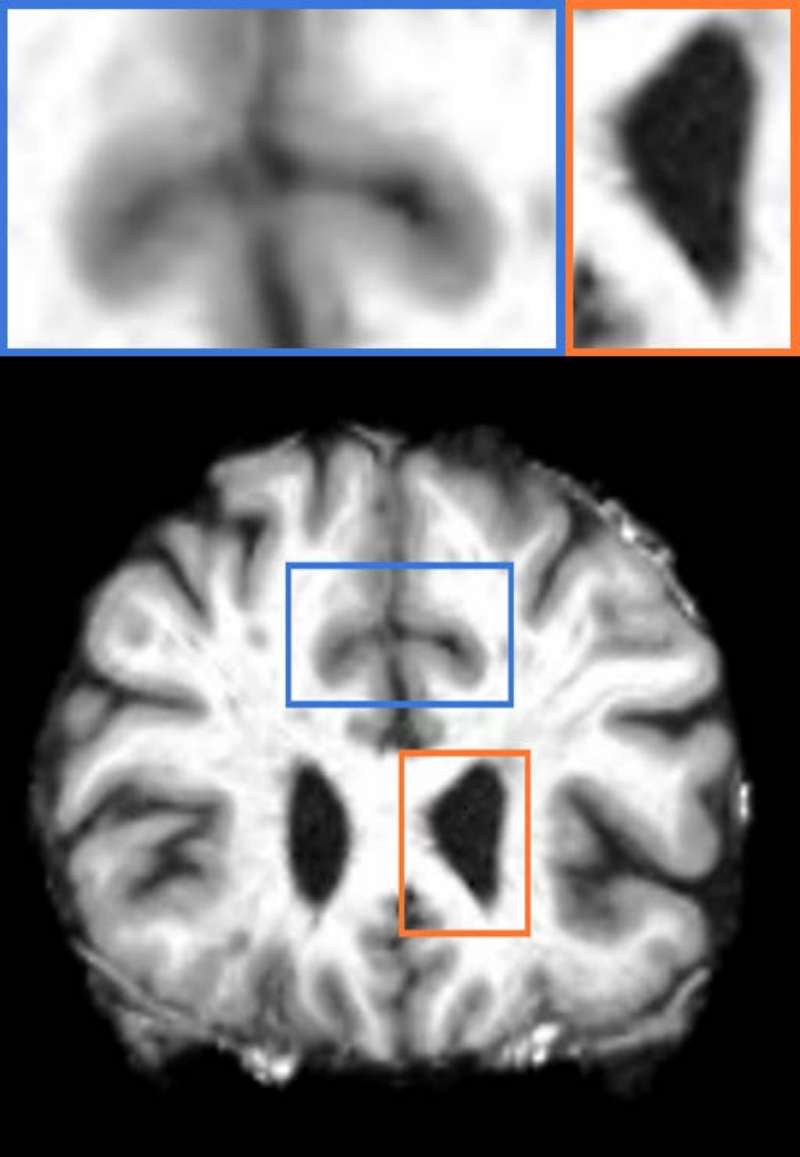}			
		&\includegraphics[width=0.122\textwidth]{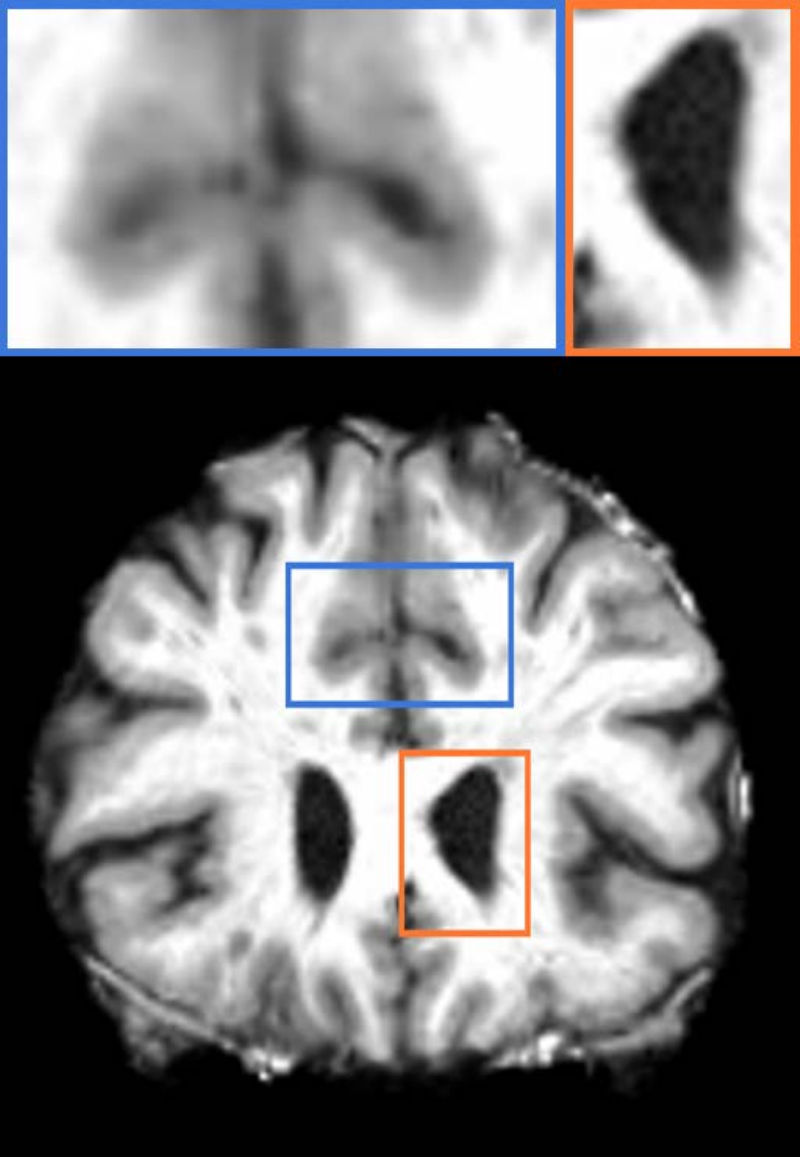}
		&\includegraphics[width=0.122\textwidth]{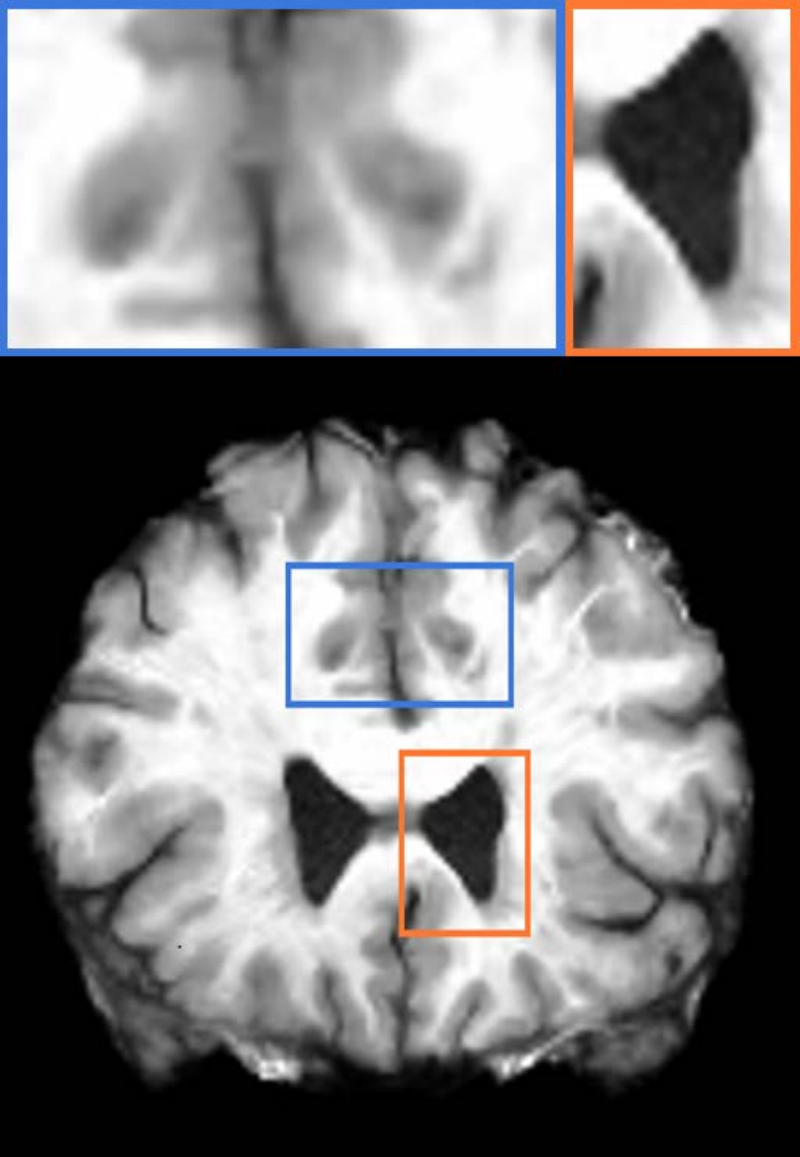}
		&\includegraphics[width=0.122\textwidth]{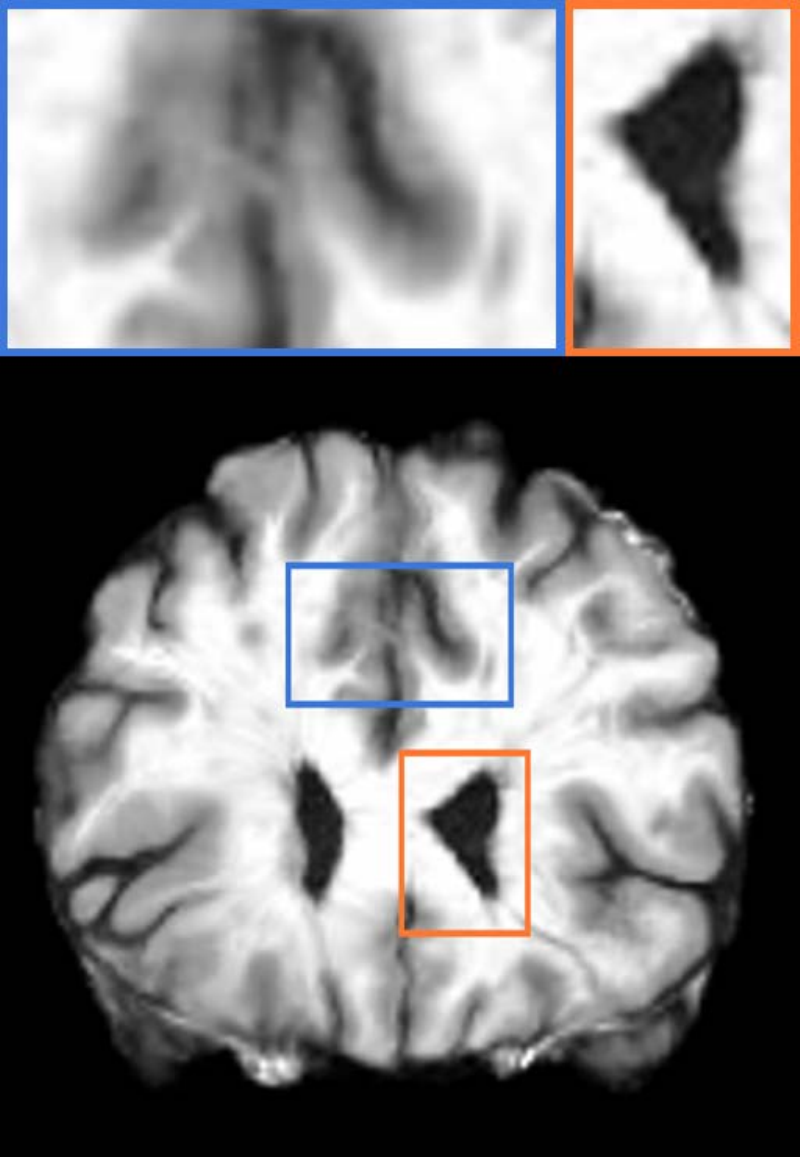}
		&\includegraphics[width=0.122\textwidth]{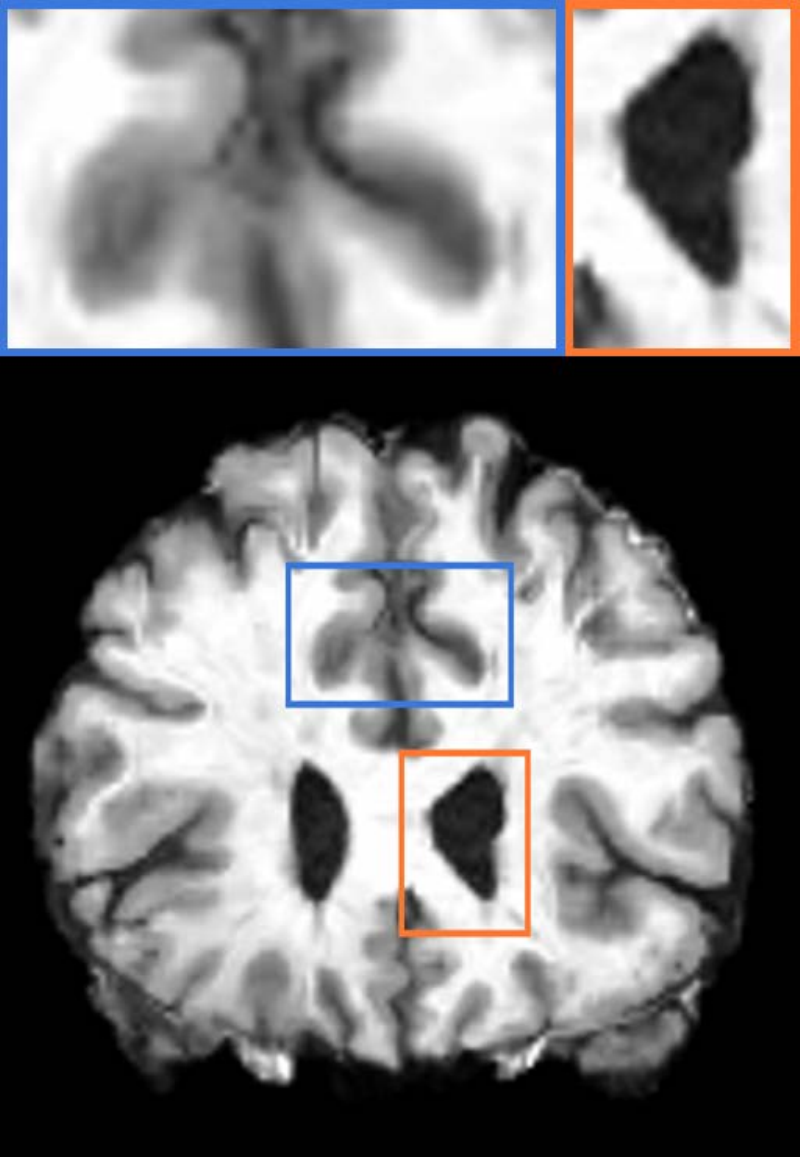}\\
		Target &  Source &  Elastix & SyN & NiftyReg & VM & VM-diff & Ours\\
		(1.0)~ & (0.442)~ & (0.658)~ & (0.729)~ & (0.729)~ & (0.662)~ & (0.720)~ & (\textbf{0.740})~\\
	\end{tabular}
	\caption{The first row demonstrates cortex visualization and zoomed-in overlaid sulci registered using different methods. The second row gives the MR slices and zoomed-in warped patches. Parenthesized values in the bottom give the Dice score. } 
	\label{fig:compare}
\end{figure*}

\begin{figure*}[t]
	\centering
	\begin{tabular}{c@{\extracolsep{0.25em}}c@{\extracolsep{0.25em}}c@{\extracolsep{0.25em}}c@{\extracolsep{0.25em}}c@{\extracolsep{0.25em}}c@{\extracolsep{0.25em}}c}
		
		\includegraphics[width=0.138\textwidth]{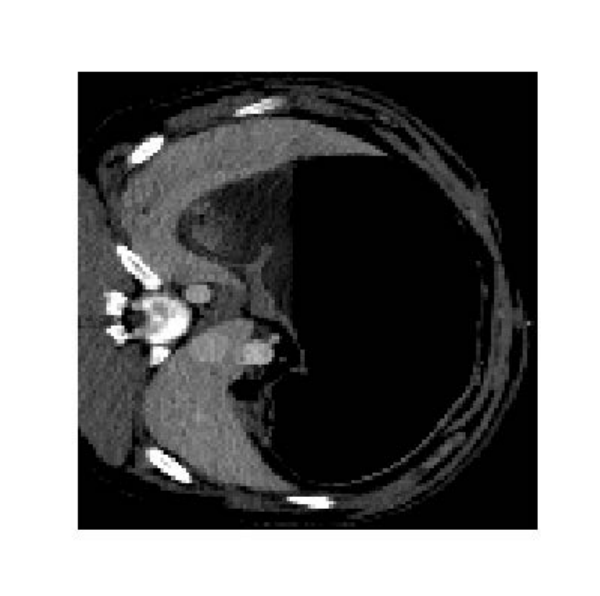}
		&\includegraphics[width=0.138\textwidth]{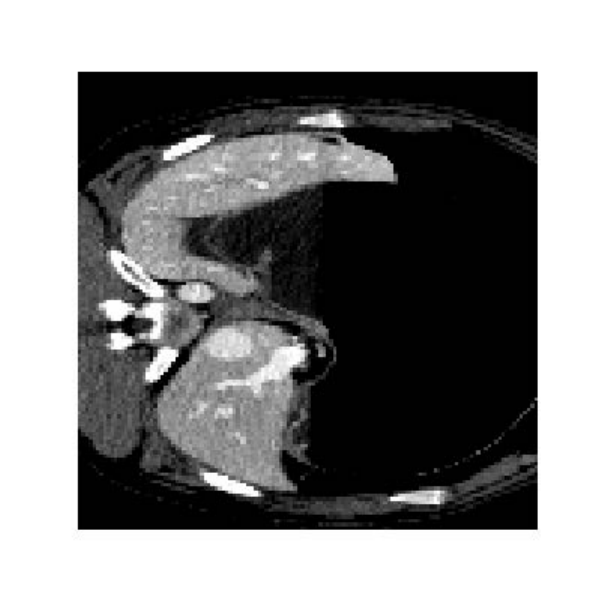}
		&\includegraphics[width=0.138\textwidth]{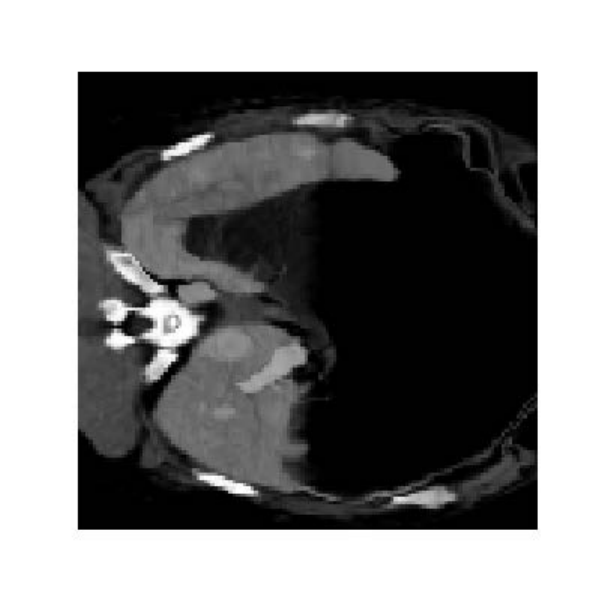}
		&\includegraphics[width=0.138\textwidth]{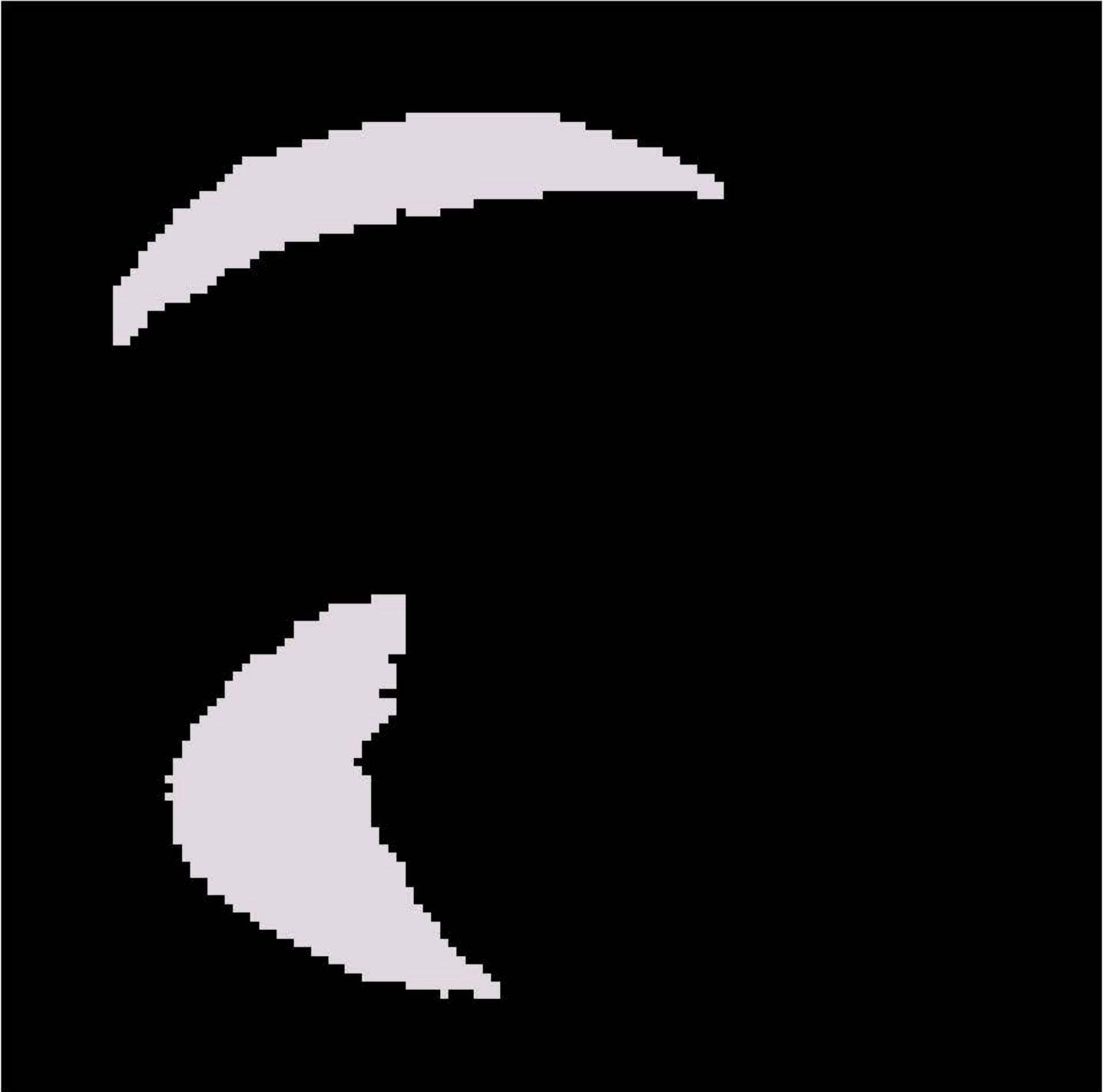}
		&\includegraphics[width=0.138\textwidth]{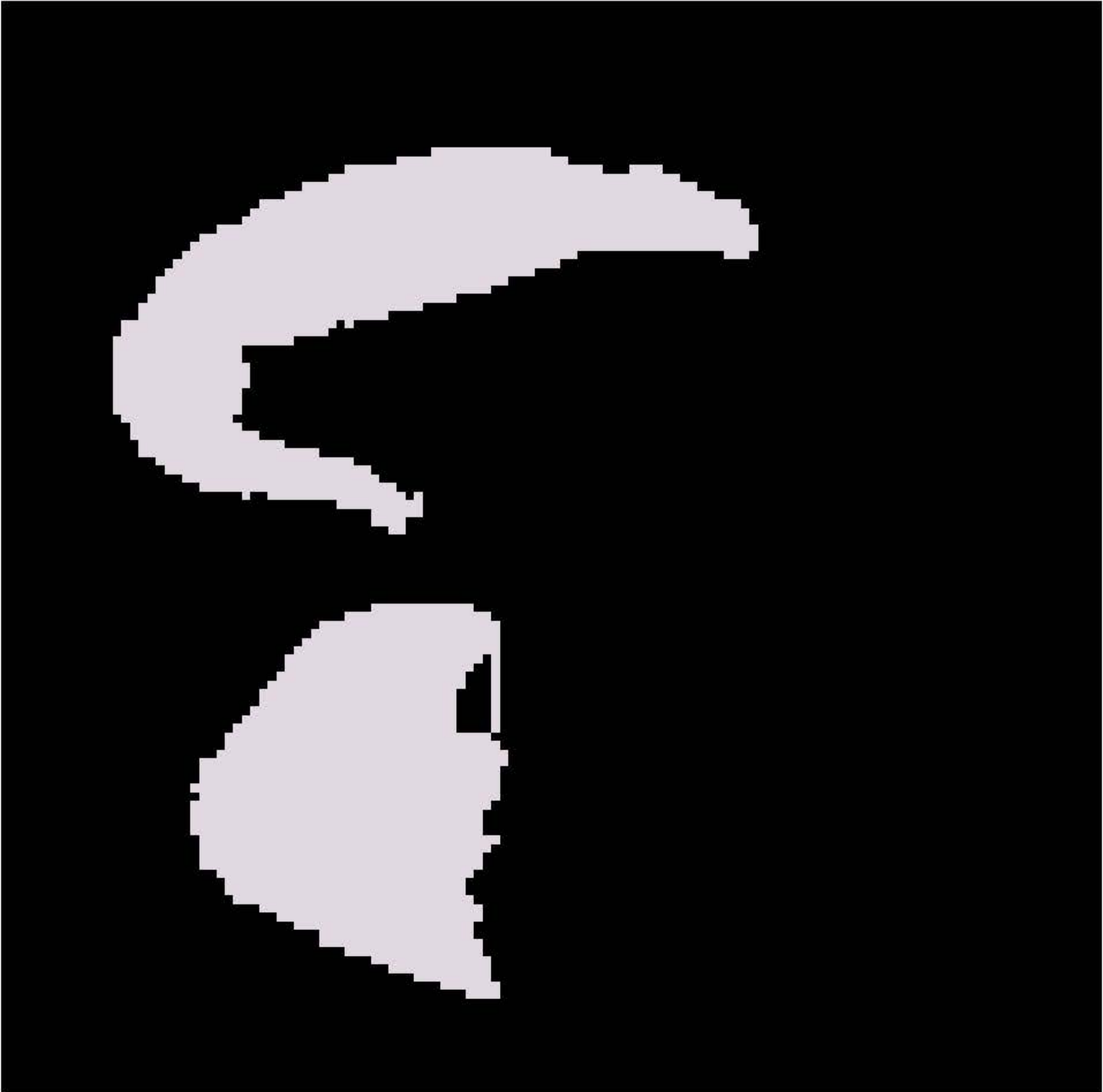}
		&\includegraphics[width=0.138\textwidth]{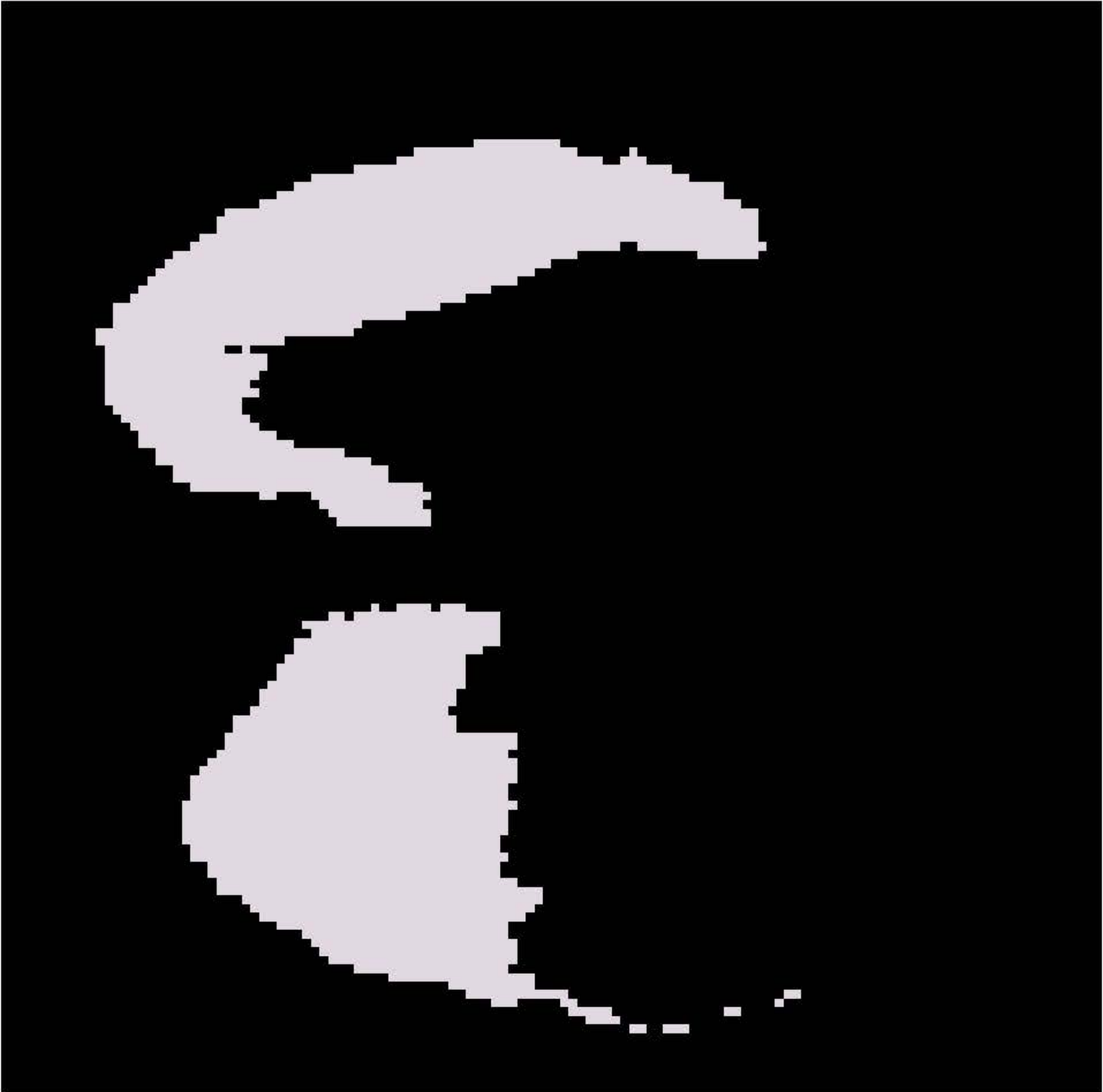}
		&\includegraphics[width=0.138\textwidth]{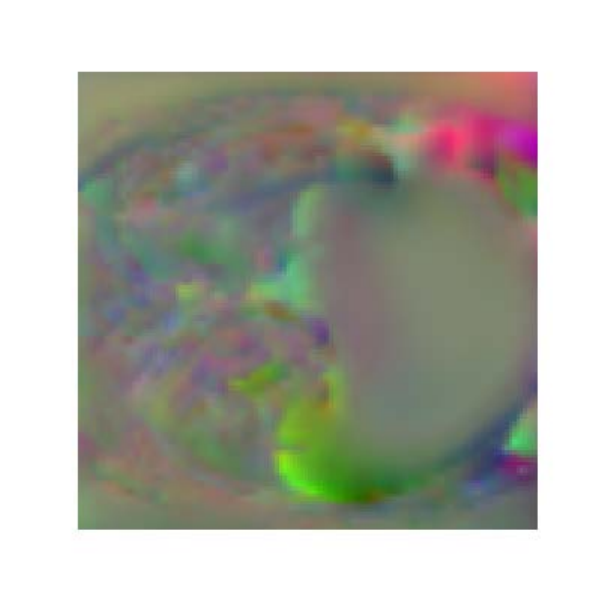}\\
		
		\includegraphics[width=0.138\textwidth]{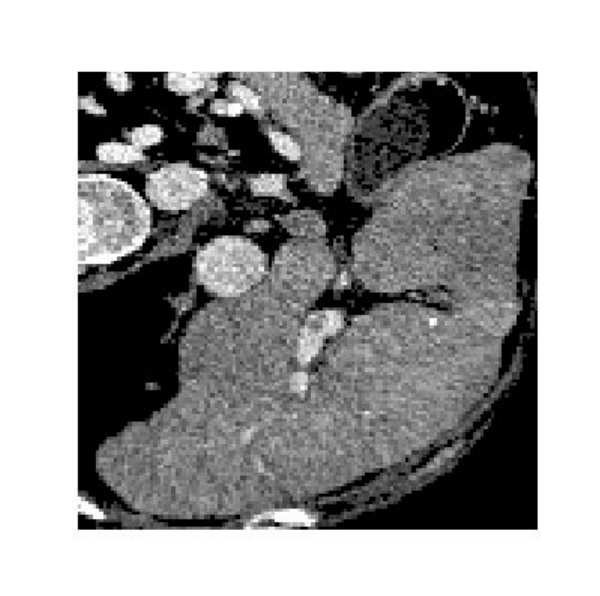}
		&\includegraphics[width=0.138\textwidth]{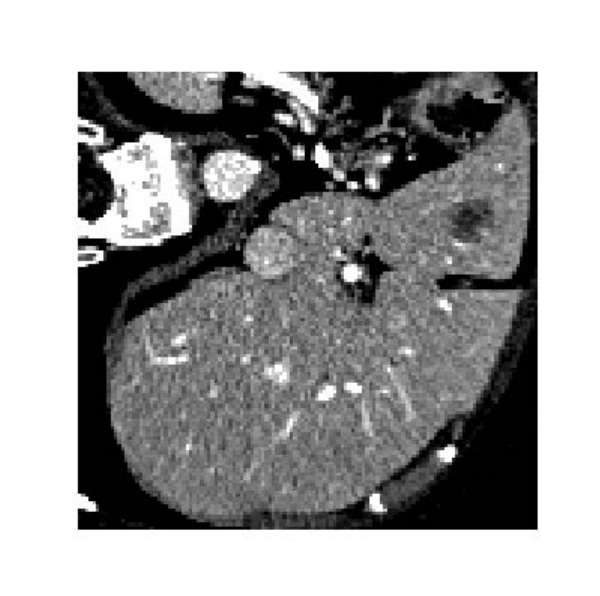}
		&\includegraphics[width=0.138\textwidth]{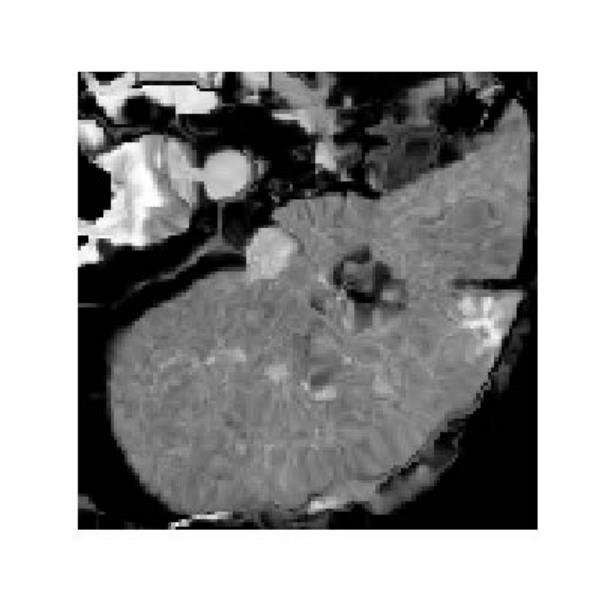}
		&\includegraphics[width=0.138\textwidth]{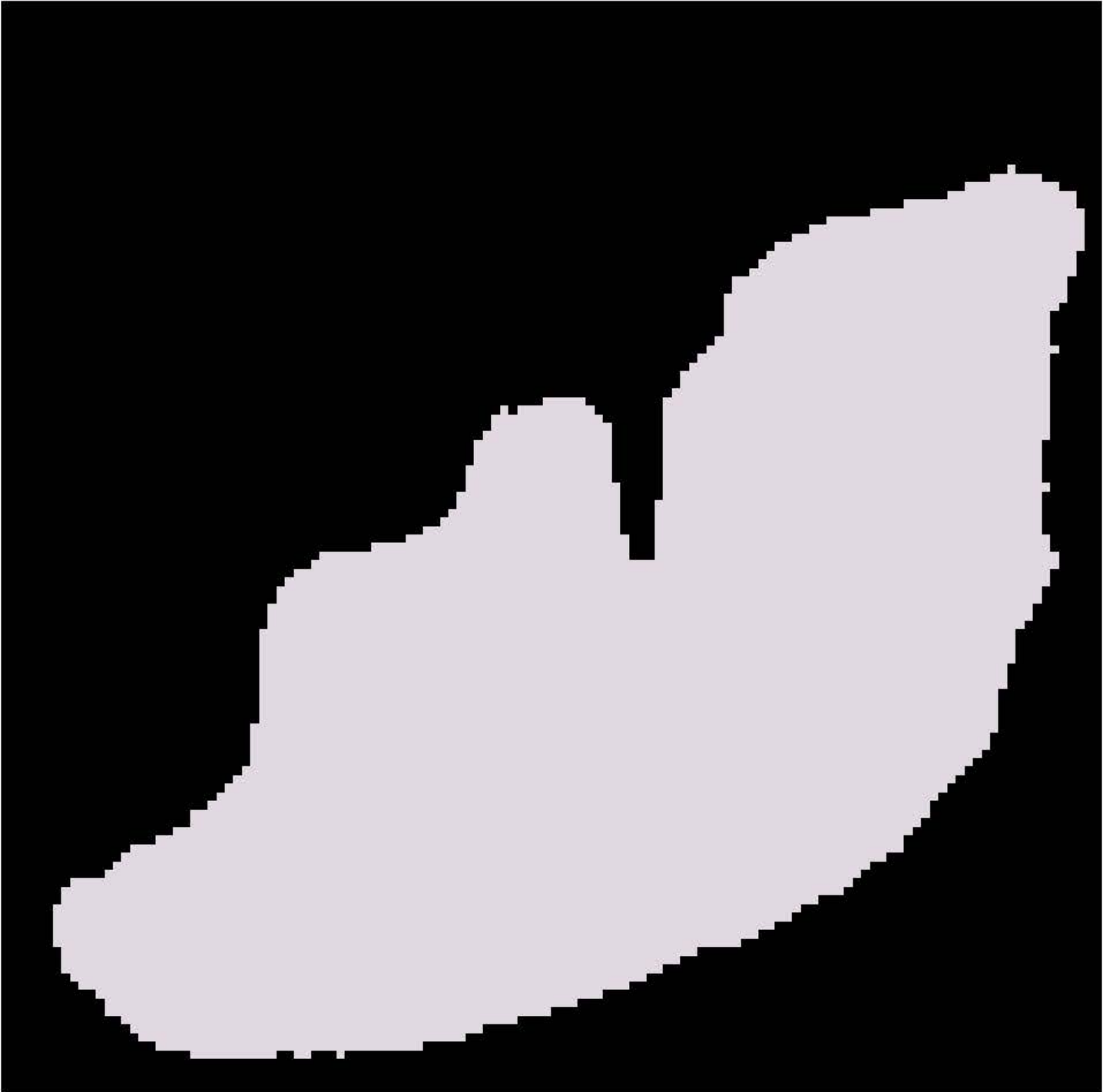}
		&\includegraphics[width=0.138\textwidth]{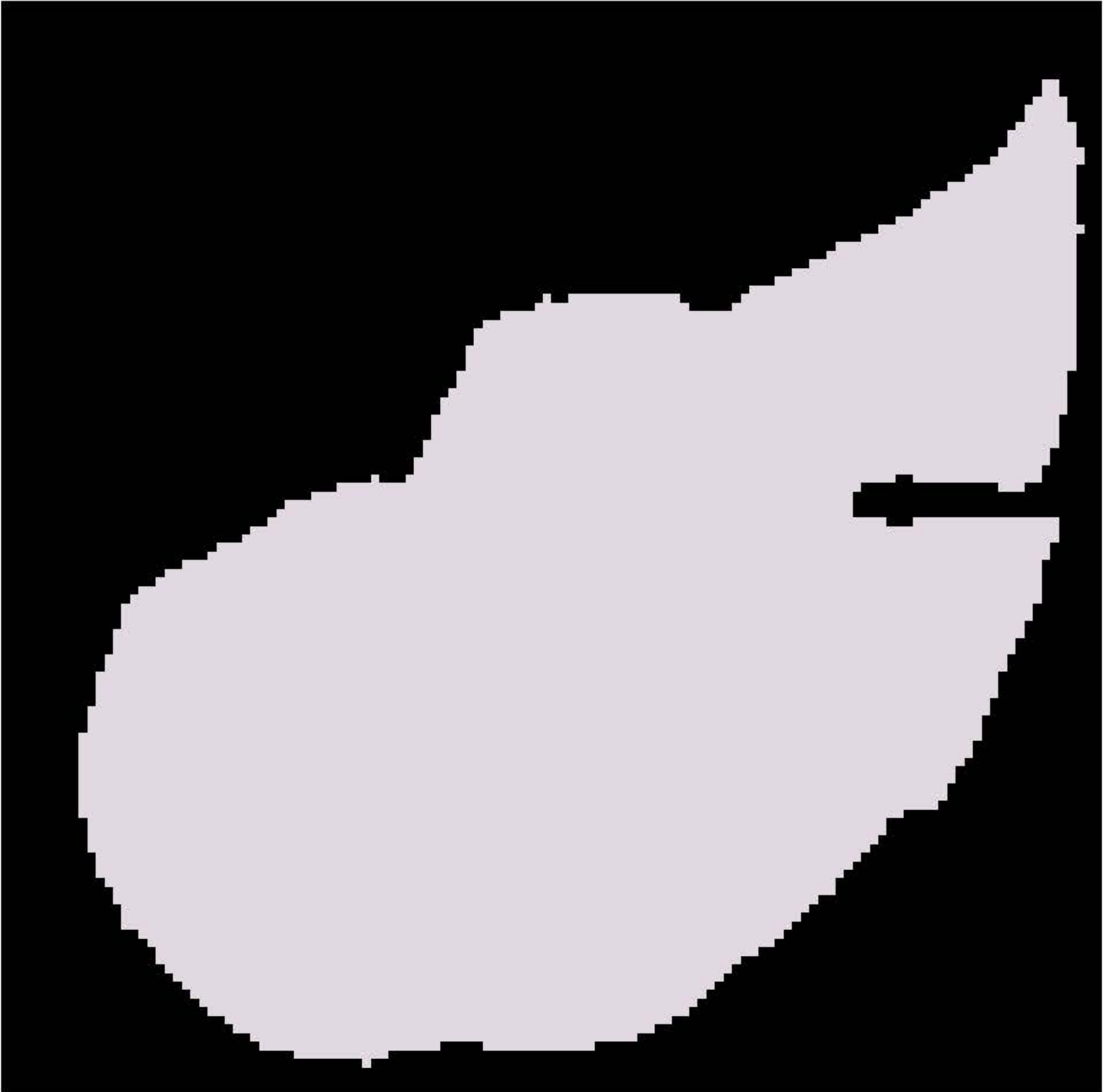}
		&\includegraphics[width=0.138\textwidth]{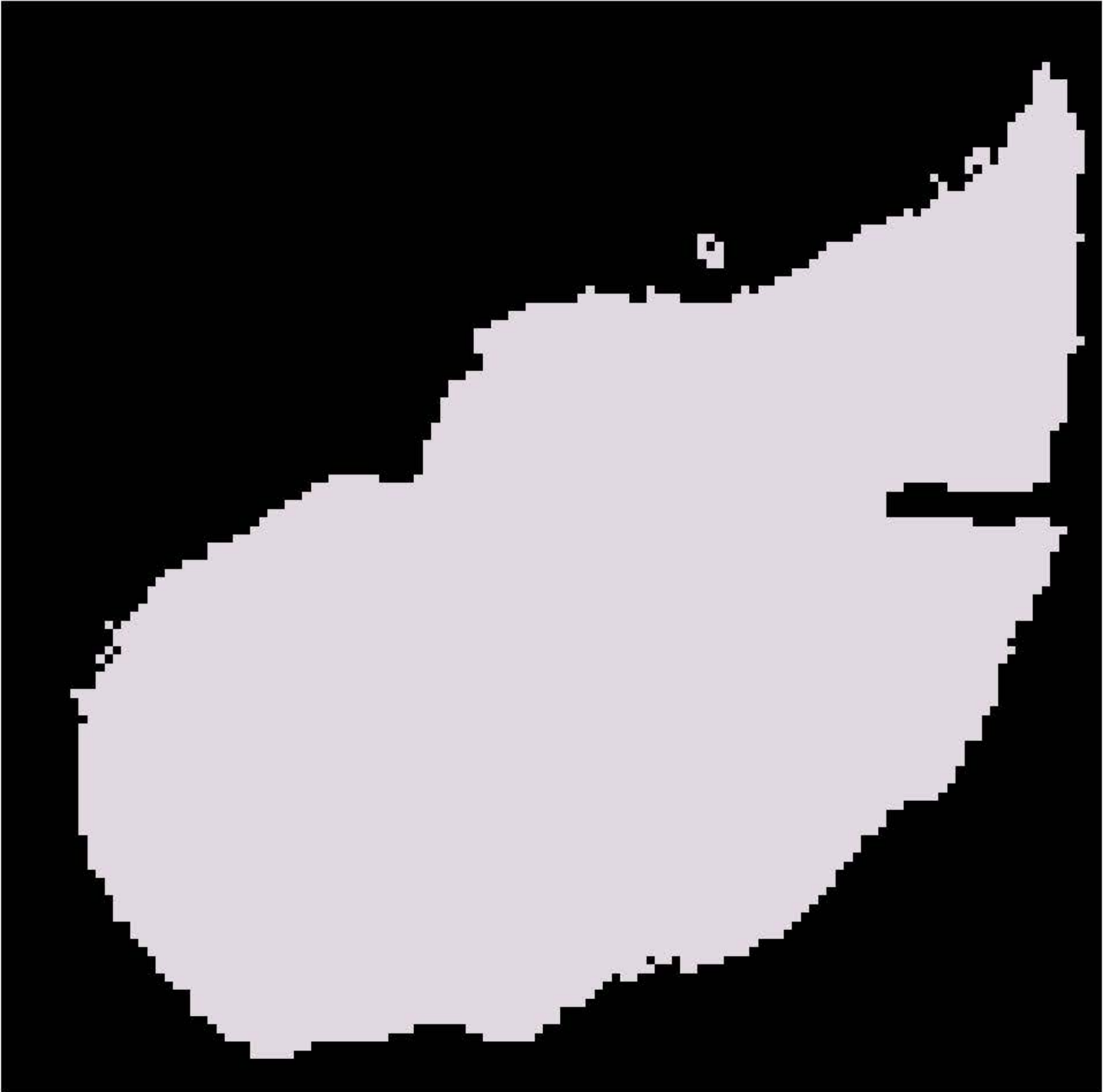}
		&\includegraphics[width=0.136\textwidth]{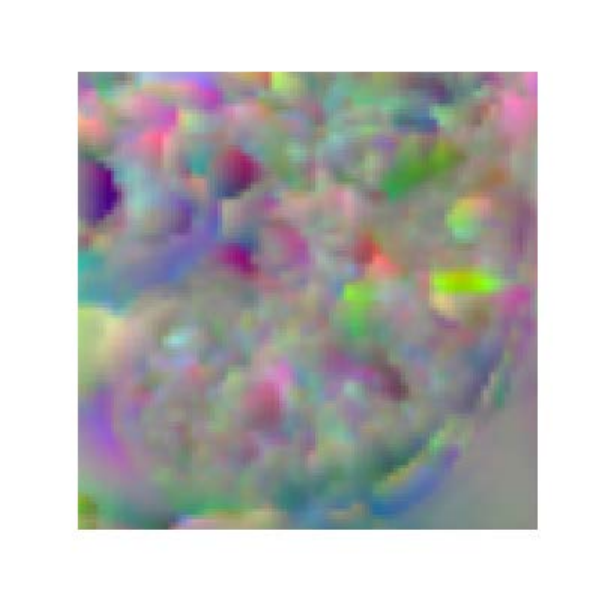}\\
		
		Source  & Target  &  Warped  & Source-label   &  Target-label & Warped-label & Flow field \\
		
	\end{tabular}
	\caption{ Two example registration cases of liver CT data, including target, source, registered images, and the corresponding labels and flow fields.  }
	\label{fig:example_ct} 
\end{figure*}

\textbf{Constraint module evaluation.}~
Except for the propagation networks, the advantage of constraint on the registration field is also explored. Fig.~\ref{fig:integration} illustrates the warped images and the corresponding flow grids generated by the model with and without integration operation. Apparently, the constraint can properly reduce unreasonable overlaps of the deformation field and preserve the topology of the warped volume, promising a smoother and more reliable registration.
To provide an intuitive comprehension of the effect of each component, Fig.~\ref{fig:multi-step} visualizes the deformation fields generated in each module during the propagation process, and the enlargement of corresponding local detail is attached on the bottom right. As the visualization shows, the data matching network firstly provides a primary estimation of field $\bm{{\bm{u}}^{1}}$, whereafter the regularization network refines the field to make $\bm{{\bm{v}}^{1}}$ smoother. Further, the integration operation ideally reduces the unreasonable overlaps within the field and guarantees a diffeomorphic deformation $\bm{{\varphi}^{1}}$. Repeating this way, the propagated field can be guided towards the desirable results and achieves satisfying accuracy.

\begin{table*}[t]
	\centering
	\caption{ Qualitative comparison results on liver CT registration tasks. The mean and standard deviations of the Dice score are listed.  }
	\vspace{-0.5em}
	\begin{tabular}{|m{1.0cm}<{\centering}| m{1.8cm}<{\centering}|m{1.8cm}<{\centering}| m{1.8cm}<{\centering}| m{1.8cm}<{\centering}|m{1.8cm}<{\centering}| m{1.8cm}<{\centering}| m{1.8cm}<{\centering}|}
		\hline
		Methods  & Affine only         & Elastix        &  NiftyReg      &  SyN      & VM    & VM-diff    & Ours  \\
		\hline
		SLIVER   & 0.794 $\pm$ 0.042  & 0.910 $\pm$ 0.038  & \textbf{0.931 $\pm$ 0.031}  & 0.895 $\pm$ 0.037  & 0.883 $\pm$ 0.034 & 0.878 $\pm$ 0.042  & 0.910 $\pm$ 0.027   \\	
		LSPIG   & 0.727 $\pm$ 0.054  & 0.825 $\pm$ 0.059  & 0.821 $\pm$ 0.122  & 0.825 $\pm$ 0.059  & 0.715 $\pm$ 0.009 & 0.788 $\pm$ 0.099  & \textbf{0.855 $\pm$  0.045}   \\
		
		\hline
	\end{tabular}
	\label{tab:compare_ct}
	
\end{table*}

\begin{table}[t]
	\centering
	\caption{Ablation analysis of the affine network on two liver CT datasets in terms of Dice score and NCC. }
	\vspace{-0.5em}
	\begin{tabular}{|m{1.0cm}<{\centering}| m{1.4cm}<{\centering}|m{1.8cm}<{\centering}| m{1.8cm}<{\centering}| }
		\hline
		\multicolumn{2}{|c |}{Methods}       &  W/O Affine      &  W/ Affine  \\
		\hline
		\multirow{2}*{SLIVER} & Dice score    &   0.858 $\pm$ 0.053& \textbf{0.910 $\pm$ 0.027}   \\
		~ & NCC    &  0.338 $\pm$ 0.032  & \textbf{0.374 $\pm$ 0.035}   \\ \hline
		\multirow{2}*{LSPIG} & Dice score      & 0.814 $\pm$ 0.057  & \textbf{0.855 $\pm$  0.045}   \\
		~ & NCC     & 0.286 $\pm$ 0.032 & \textbf{0.348 $\pm$ 0.056}   \\ 
		\hline
	\end{tabular}
	\label{tab:ablation_affine}
\end{table}

\begin{table}[t]
	\centering
	\renewcommand\tabcolsep{4pt} 
	\caption{ Average registration time in second for Brain MR and Liver CT test pairs. }
	\vspace{-0.5em}
	\begin{tabular}{|m{1.2cm}<{\centering} |m{0.8cm}<{\centering} |m{1.0cm}<{\centering} |m{0.8cm}<{\centering} |m{0.8cm}<{\centering}  |m{1.2cm}<{\centering} |m{0.8cm}<{\centering} |}
		\hline
		Methods  & Elastix        &  NiftyReg      & SyN      & VM    & VM-diff    & Ours   \\
		\hline
		Brain MR &  167    & 323 & 4799   & 0.69   & 0.58   & \textbf{0.36}   \\
		Liver CT &  115    & 53  & 748    & 0.20   &  0.19    & \textbf{0.13}   \\
		\hline
	\end{tabular}
	\label{tab:compare_time}
\end{table}

\begin{table*}[t]
	\centering
	\caption{ Qualitative comparison results for multi-modal registration tests.   The mean and standard deviations of the Dice score are listed.      }
	\vspace{-0.5em}
	\begin{tabular}{|m{1.4cm}<{\centering}| m{1.8cm}<{\centering}|m{1.8cm}<{\centering}| m{1.8cm}<{\centering}| m{1.8cm}<{\centering}| m{1.8cm}<{\centering}| m{1.8cm}<{\centering}| m{1.8cm}<{\centering}|}
		\hline
		Methods         & Affine only & Elastix        &  NiftyReg      &  SyN      & VM    & Ours    & Ours-MIND   \\
		\hline
		T1-T2atlas   & 0.539 $\pm$ 0.007 & 0.532 $\pm$ 0.014  & 0.619 $\pm$ 0.007 & 0.528 $\pm$ 0.011  & 0.579 $\pm$ 0.013 &  0.586 $\pm$ 0.009  &  \textbf{0.625 $\pm$  0.009}  \\	
		T2-T1atlas   & 0.539 $\pm$  0.007 & 0.541 $\pm$ 0.016  & 0.639 $\pm$ 0.011  & 0.610 $\pm$ 0.010  & 0.579 $\pm$ 0.013 & 0.600 $\pm$ 0.014 &  \textbf{0.644 $\pm$ 0.007}   \\		
		\hline
	\end{tabular}
	\label{tab:compare_t1t2}	
\end{table*}

\textbf{Training strategies evaluation.}~
Given the training and validation data, thanks to our joint learning of the task-specific hyper-parameters and model parameters, our network does not have to manually select hyper-parameters during training. Instead, we design the bilevel training strategy to automatically discover the optimal value for task-specific hyper-parameters. As Tab.~\ref{tab:ablation_HO} shows, on the brain MR registration task the model taking bilevel self-tuned training can achieve much more accurate performance than the default training setting, where the hyper-parameters are manually tuned. 
When switching to liver CT data, the default hyper-parameters could not generalize well. Generally, a grid search algorithm or manually tuning will be applied to obtain these task-specific parameters, which require many training runs. In contrast, the proposed bilevel self-tuned training could auto-adapt to liver CT data and achieve satisfying performance.
These results indicate that our bilevel training strategy may offer a compelling path towards automated hyper-parameter tuning for registration networks.

\begin{figure}[b]
	\centering
	\begin{tabular}{c@{\extracolsep{0.15em}}c@{\extracolsep{0.15em}}c@{\extracolsep{0.15em}}c}	
		\includegraphics[width=0.116\textwidth]{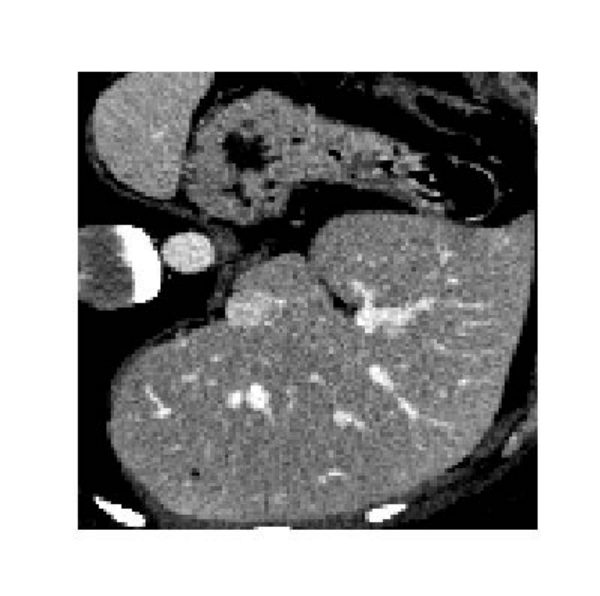}
		&\includegraphics[width=0.116\textwidth]{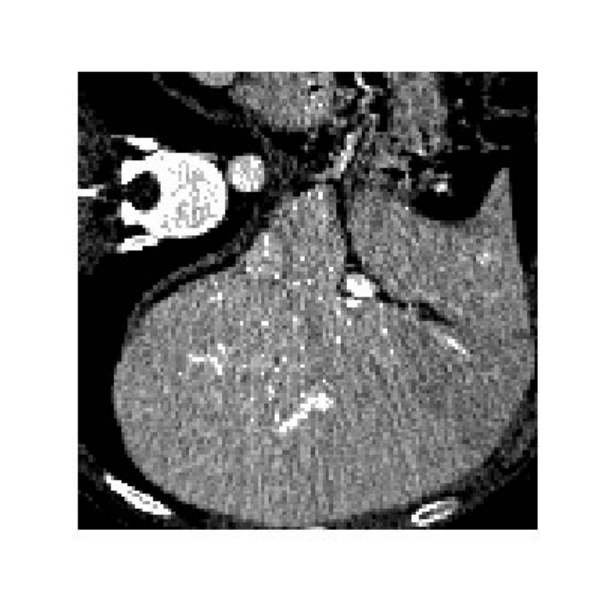}
		&\includegraphics[width=0.116\textwidth]{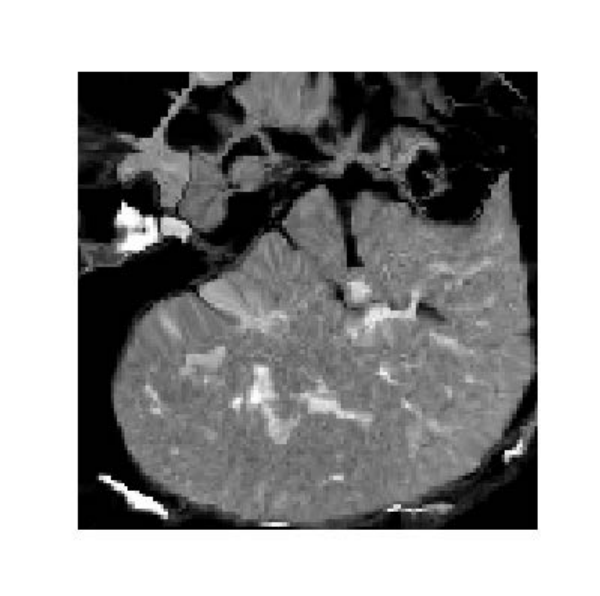}	&\includegraphics[width=0.116\textwidth]{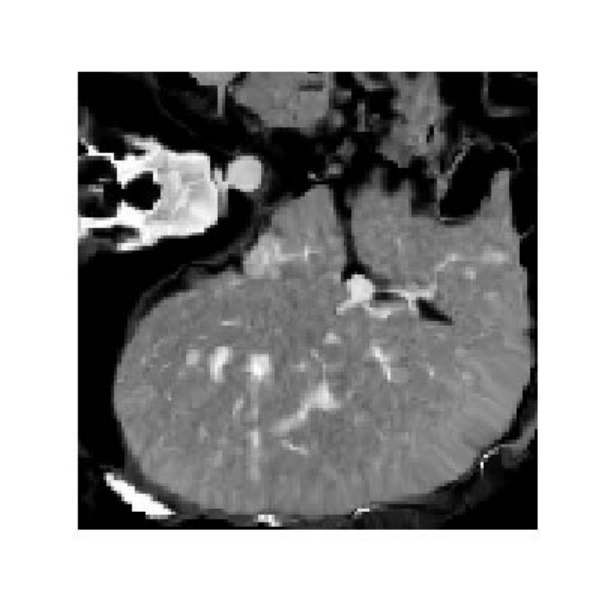}	\\
		Source  & Target &  W/O Affine 	 &  W/ Affine  \\		
	\end{tabular}
	\caption{ Illustration of the importance of affine pre-processing for registration result. The source, target, warped image of our model trained without and with the affine network.}
	\label{fig:affine} 
\end{figure}

\begin{figure*}[t]
	\centering
	\begin{tabular}{c@{\extracolsep{0.25em}}c@{\extracolsep{0.25em}}c@{\extracolsep{0.25em}}c@{\extracolsep{0.25em}}c@{\extracolsep{0.25em}}c@{\extracolsep{0.25em}}c}
		
		\includegraphics[width=0.138\textwidth]{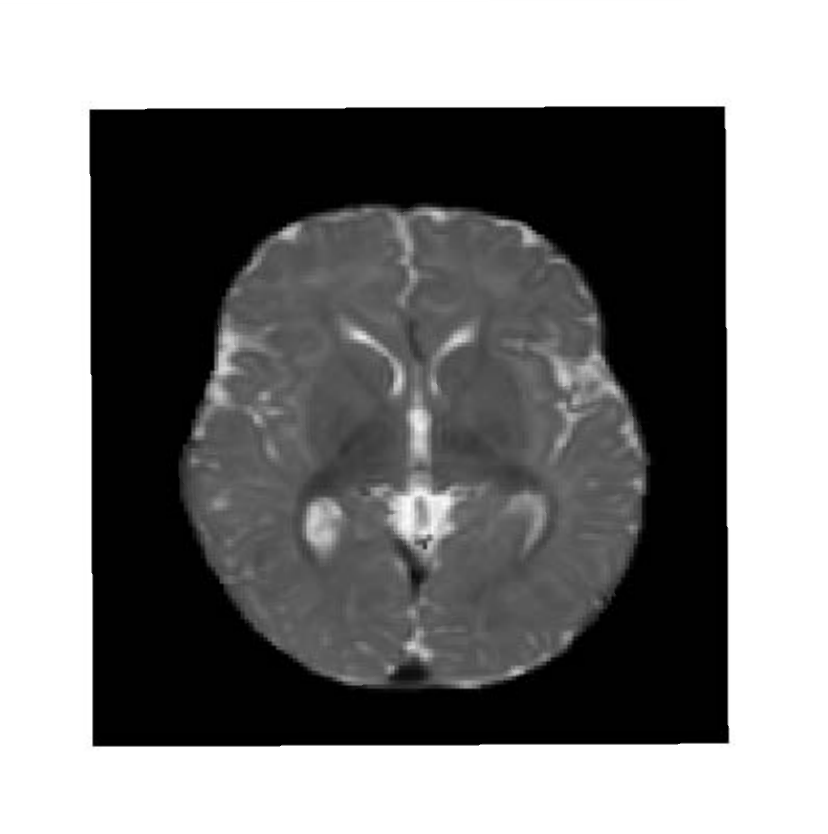}
		&\includegraphics[width=0.138\textwidth]{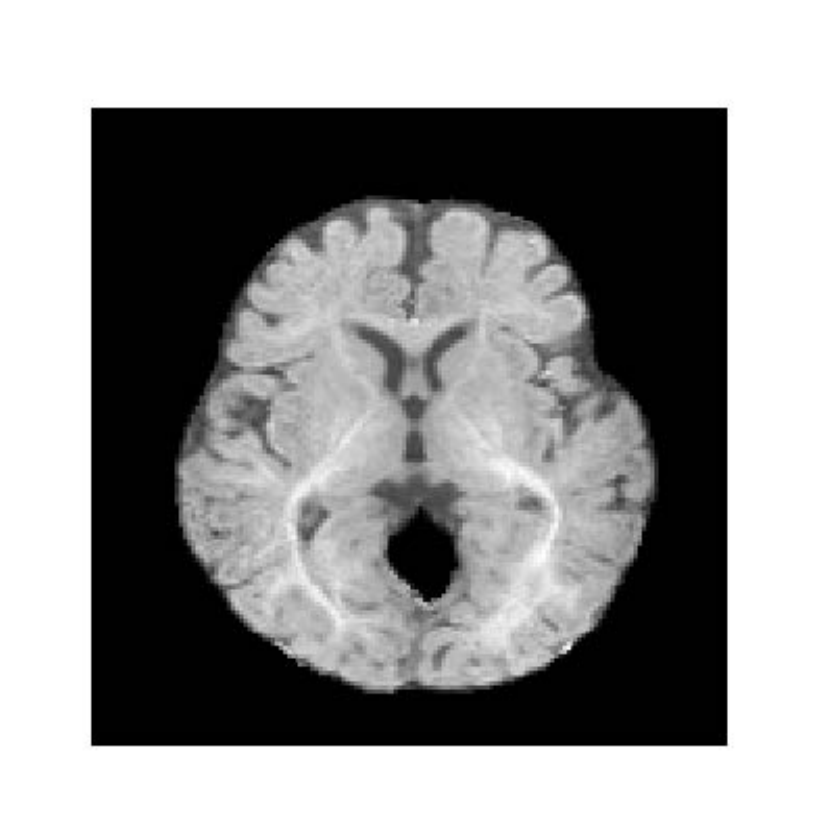}
		&\includegraphics[width=0.138\textwidth]{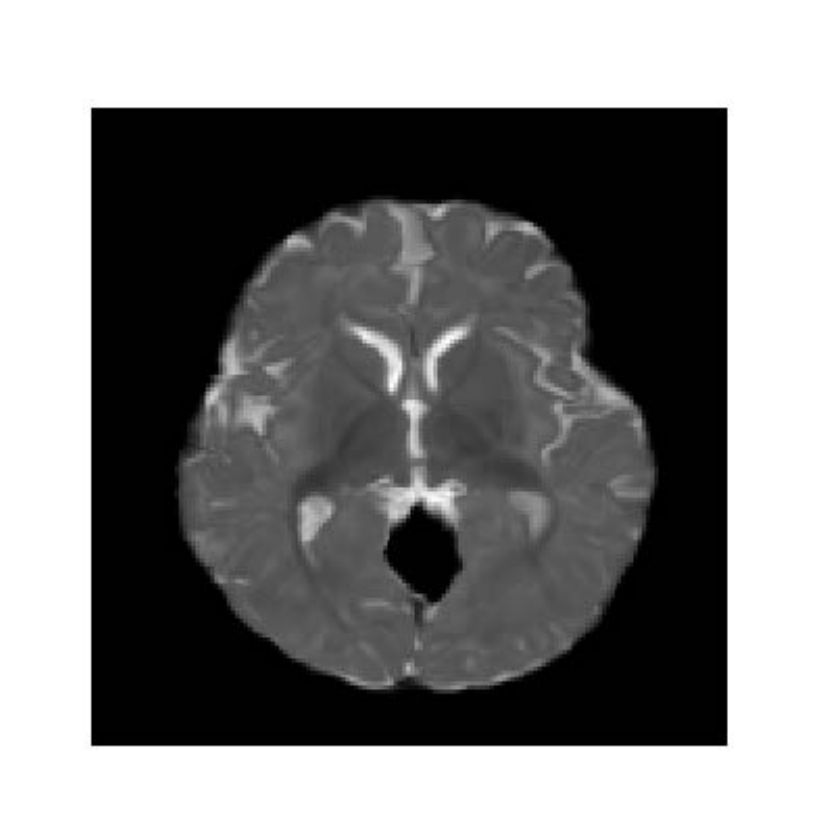}
		&\includegraphics[width=0.138\textwidth]{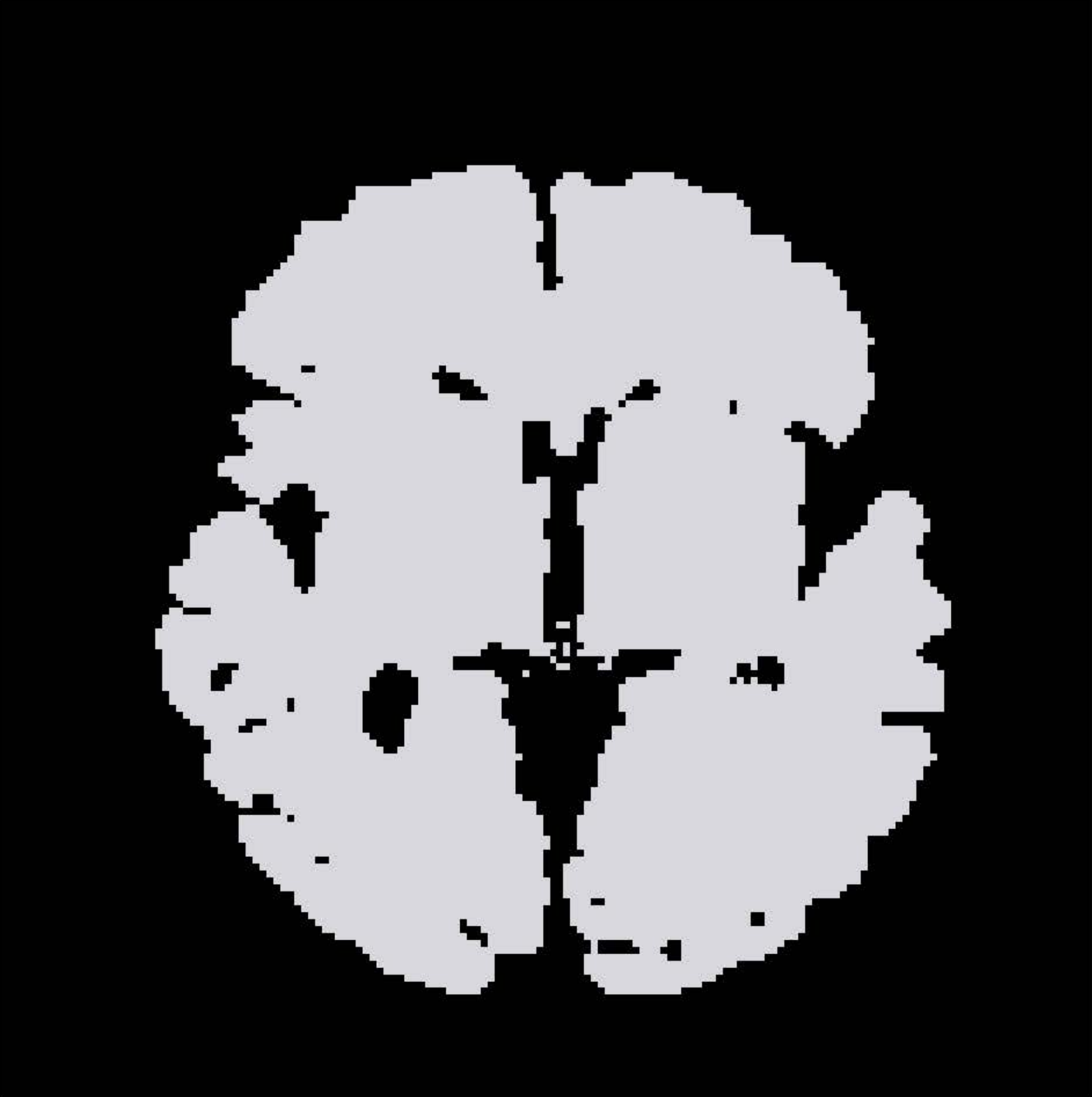}
		&\includegraphics[width=0.138\textwidth]{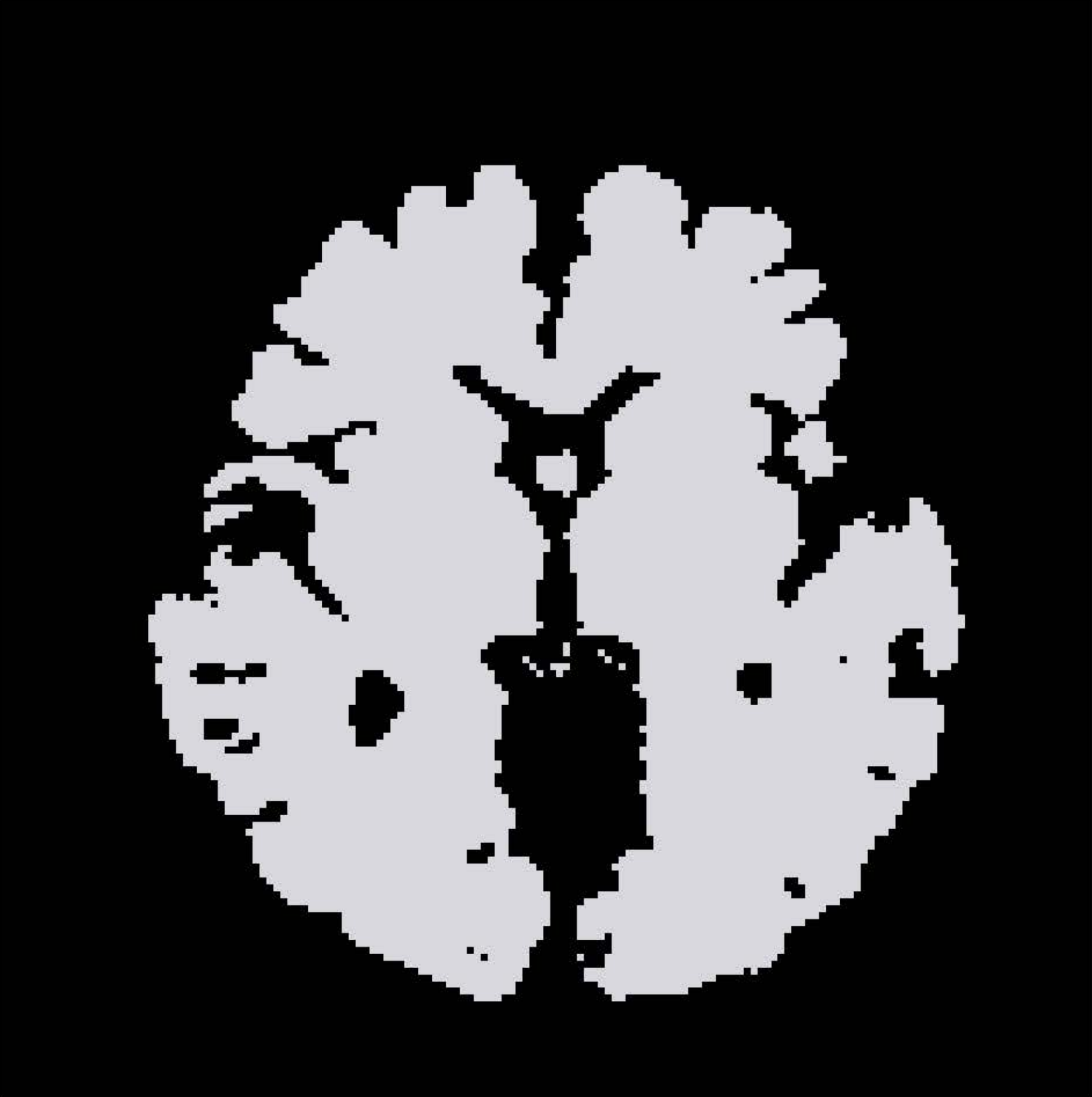}
		&\includegraphics[width=0.138\textwidth]{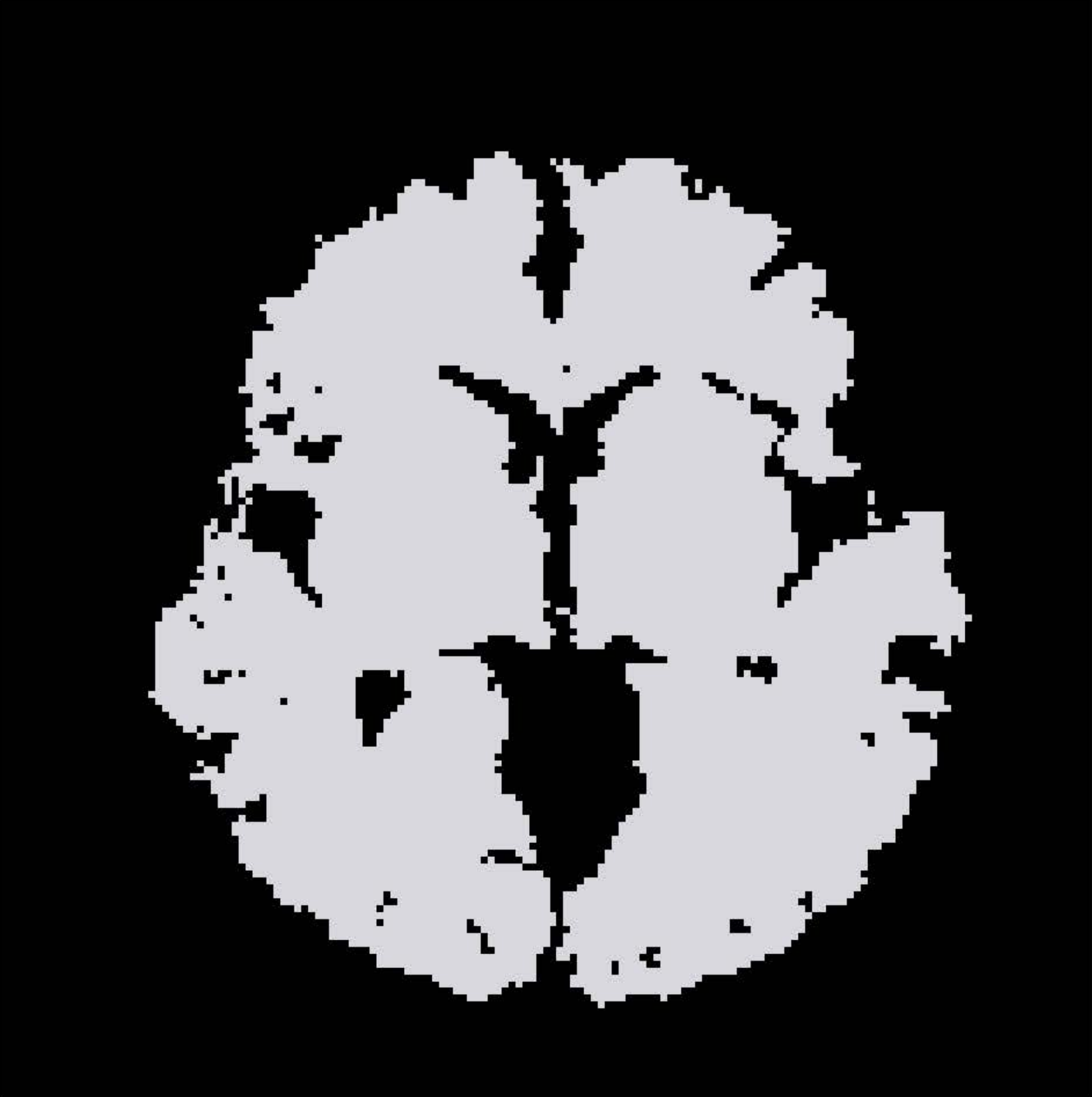}
		&\includegraphics[width=0.138\textwidth]{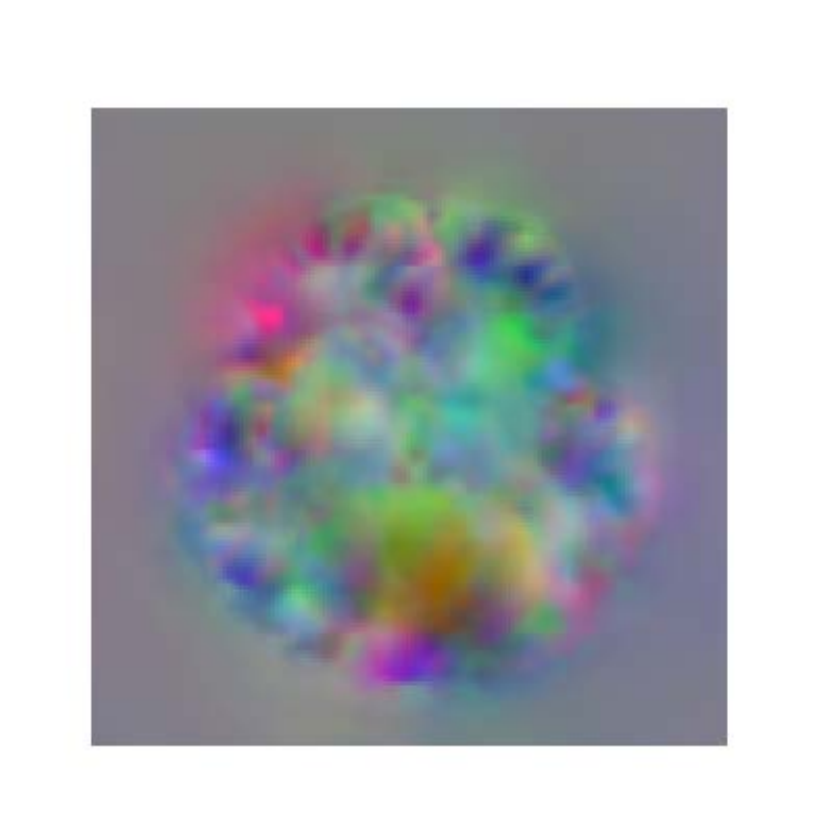}\\
		Source  & T1-atlas  &  Warped  & Source-label   &  T1-atlas-label & Warped-label & Flow field \\
		
		\includegraphics[width=0.138\textwidth]{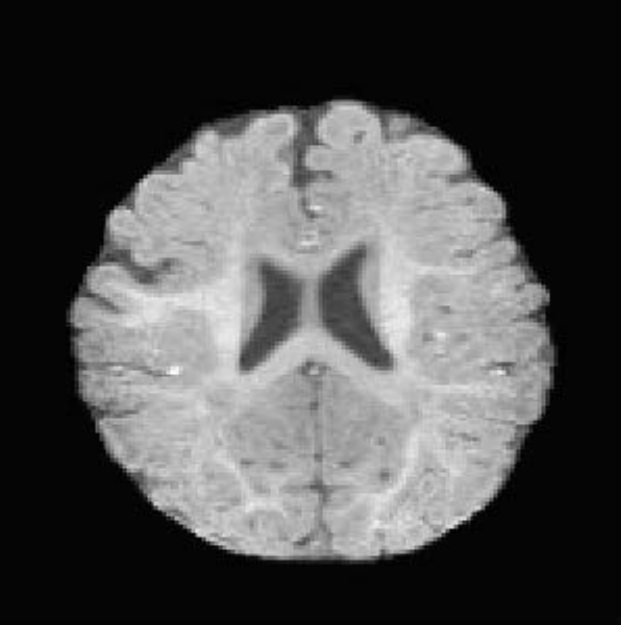}
		&\includegraphics[width=0.138\textwidth]{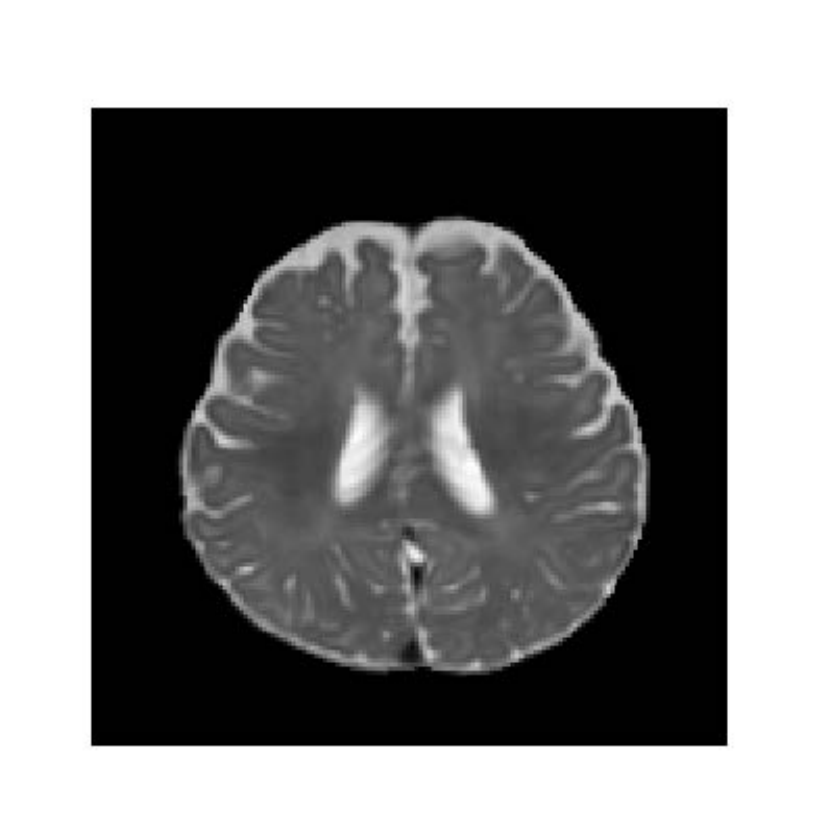}
		&\includegraphics[width=0.138\textwidth]{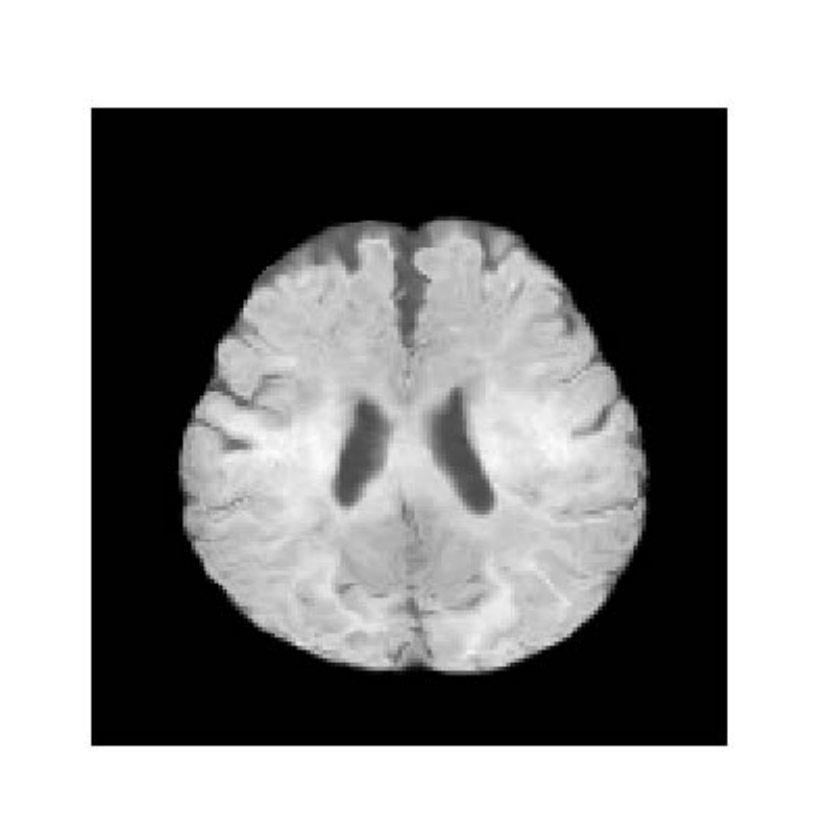}
		&\includegraphics[width=0.138\textwidth]{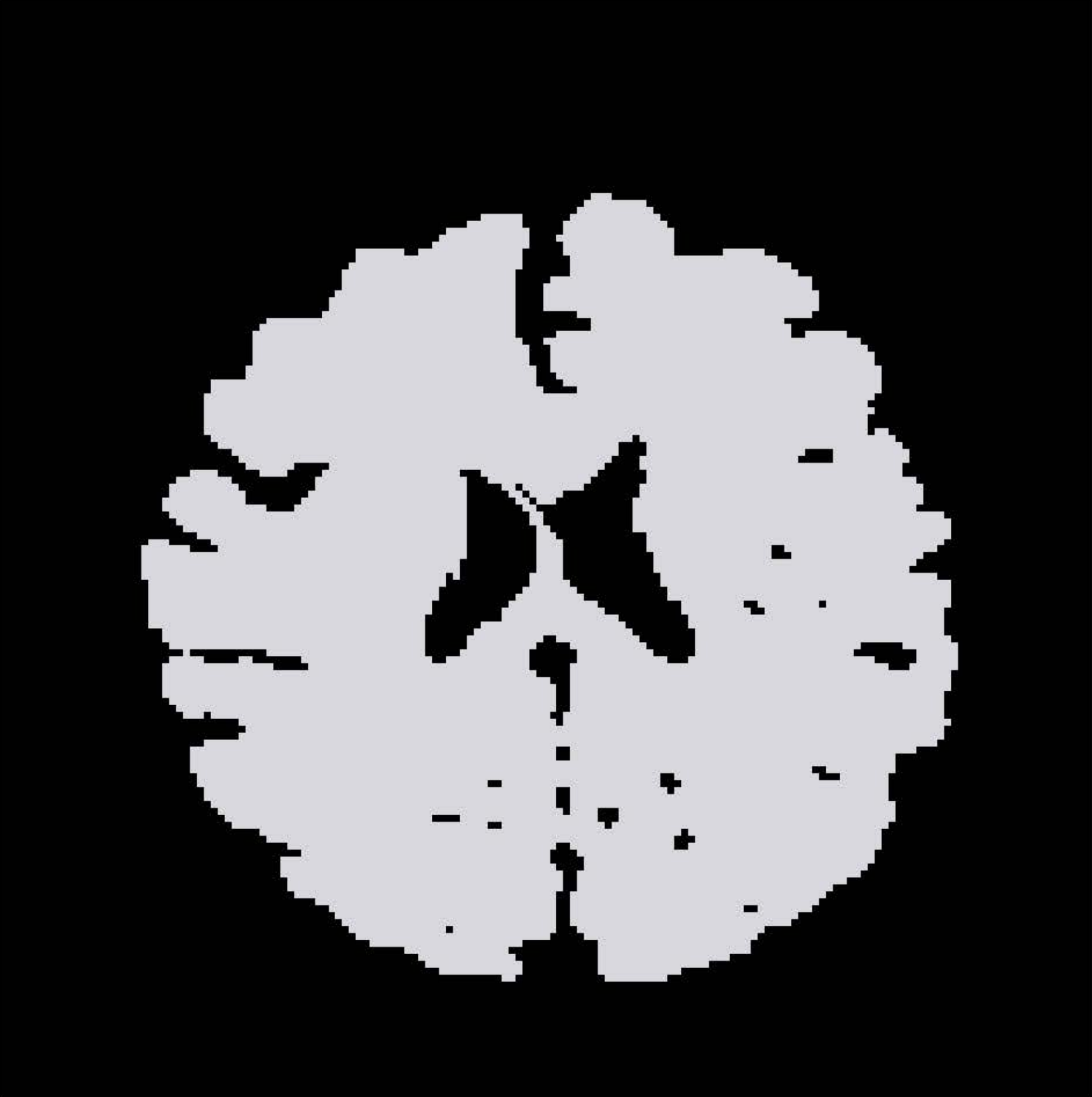}
		&\includegraphics[width=0.138\textwidth]{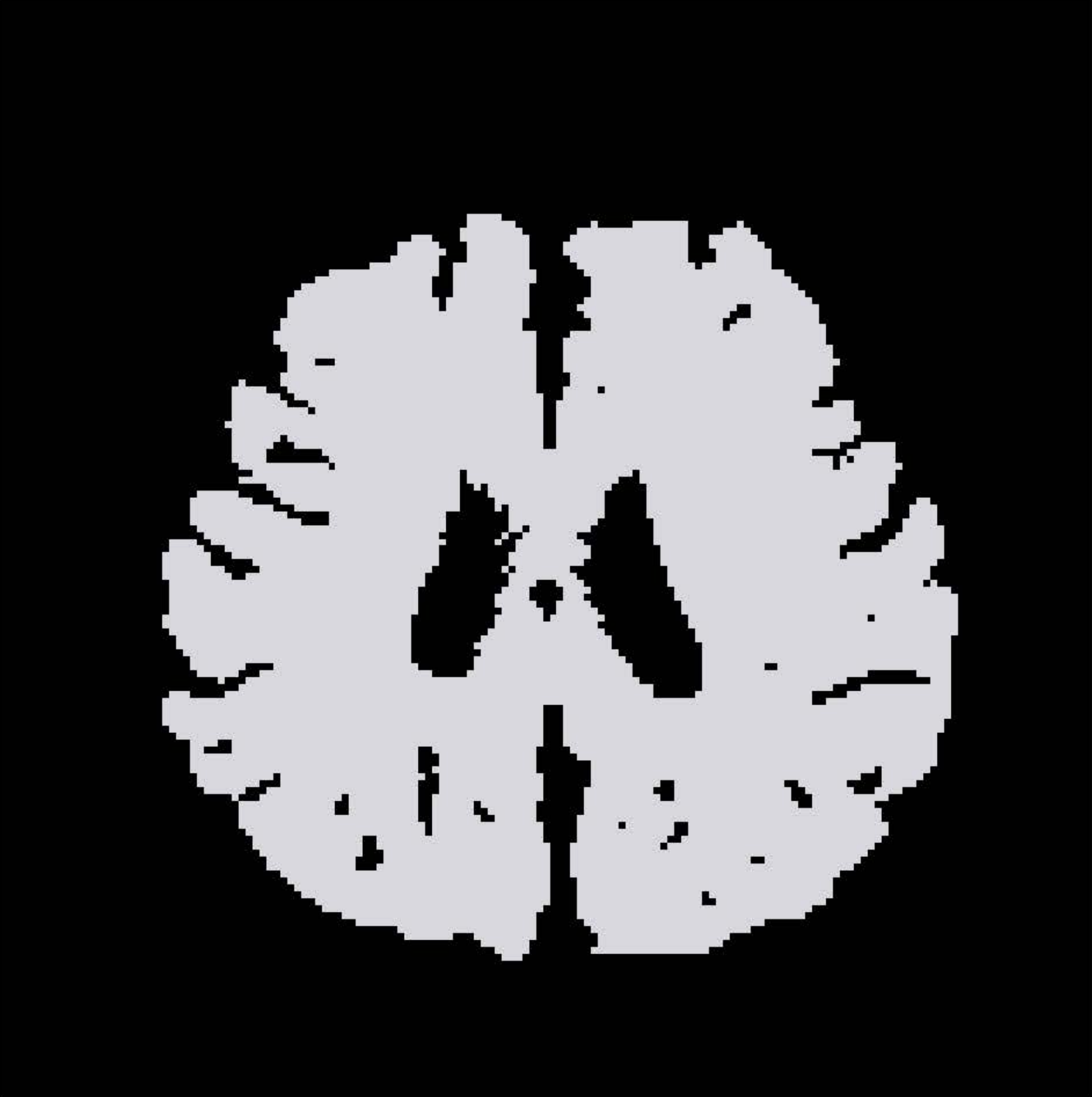}
		&\includegraphics[width=0.138\textwidth]{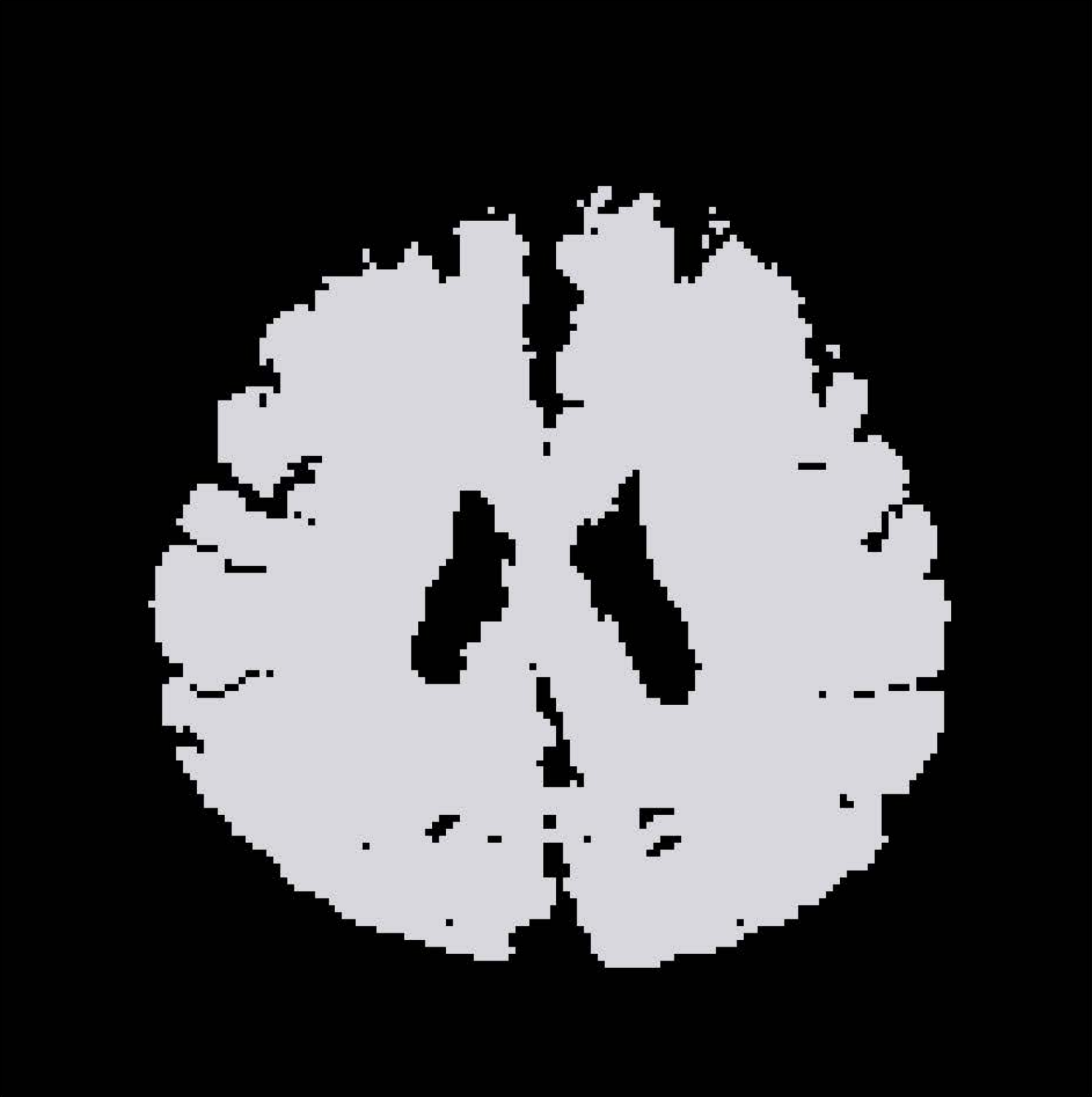}
		&\includegraphics[width=0.138\textwidth]{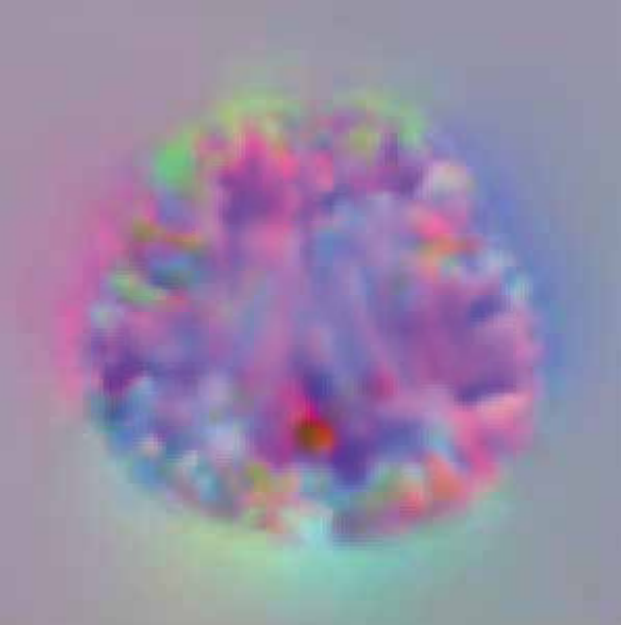}\\
		
		Source  & T2-atlas  &  Warped  & Source-label   &  T2-atlas-label & Warped-label & Flow field \\				
	\end{tabular}
	\caption{ Two example registration cases of multi-modal data. A clearly improved alignment of the ventricles is visible after registration.}
	\label{fig:example_t1t2} 
\end{figure*}

\subsection{Image-to-atlas Registration}
Firstly, Fig.~\ref{fig:example_mri} illustrates our representative registration results, two example registration cases including adult brain data and teen brain data. The large deformations that exist in the adult scans make registration challenging. As for teen MR scans, due to still in inherent myelination and maturation process, white matter and gray matter exhibit obvious differences in contrast to the fixed image, also making registration difficult. As result, all the source images are well aligned to the target, demonstrating our excellent registration performance. 

Then, we quantitatively evaluate the accuracy and rationality of all the registration techniques. 
Tab.~\ref{tab:compare_mri} depicts the accuracy and stability of the methods in terms of the Dice score on the five different datasets, where higher values and lower variance indicate a more accurate and stable registration. Our method gives an obvious lower variance and a comparable mean of Dice score on most of the datasets.  
As shown in Tab.~\ref{tab:compare_jac}, only Elastix and our method can decrease the number of folds to zero on specific datasets. However, the registration accuracy of Elastix on these datasets is far from satisfactory. 
Only our method achieves both high accuracy and strong stability while also having nearly zero non-negative Jacobian locations, benefiting from well-designed network architectures, regularization loss functions, and introduction of ordinary differential equation constraint.

To take a deeper perspective of the alignment of anatomical segmentation, we illustrate the Dice score of 30 anatomical structures in Fig.~\ref{fig:boxplot2}. Limited by space, besides our method, we take SyN and VM as the representatives for the optimization-based and learning-based techniques. We can see that compared with the top-performing conventional method SyN, the popular deep methods VM gives evenly accuracy but perform much less stable among different anatomical segmentations. In contrast, our deep model could achieve a good balance between accuracy and stability in virtue of a proper trade-off between the model-based domain knowledge and data-based deep representation. 
Fig.~\ref{fig:compare_seg} visualizes one slice of the registered segmentations generated by different methods. The target is set as semitransparent on the upper layer to present an intuitive discrepancy between the results and target. 
Compared with other approaches, ours has higher consistency with the target for both interior and outline.

Except for the boost of accuracy, we can also ideally promote efficiency by virtue of a well-designed network and regularization on estimated flow, such as context information and integration operation. Fig.~\ref{fig:arrow} visualizes both the corresponding 2D slice with a blowup of details on the right and the direction and magnitude of the generated flow with a brain volume. As the pictures on the below row show, the flow generated by our method contains fewer displacements, and the magnified details on the lower pictures demonstrate qualitatively the superiority of our method over the competitors, indicating our simpler and efficient transformation.
Fig.~\ref{fig:compare} depicts the 3D view of cortical modeling and 2D slices of the registration results, from which we can see that our method can ideally guarantee the topology of the registered volumes and preserve the contour of anatomical structure like cerebral cortex and ventricles.

\begin{figure*}[t] 
	\centering
	\begin{tabular}{m{3.3cm}<{\centering}@{\extracolsep{0.25em}}m{3.3cm}<{\centering}@{\extracolsep{0.25em}}m{3.3cm}<{\centering}@{\extracolsep{0.25em}}m{3.3cm}<{\centering}@{\extracolsep{0.25em}}m{3.3cm}<{\centering}}

		\includegraphics[width=0.18\textwidth]{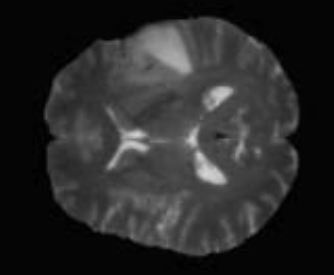}
		&\includegraphics[width=0.18\textwidth]{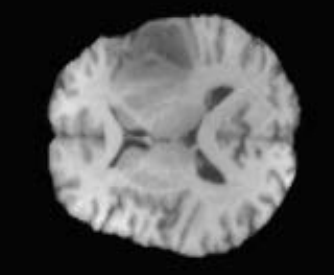}
		&\includegraphics[width=0.18\textwidth]{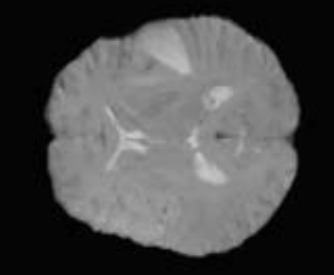}
		&\includegraphics[width=0.18\textwidth]{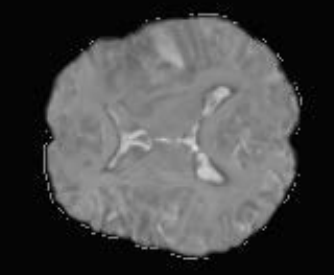}
		&\includegraphics[width=0.18\textwidth]{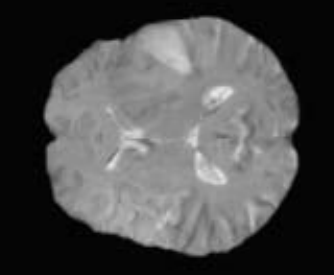} \\

		\includegraphics[width=0.18\textwidth]{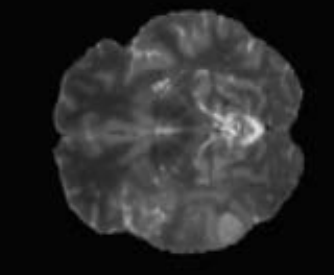}
		&\includegraphics[width=0.18\textwidth]{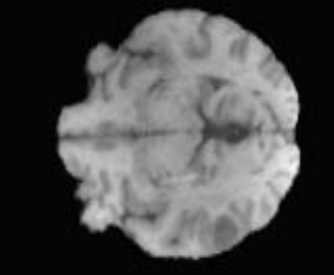}
		&\includegraphics[width=0.18\textwidth]{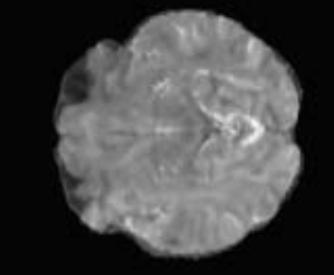}
		&\includegraphics[width=0.18\textwidth]{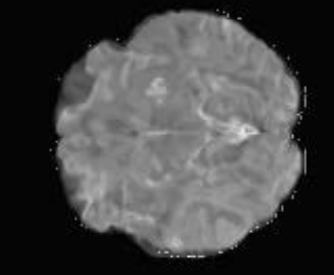}
		&\includegraphics[width=0.18\textwidth]{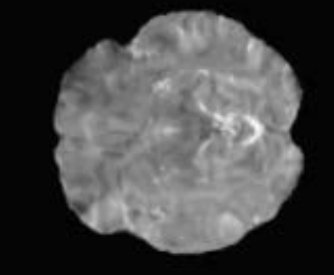} \\

		\includegraphics[width=0.18\textwidth]{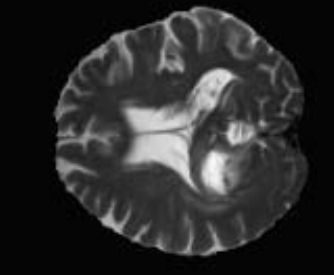} 
		&\includegraphics[width=0.18\textwidth]{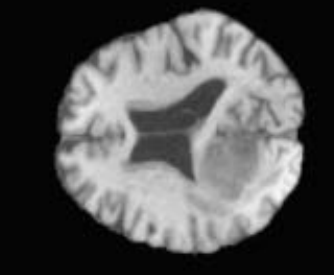} 
		&\includegraphics[width=0.18\textwidth]{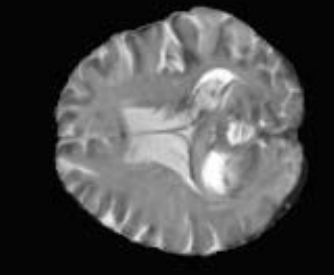} 	
		&\includegraphics[width=0.18\textwidth]{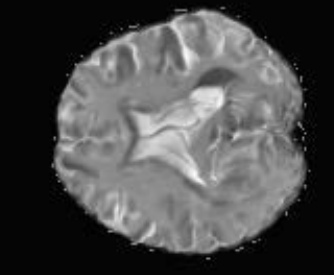} 
		&\includegraphics[width=0.18\textwidth]{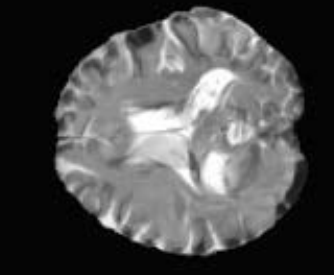}\\		
		
		(a)	&  (b)  &  (c)  &  (d)  &  (e)  \\

	\end{tabular}
	\caption{ Comparisons on multi-modal fusion results by fusing the source and target without and with registration pre-processing. The (a)-(e) columns refer to target, source, fusion result without registration, fusion result with NiftyReg registration method and fusion result with our registration method. Limited by space, we take the second-best NiftyReg as the representative comparison method (cf Tab.~\ref{tab:compare_t1t2}).}
	\label{fig:fusion_t1t2} 
\end{figure*}

\subsection{Image-to-image Registration}
To show the generalizability, we extended our paradigm to the case of challenging image-to-image Liver CT registration tasks. 
We evaluated the accuracy, rationality of all the registration methods. Additionally, we propose an initial affine network to perform affine transform before predicting deformation fields, substituting the traditional affine preprocessing. We also demonstrated the benefits of the affine network.

Visual registration examples of liver CT scans are shown in Fig.~\ref{fig:example_ct}, with the first raw data from LSPIG~\cite{ZhaoDCX19}, covering intrasubject registration with the image pair which comes from the same pig (preoperative to perioperative), and others from SLIVER~\cite{HeimannGS09}, covering intersubject registration for different persons. Although large deformations exist, source images are well aligned to the target, demonstrating the good registration performance of our approach. 
Tab.~\ref{tab:compare_ct} depicts the stability of the methods in terms of the Dice score on the different datasets, where higher values and lower variance indicate a more accurate and stable registration.
Our method gives an obvious lower variance with a comparable mean of Dice score, demonstrating superiority over the competitors.

Tab.~\ref{tab:ablation_affine} depicts the qualitative comparison between the network trained without and with the affine network under the experiment setting of liver CT registration. Fig.~\ref{fig:affine} gives the illustration of the impact of affine transformation on the final registration result. As shown, the registered image is more rational after introducing the affine network. Moreover, the affine network helps to achieve a much more accurate and robust alignment quality.

We reported the elapsed time for computations on brain MR and liver CT registration tasks. 
Cause of the data size, the runtime of registring the brain MR test pair is a little more than the liver CT scans.
As shown in Tab.~\ref{tab:compare_time}, on brain MRI registration task, conventional optimization-based SyN  require two or more hours and NiftyReg requires roughly ten minutes, while the proposed method completes the registration for 3D medical image pairs in under half-second and runs orders of magnitude faster than conventional techniques. 
Owing to the well-designed lightweight network and dealing with half-resolution, we require less runtime even compared with learning-based methods.

\begin{figure*}[t] 
	\centering
	\begin{tabular}{m{3.4cm}<{\centering}@{\extracolsep{0.25em}}m{3.4cm}<{\centering}@{\extracolsep{0.25em}}m{3.4cm}<{\centering}@{\extracolsep{0.25em}}m{3.4cm}<{\centering}@{\extracolsep{0.25em}}m{3.4cm}<{\centering}}
		
		\includegraphics[width=0.184\textwidth]{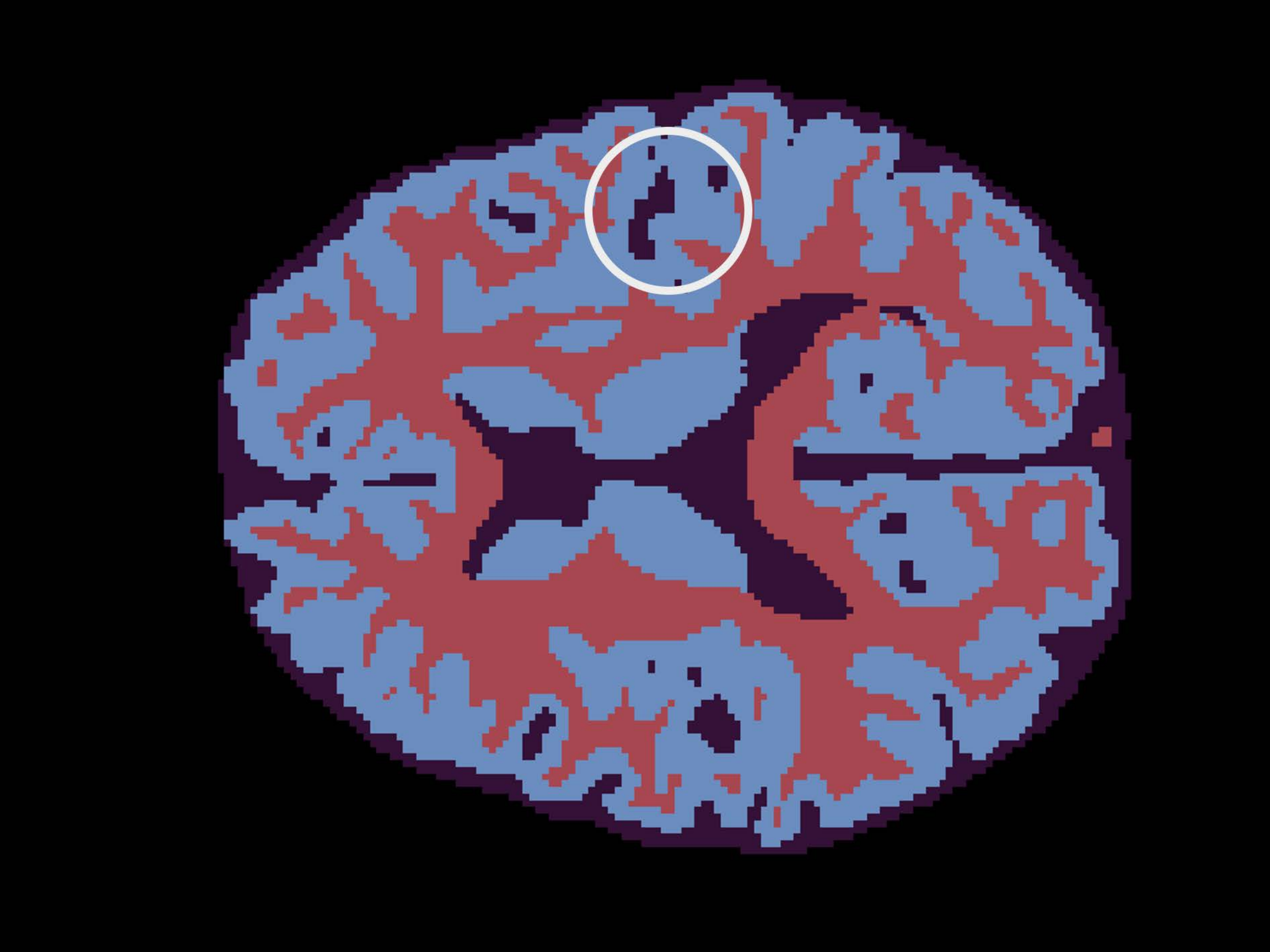}
		&\includegraphics[width=0.184\textwidth]{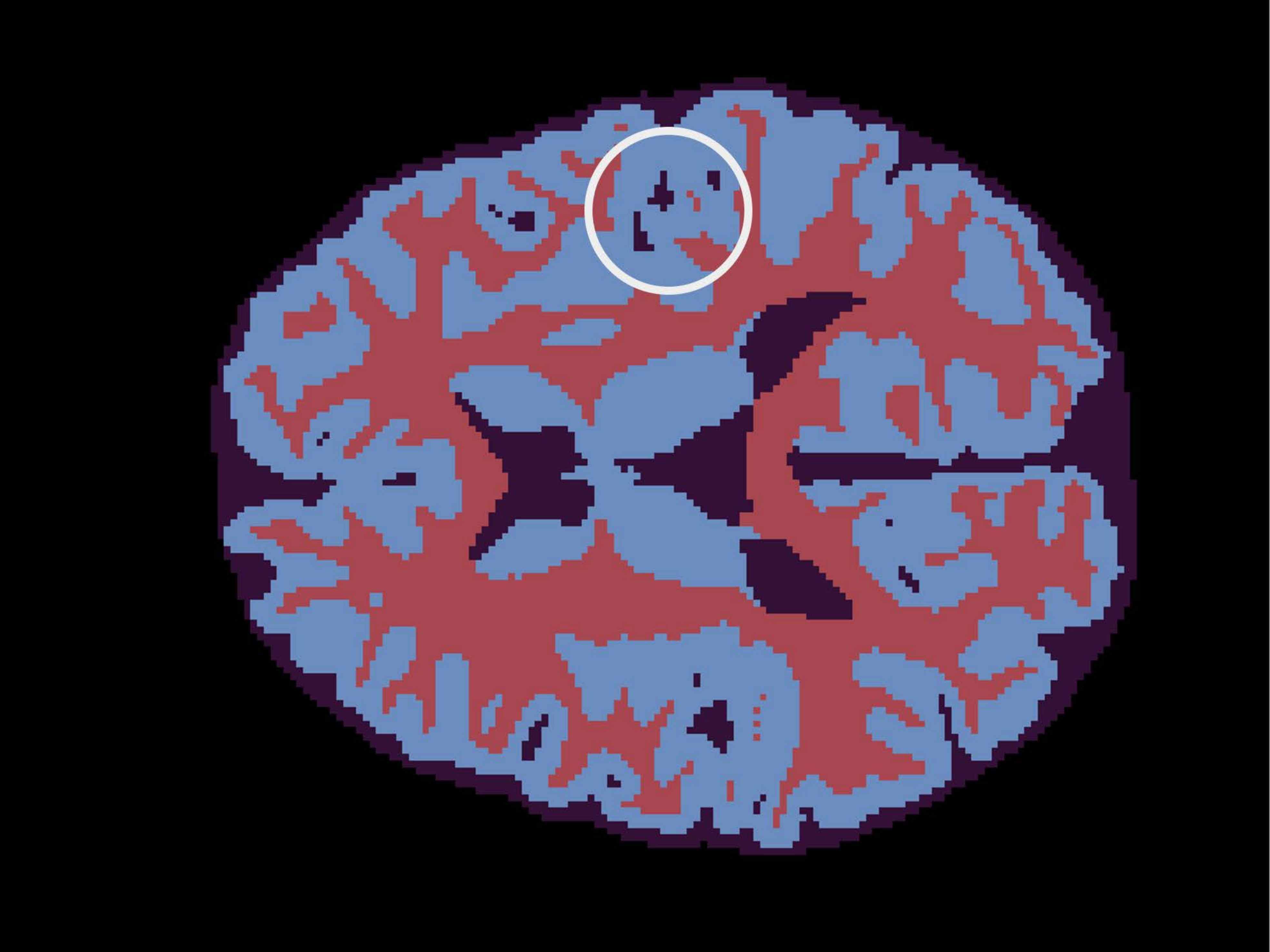} 
		&\includegraphics[width=0.184\textwidth]{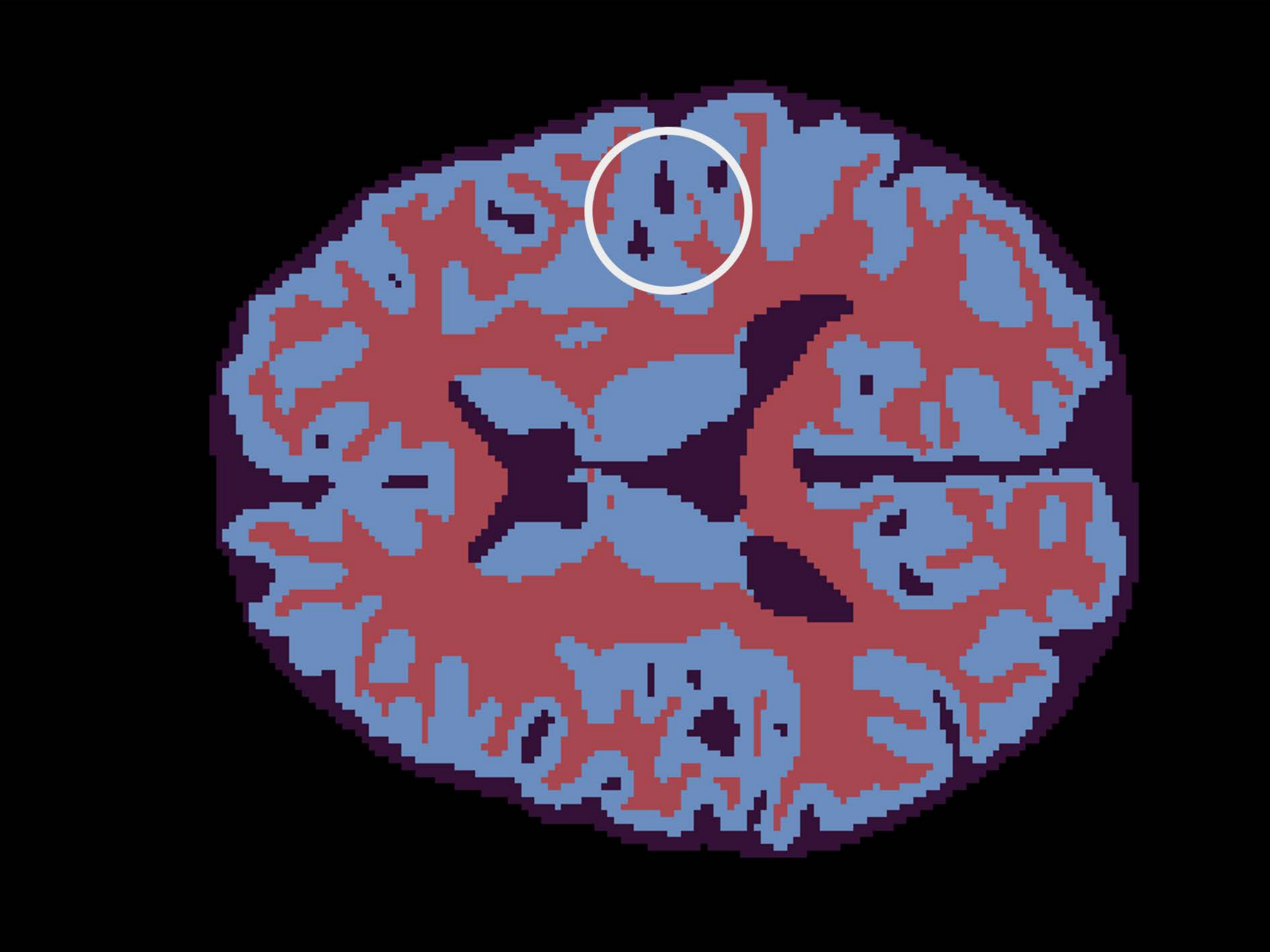}
		&\includegraphics[width=0.184\textwidth]{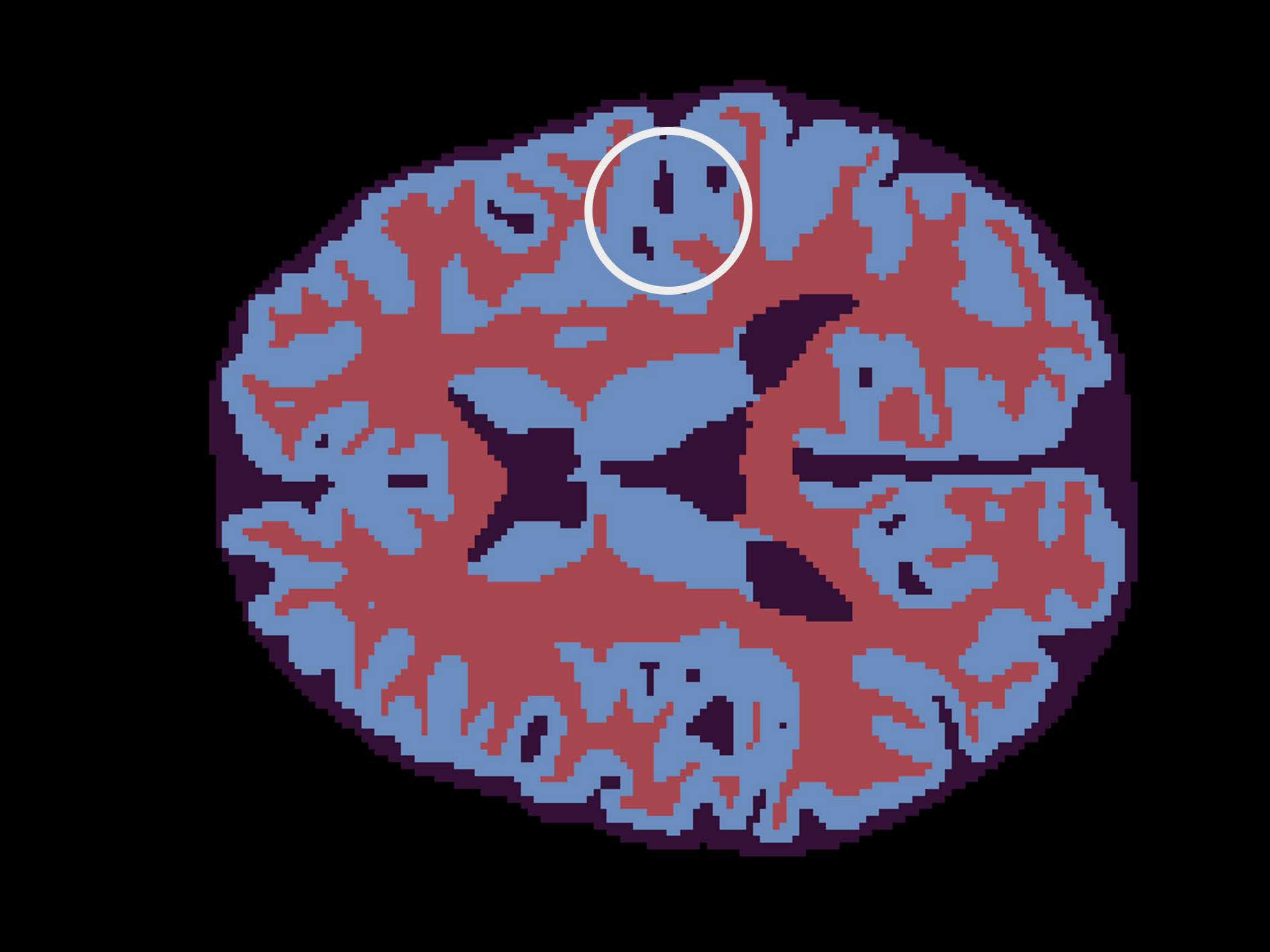} 
		&\includegraphics[width=0.184\textwidth]{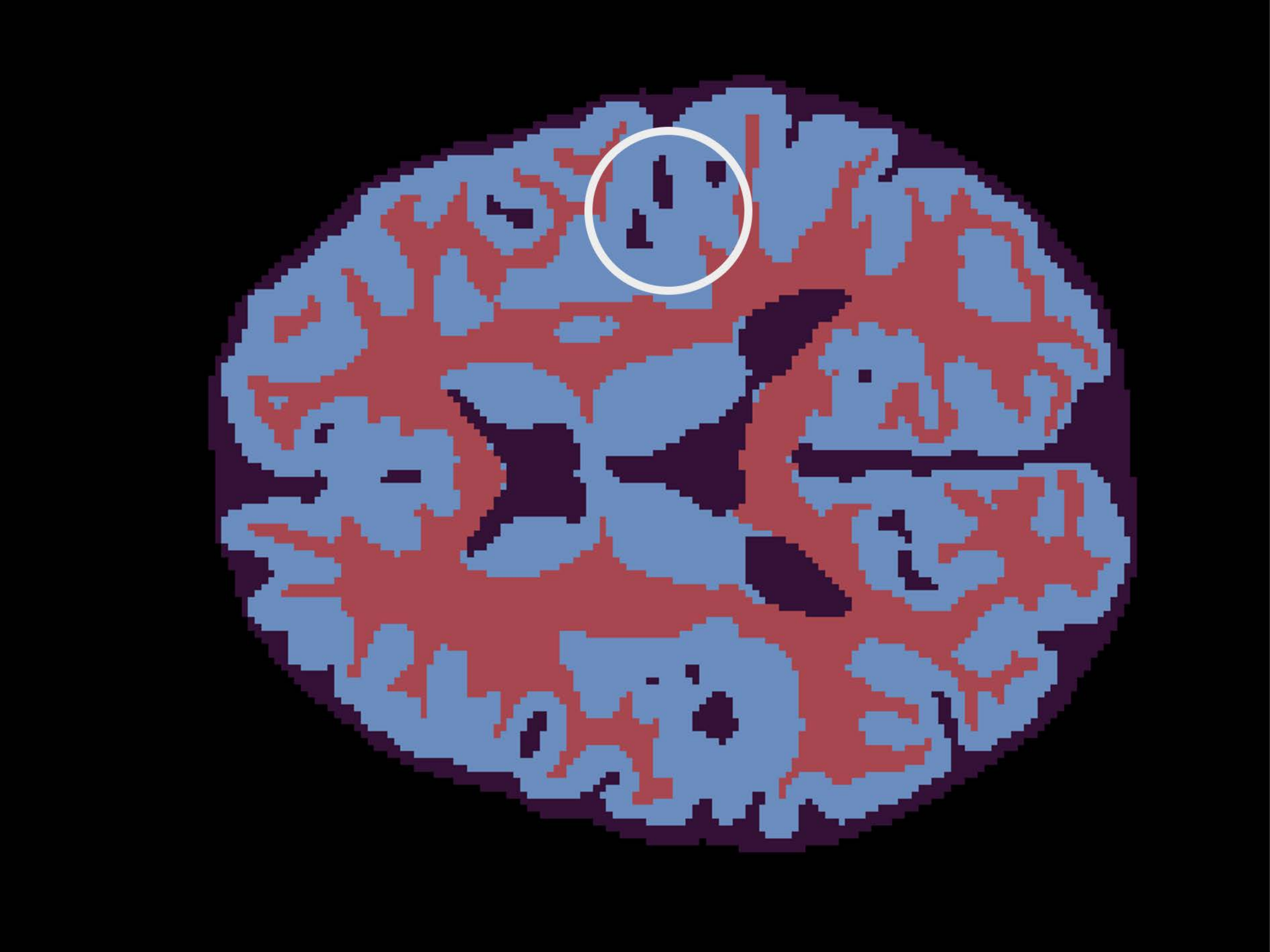} \\

		\includegraphics[width=0.184\textwidth]{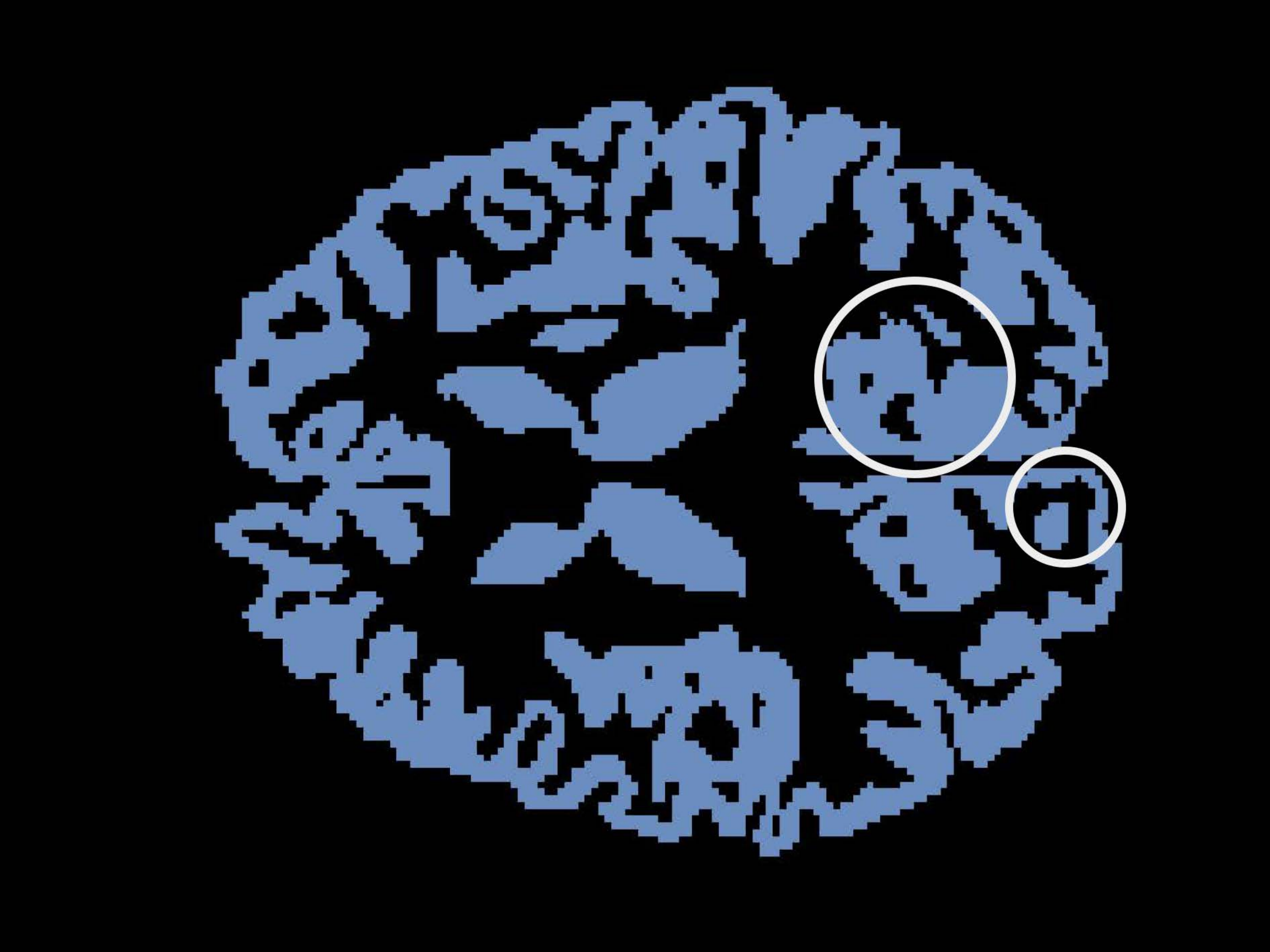}
		&\includegraphics[width=0.184\textwidth]{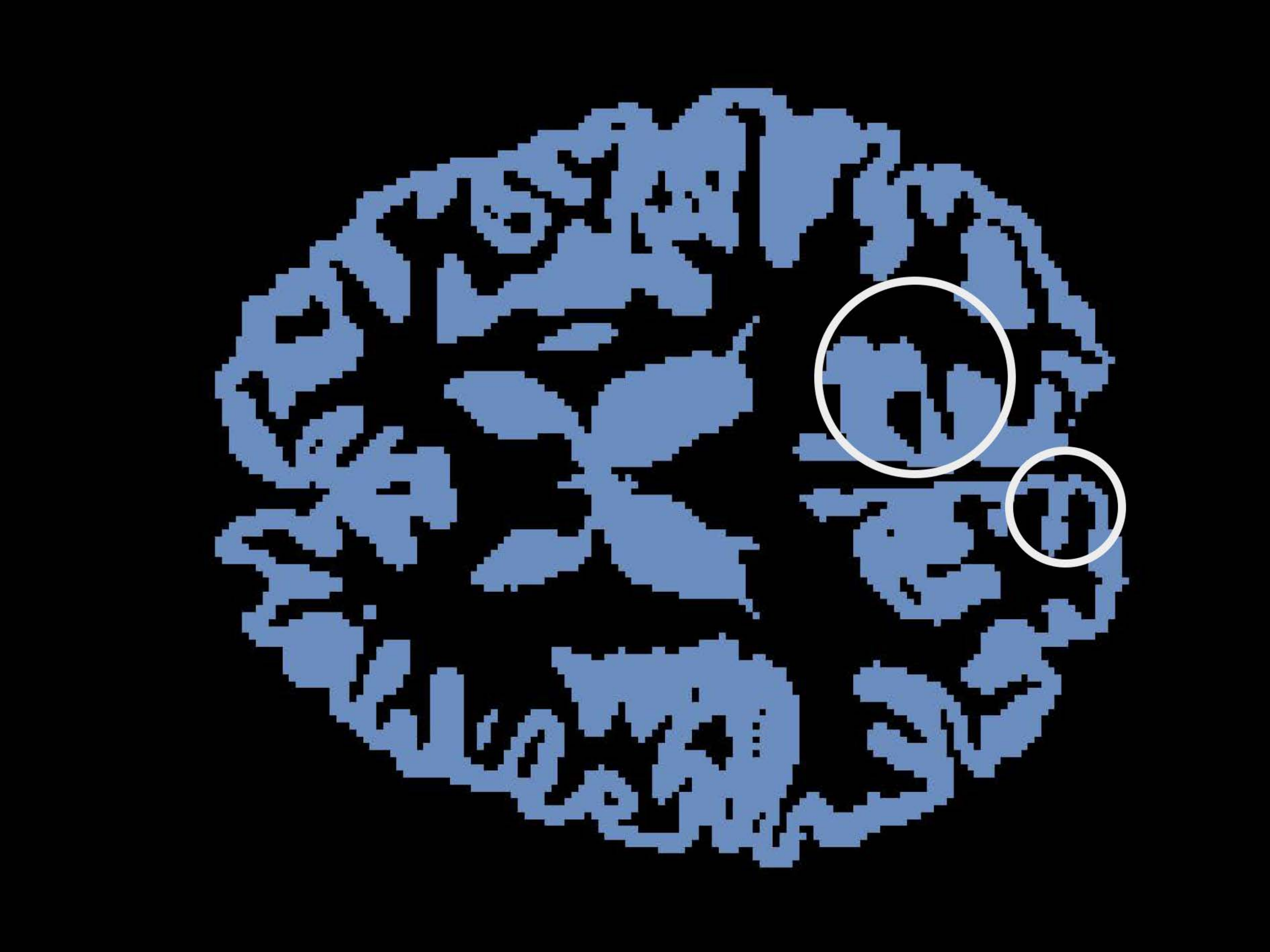}
		&\includegraphics[width=0.184\textwidth]{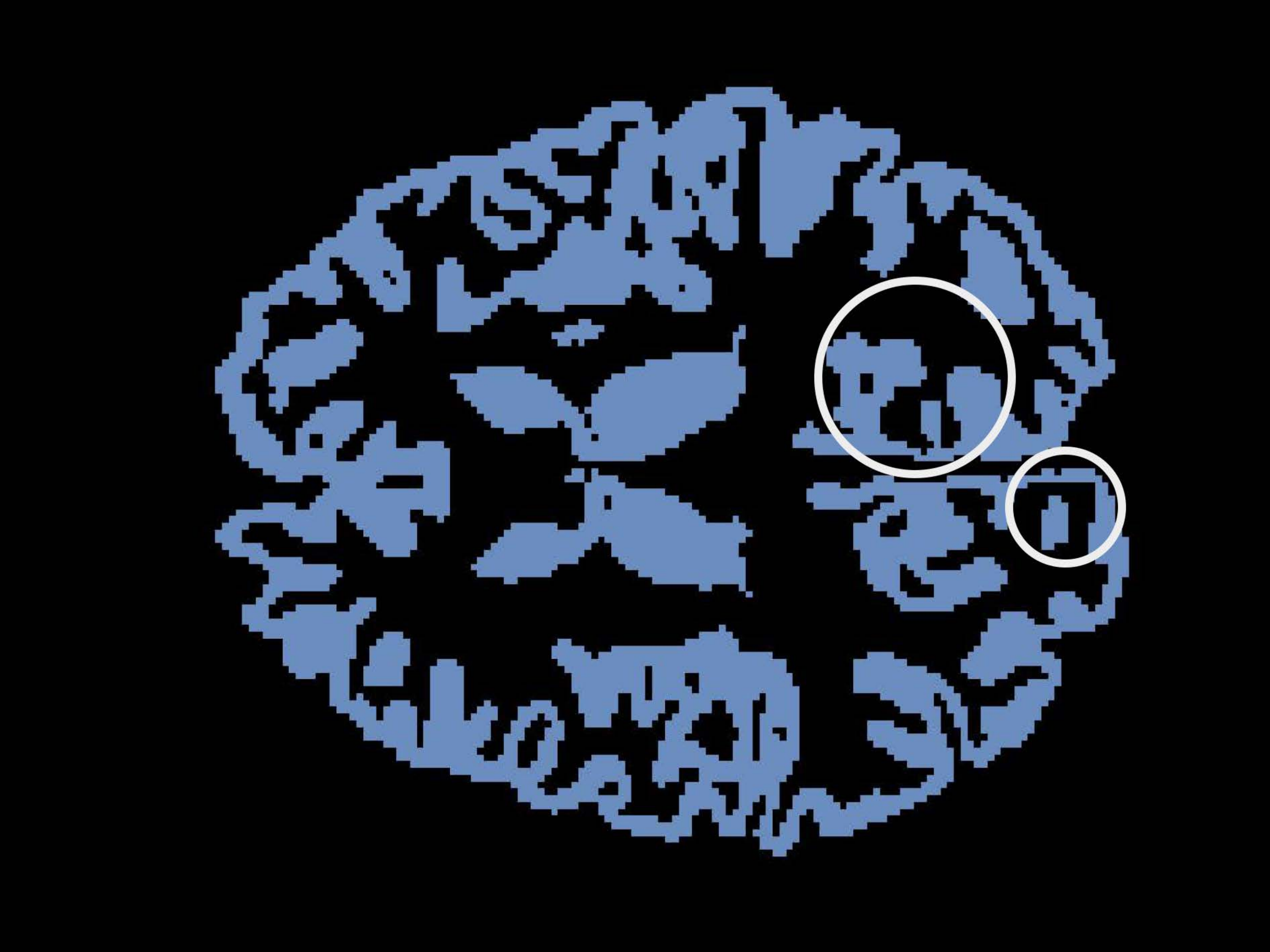}
		&\includegraphics[width=0.184\textwidth]{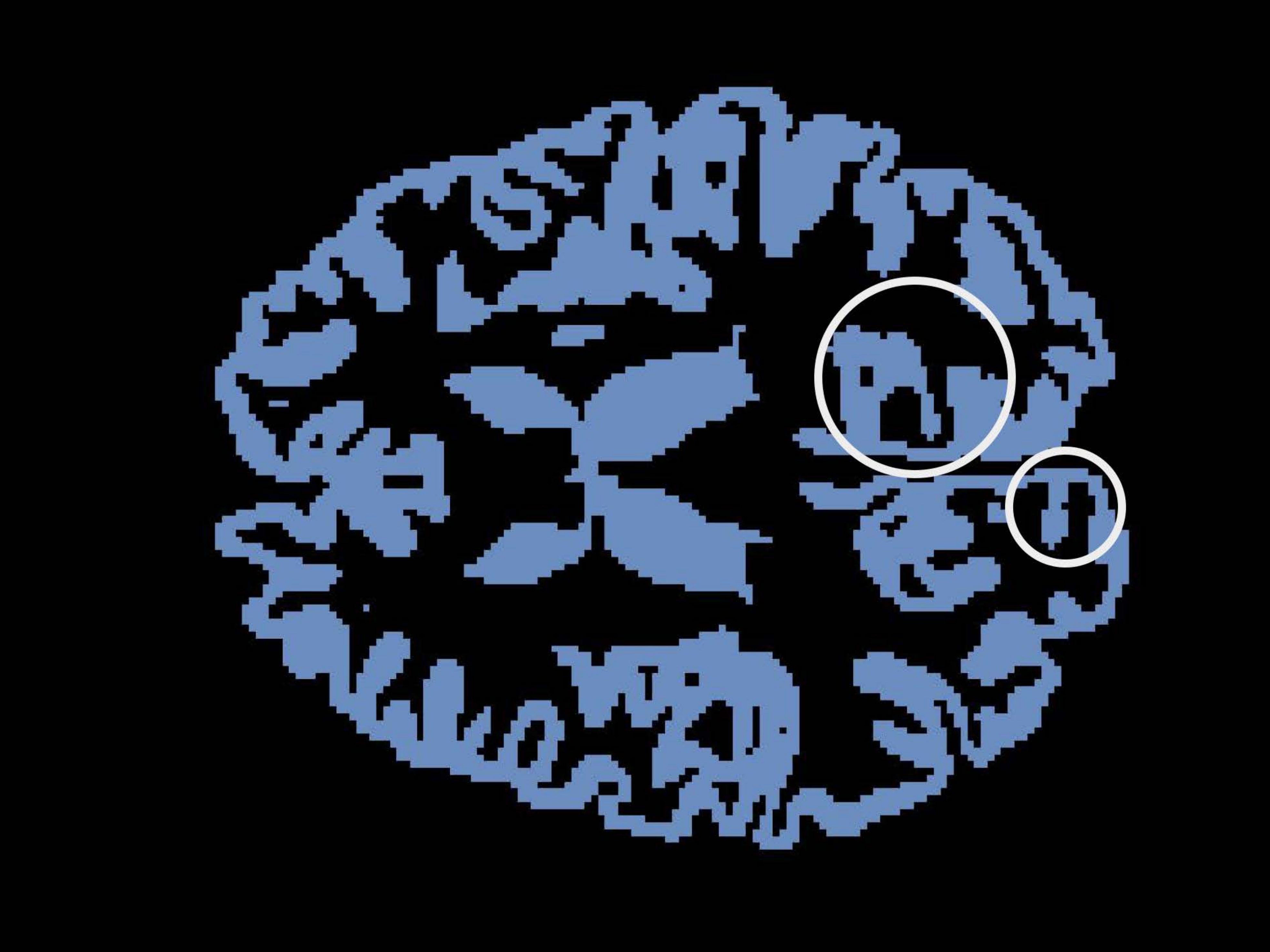}
		&\includegraphics[width=0.184\textwidth]{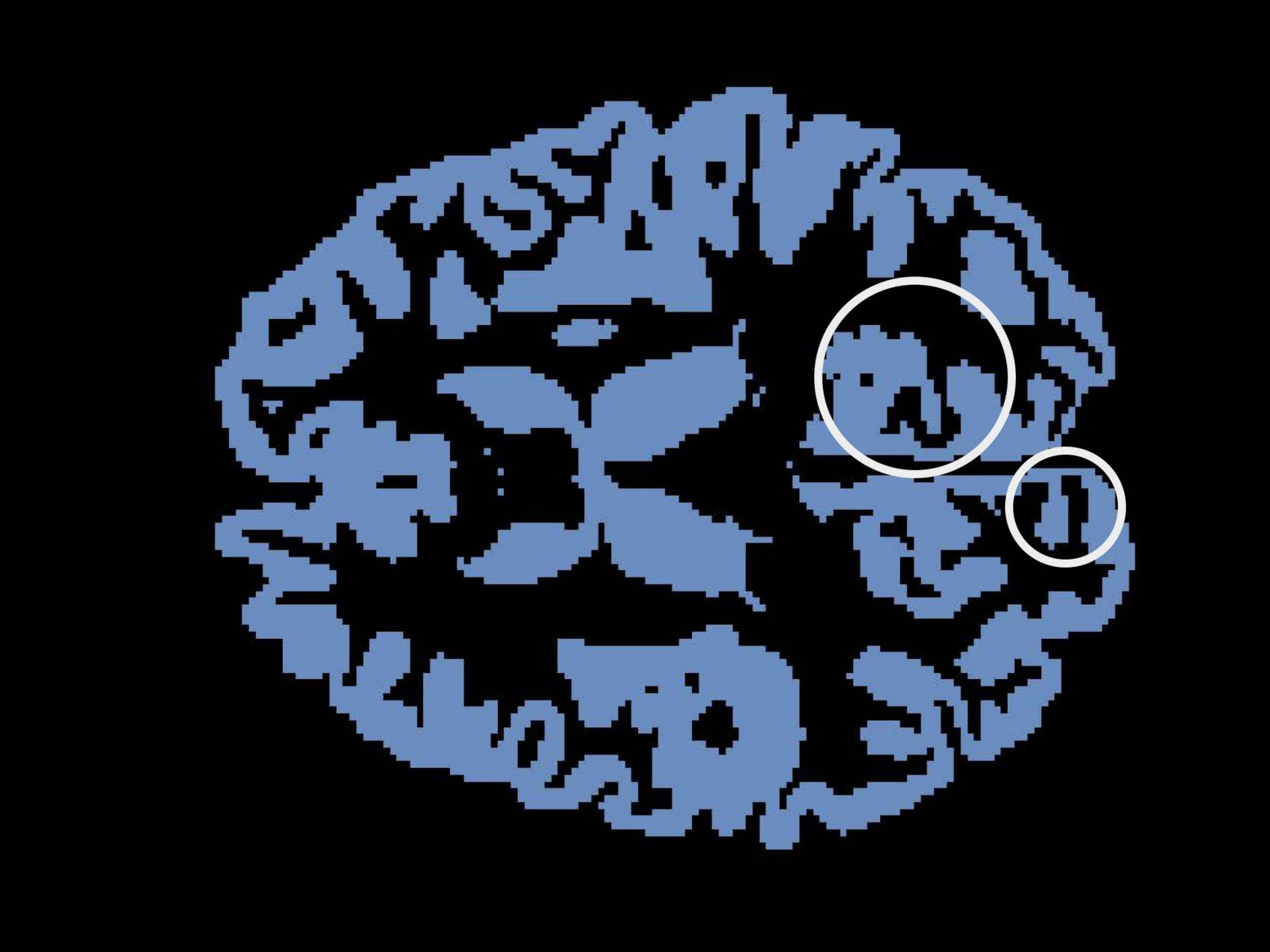} \\
		
		\includegraphics[width=0.184\textwidth]{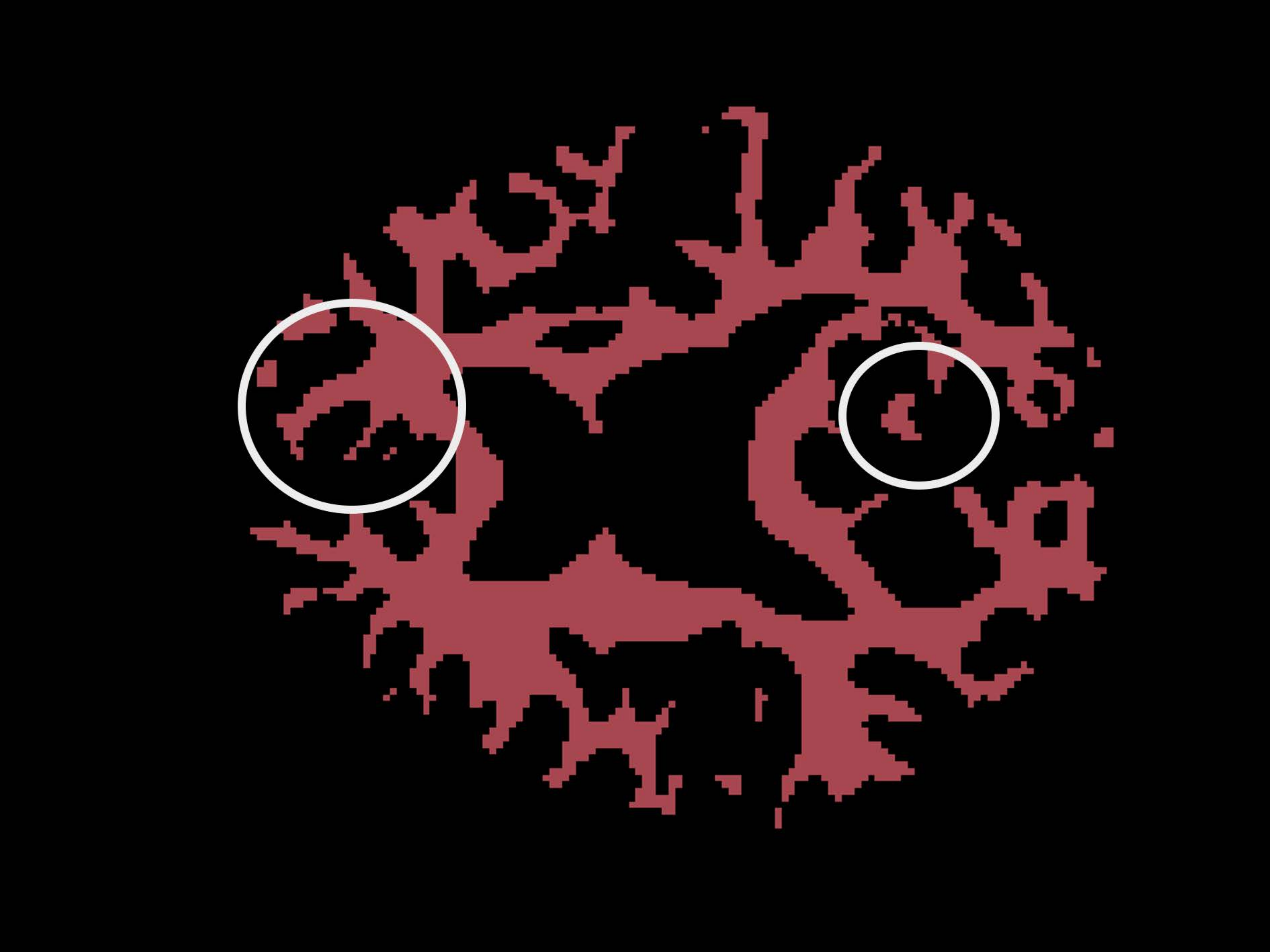}
		&\includegraphics[width=0.184\textwidth]{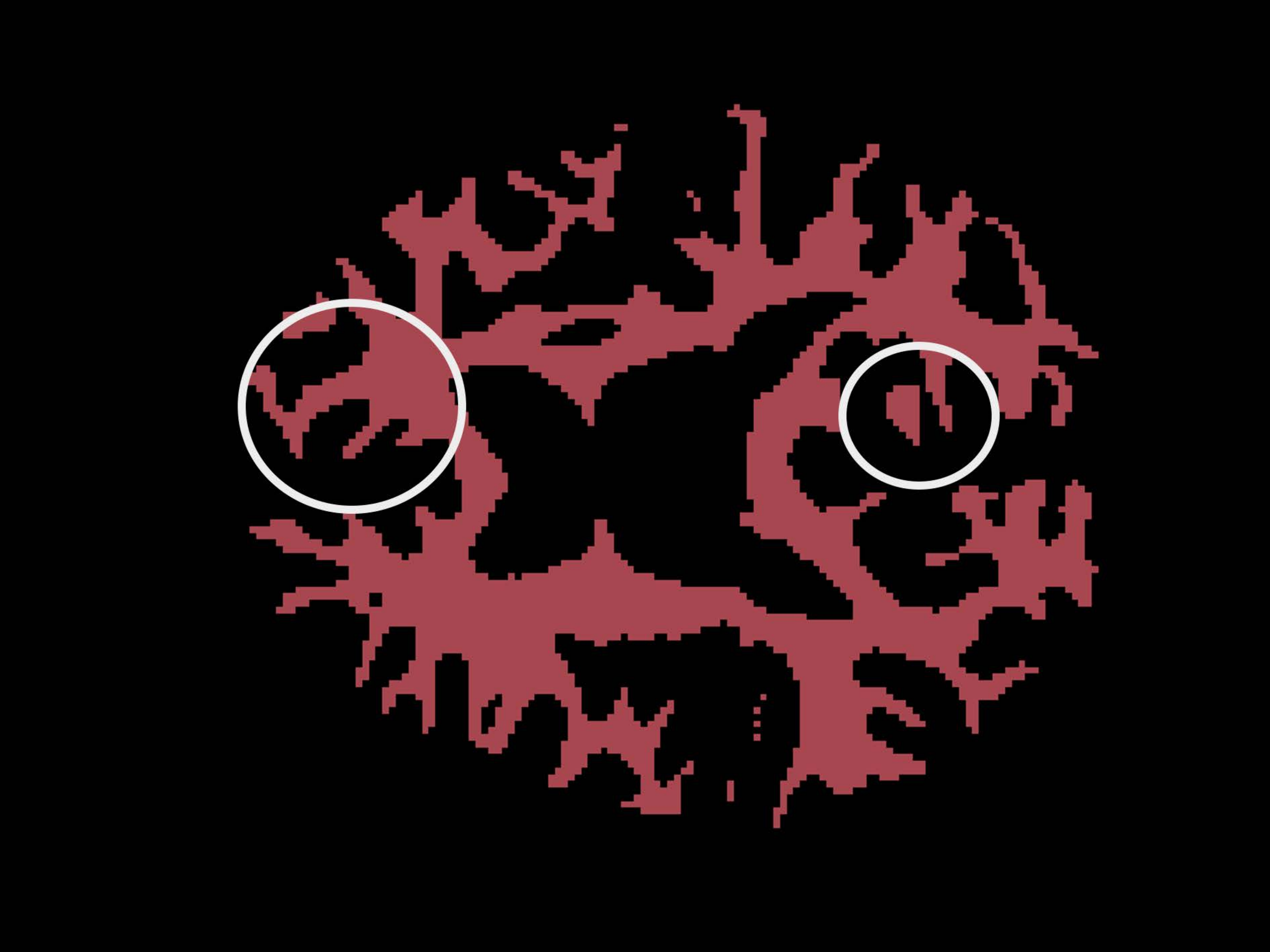}
		&\includegraphics[width=0.184\textwidth]{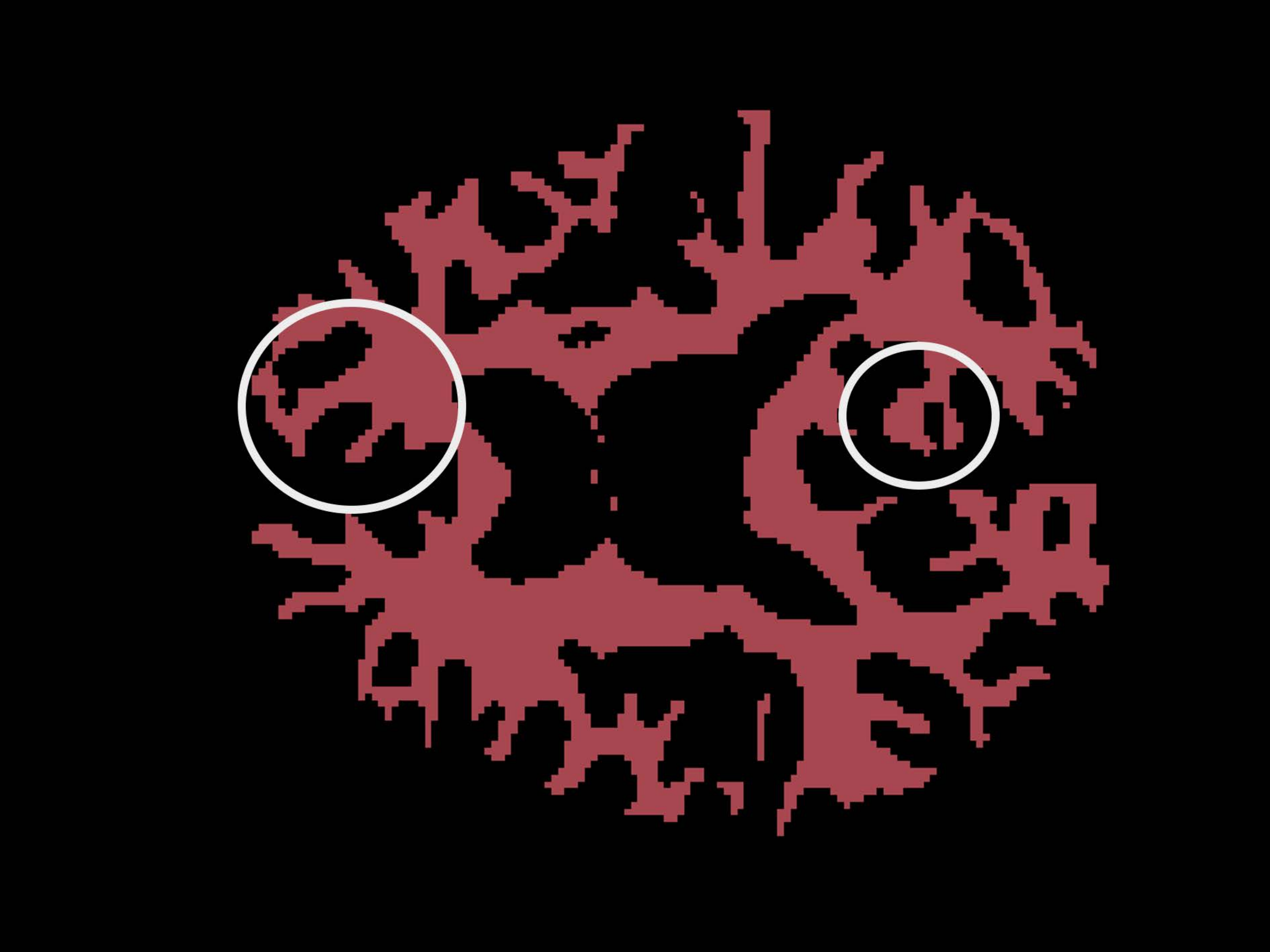}
		&\includegraphics[width=0.184\textwidth]{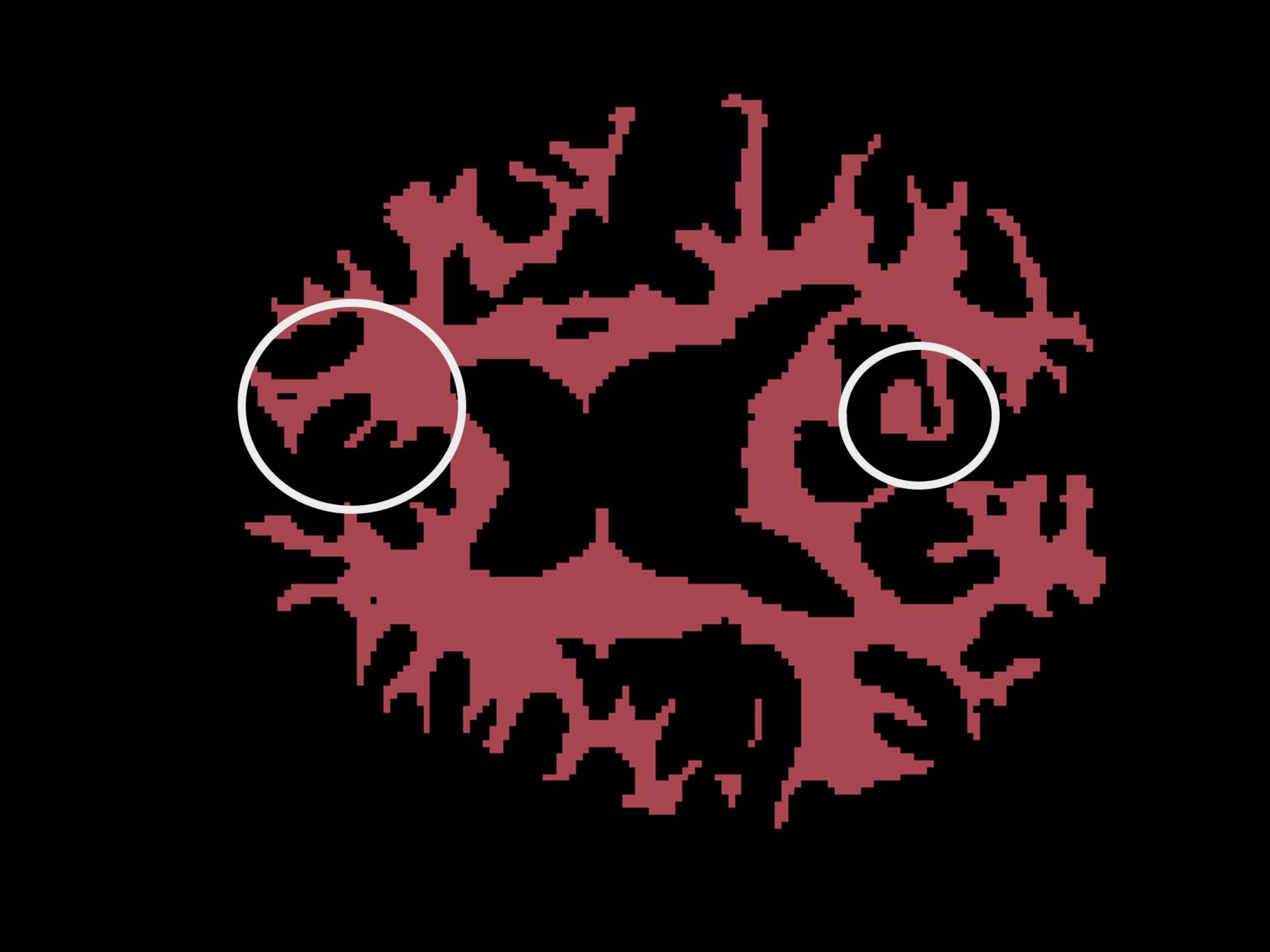}
		&\includegraphics[width=0.184\textwidth]{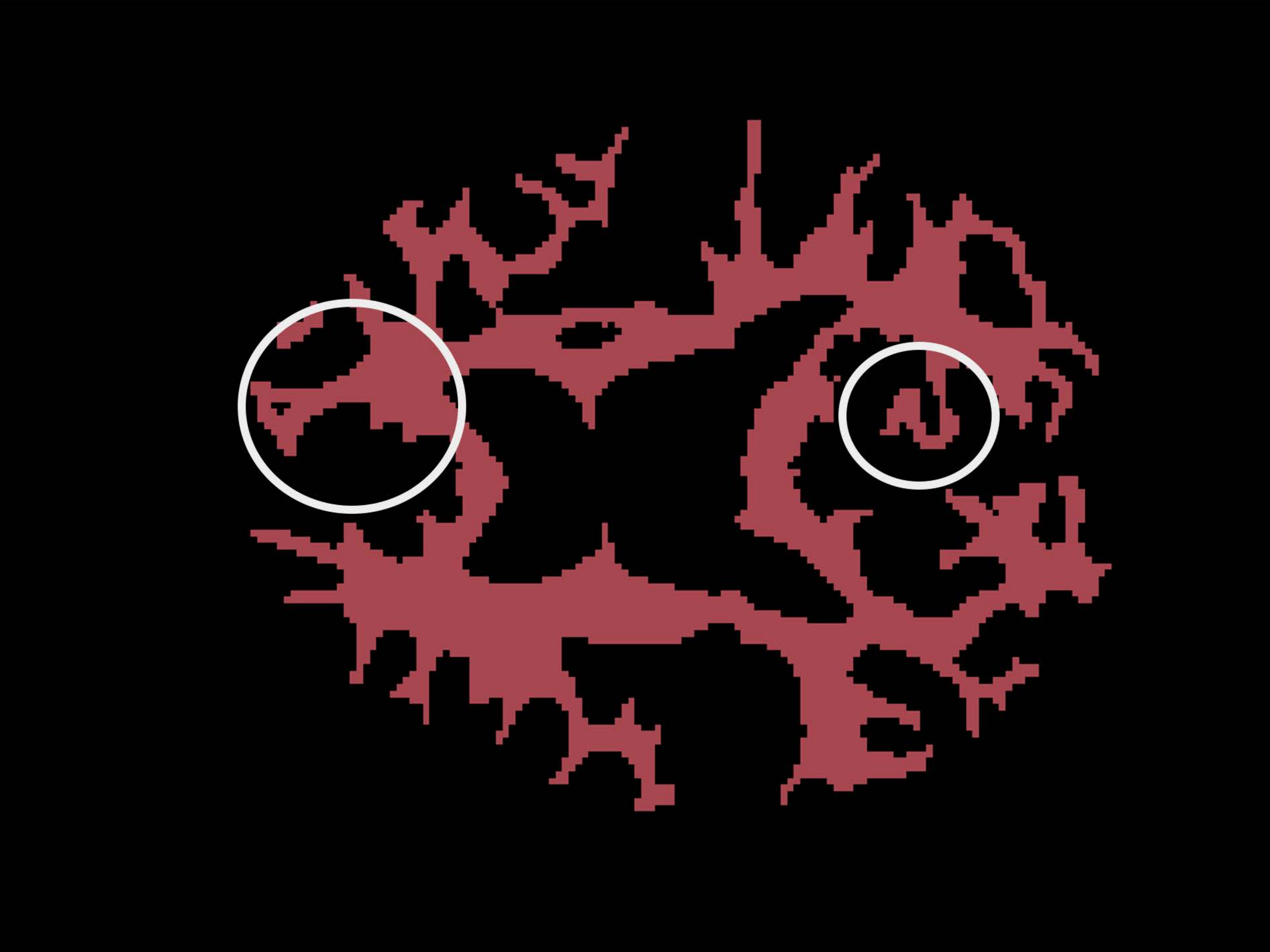} \\
		
		NLU-Net  & 	NLU-Net + Ours   & 	NLU-Net & 	NLU-Net + Ours  &  Ground Truth  \\
		
		(A)  & 	(A)   & (B)  & (B)  &    \\

	\end{tabular}
	\caption{ 
		Visualization of the segmentation results without and with data augmentation using our registration framework. The first row shows the original segmentation maps. The second and third rows show the segmentation maps for  GM and WM. The (A) refers to the experiment setup using one training data, and (B) represents the case of six training data. }
	\label{fig:seg_augmentation} 
\end{figure*}

\subsection{Applications on Multi-modal Data}
\label{sec:Multi-modal}

\textbf{Multi-modal registration experiments.}
We evaluate our method on datasets of brain scans acquired with T1 weighted and T2 weighted MRI, where T1 images do well to distinguish between different healthy tissues, whereas T2 images are best for highlighting abnormal structures in the brain such as tumors. 
Because of the complex intensity relationship between different modalities, in addition to default local normalized cross-correlation coefficient, we also use the multi-dimensional Modality Independent Neighborhood Descriptor (MIND)~\cite{HeinrichJBMGBS12} as the loss function for the multi-modal registration. These self-similarity context-based descriptors may transform images into representations independent of the underlying image acquisition and have been frequently used in multi-modal registration.

Fig.~\ref{fig:example_t1t2} first illustrates the examples of the multi-modal registration problem and the result alignments. As shown, the source images and labels are well aligned to the targets, although large deformations and complex intensity relationship between the two modalities exist.
Tab.~\ref{tab:compare_t1t2} depicts the accuracy and stability of the methods in terms of the Dice score on the different multi-modal setting.  Our method gives an obvious higher mean and a lower variance of Dice score on all the datasets, which indicates a more accurate and stable registration. 
These experiments demonstrate the capability of our framework to implicitly learn the complex similarity measure between different modalities.

\textbf{Multi-modal fusion experiments.}
We investigate the utility of our framework for assisting downstream image fusion tasks. Specifically, the registered multi-modal images with different registration algorithms are set as the input of the same fusion algorithm.
We introduce an image fusion algorithm~\cite{LiW19}, to fuse properties of medical images of MRI T1 and T2 brain images.  
We trained this fusion algorithm on the MR T1 and T2 multi-modal data in BraTS18.

Fig.~\ref{fig:fusion_t1t2} shows three groups of medical images with T1 and T2 MR. T1 images contain anatomical structure details, while T2 images provide normal and pathological content. 
In column (c), the fused images without pre-registration processing suffer from low contrast and lose some weak details. In column (d), the fused images obtained with pre-registration using NiftyReg may improve structure details and contrast but can not decrease the artifacts.  In column (e), the fused images obtained with pre-registration using our method can ideally improve structure details and contrast and greatly decrease the artifacts.

\begin{table}[t]
	\centering
	\caption{Ablation analysis of our efficient data augmentation strategy under two segmentation experiment settings in terms of Dice score. }
	\begin{tabular}{|m{1.0cm}<{\centering}| m{1.1cm}<{\centering}|m{2.2cm}<{\centering}| m{2.2cm}<{\centering}| }
		\hline
		\multicolumn{2}{|c |}{Methods}       &  NLU-Net      &  NLU-Net + Ours \\
		\hline
		\multirow{3}*{(A)} & CSF    &   0.902 $\pm$ 0.021& \textbf{0.919 $\pm$ 0.012}   \\
		~ & GM  &  0.885 $\pm$ 0.013  & \textbf{0.900 $\pm$ 0.009} \\
		~ & WM  &  0.851 $\pm$ 0.005  & \textbf{0.877 $\pm$ 0.008}    \\ 
		~ & Average  &  0.879 $\pm$ 0.009  & \textbf{0.899 $\pm$ 0.007}    \\ 
		\hline
		\multirow{3}*{(B)} & CSF      & 0.934 $\pm$ 0.017  & \textbf{0.939 $\pm$  0.011}   \\
		~ & GM     & 0.903 $\pm$ 0.008 & \textbf{0.913 $\pm$ 0.007}   \\ 
		~ & WM    & 0.888 $\pm$ 0.021 & \textbf{0.894 $\pm$ 0.013}   \\ 
		~ & Average   &  0.908 $\pm$ 0.009  & \textbf{0.915 $\pm$ 0.007}    \\ 
		\hline
	\end{tabular}
	\label{tab:ablation_seg}
\end{table}

\subsection{Applications on Medical Image Segmentation}
\label{sec:Seg}
We apply our framework to data augmentation of  segmentation. To be specific, we simulate deformations by sampling from the feature distribution, these deformations can be used to register labeled atlas images to produce more labels images for segmentation tasks. We use NLU-Net~\cite{WangZSJ20} as the baseline method. We conduct experiments on the dataset that comes from the ISeg19 and contains images and segmentation labels from 10 healthy 6-month-old infants.
By using the model without and with KL loss/feature distribution design, we may change the number of training data to get two experimental setups that use one training data  (A) and six training data (B). 
Tab.~\ref{tab:ablation_seg} provides the segmentation accuracy under two experiment settings in terms of Dice score. 
Fig.~\ref{fig:seg_augmentation} shows the example segmentation results of different variants.
These experimental results indicate that thanks to the data augmentation using our registration model, the segmentation models outperform previous baseline models significantly.

\section{Conclusions and Future Work}

We introduced a fundamental optimization model to formulate diffeomorphic registration and then established a learning framework to optimize it on a multi-scale feature space.
This framework may propagate learned multi-scale features and deep parameters for optimization, and thus could render fast optimization without needing iteratively computing gradients on the image domain.
We developed a series of deep modules to yield the multi-scale propagating process and to design the training objective.
This optimization perspective could differentiate our framework from naively cascading registration networks, and provide a computational interpretation of network architectures that guarantees diffeomorphism.
Moreover, we proposed our new bilevel self-tuned training, which allows the efficient search of the task-specific hyper-parameters and leads to the increased model flexibility and reduced computational burden.
Extensive experiments on image-to-atlas and image-to-image registration tasks showed that our method achieved state-of-the-art performance with diffeomorphic guarantee and extreme efficiency.
We also employed our framework to ideally solve challenging multi-modal registration tasks and investigated the utility of our framework to support the down-streaming image fusion and segmentation.

We demonstrate the performance on uni-modal and multi-modal registration tasks, and future validation remains on more challenging cross-modal registration and other scenarios where the moving and target exhibit more significant appearance differences.

\ifCLASSOPTIONcompsoc
  \section*{Acknowledgments}
\else
  \section*{Acknowledgment}
\fi

This work was partially supported by the National Key R\&D Program of China (2020YFB1313503), the National Natural Science Foundation of China (Nos. 61922019, 61733002, and 61672125), LiaoNing Revitalization Talents Program (XLYC1807088), and the Fundamental Research Funds for the Central Universities.
We thank Dr. Adrian V. Dalca at Massachusetts Institute of Technology for sharing the codes to generate the boxplots indicating Dice scores for anatomical structures.

\ifCLASSOPTIONcaptionsoff
  \newpage
\fi

\bibliographystyle{IEEEtran}
\bibliography{IEEEabrv,paper}

\begin{IEEEbiography}[{\includegraphics[width=1in,height=1.25in,clip,keepaspectratio]{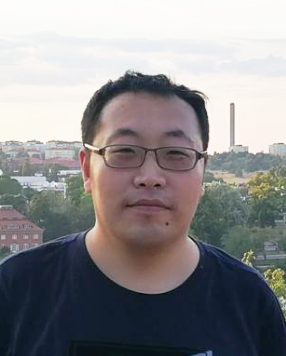}}]{Risheng Liu} received his B.Sc. (2007) and Ph.D. (2012) from Dalian University of Technology, China. From 2010 to 2012, he was doing research as joint Ph.D. in robotics institute at Carnegie Mellon University. From 2016 to 2018, He was doing research as Hong Kong Scholar at the Hong Kong Polytechnic University. He is currently a full professor with the Digital Media Department at International School of Information Science \& Engineering, Dalian University of Technology (DUT). He was awarded the ``Outstanding Youth Science Foundation" of the National Natural Science Foundation of China. He serves as editor for the Journal of Electronic Imaging (Senior Editor), The Visual Computer, and IET Image Processing. His research interests include optimization, computer vision and multimedia.
\end{IEEEbiography}

\begin{IEEEbiography}[{\includegraphics[width=1in,height=1.25in,clip,keepaspectratio]{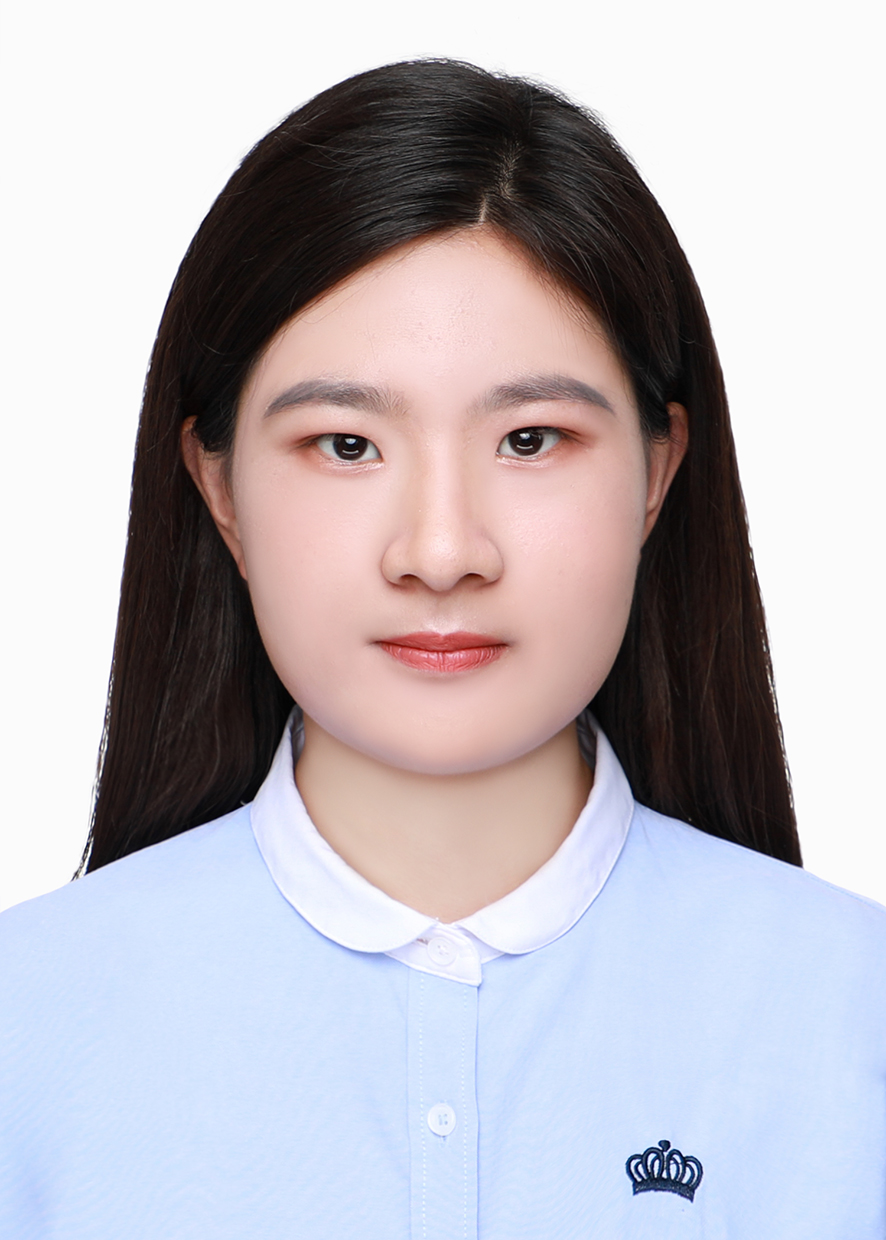}}]{Zi Li} received the B.E. degree in software engineering at Dalian University of Technology, Dalian, China, in 2019. She is currently pursuing a master's degree in software engineering at Dalian University of Technology, Dalian, China. Her research interests include medical image analysis, computer vision, deep learning.
\end{IEEEbiography}

\begin{IEEEbiography}[{\includegraphics[width=1in,height=1.25in,clip,keepaspectratio]{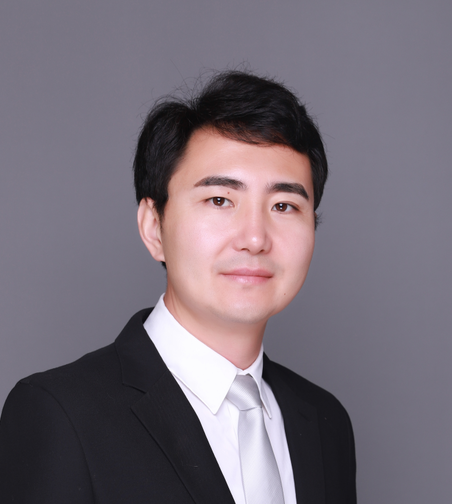}}]{Xin Fan} received the B.E. and Ph.D. degrees in information and communication engineering from Xian Jiaotong University, Xian, China, in 1998 and 2004, respectively. He was with Oklahoma State University, Stillwater, from 2006 to 2007, as a post-doctoral research Fellow. He joined the School of Software, Dalian University of Technology, Dalian, China, in 2009. His current research interests include computational geometry and machine learning, and their applications to low-level image processing and DTI-MR image analysis.
\end{IEEEbiography}

\begin{IEEEbiography}[{\includegraphics[width=1in,height=1.25in,clip,keepaspectratio]{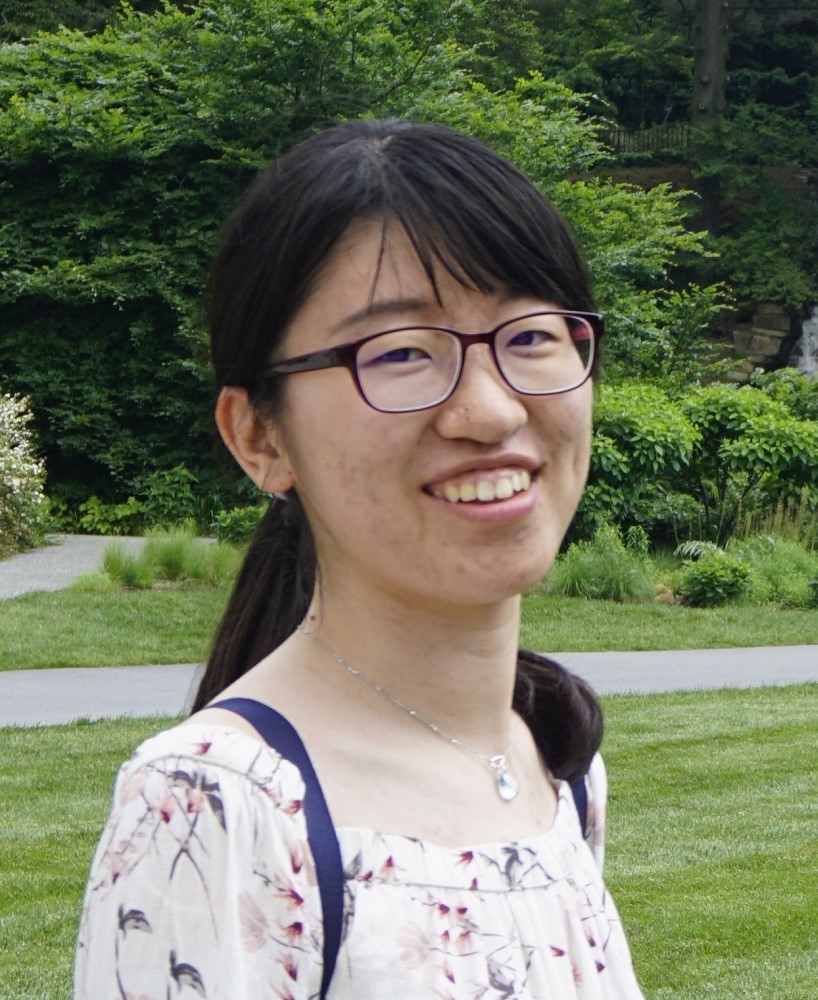}}]{Chenying Zhao} received the B.Eng. degree in Biomedical Engineering from Tsinghua University in 2017. Currently she is a Bioengineering Ph.D. student at the University of Pennsylvania. Her research interests include diffusion MRI, brain connectome, medical image analysis, and applications in pediatric population. She was the recipient of the International Society for Magnetic Resonance in Medicine (ISMRM) Summa Cum Laude Merit Award in 2020.
\end{IEEEbiography}

\begin{IEEEbiography}[{\includegraphics[width=1in,height=1.25in,clip,keepaspectratio]{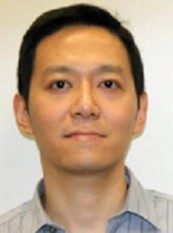}}]{Hao Huang} received the B.E. from Tsinghua University, China, in 1996 and the M.S. from Peking University, China, in 1999. In 2004 and 2005, he received the M.S.E in Computer and Electrical Engineering and the PhD in Biomedical Engineering from the Johns Hopkins University.
He continued to be a Research Associate Faculty at Johns Hopkins University School of Medicine from 2005 to 2007. In 2007, he became an Assisant Professor at University of Texas Southwestern Medical Center. From 2014 to now, he has been a faculty member at University of Pennsylvania, where he became an Associate Professor in 2014, and a Professor in 2021. 
His research is primarily focused on pushing the technical boundaries of neural magnetic resonance imaging (MRI) in health and disease, including advanced MR acquisition and analysis techniques in diffusion MRI, functional MRI and perfusion MRI. He has served in a number of leadership positions in international committees. He has been recognized as the Distinguished Investigator of the Academy for Radiology and Biomedical Imaging Research in 2019. He has been elected as the Fellow of American Institute of Medical and Biological Engineering (AIMBE) in 2021.
\end{IEEEbiography}

\begin{IEEEbiography}[{\includegraphics[width=1in,height=1.25in,clip,keepaspectratio]{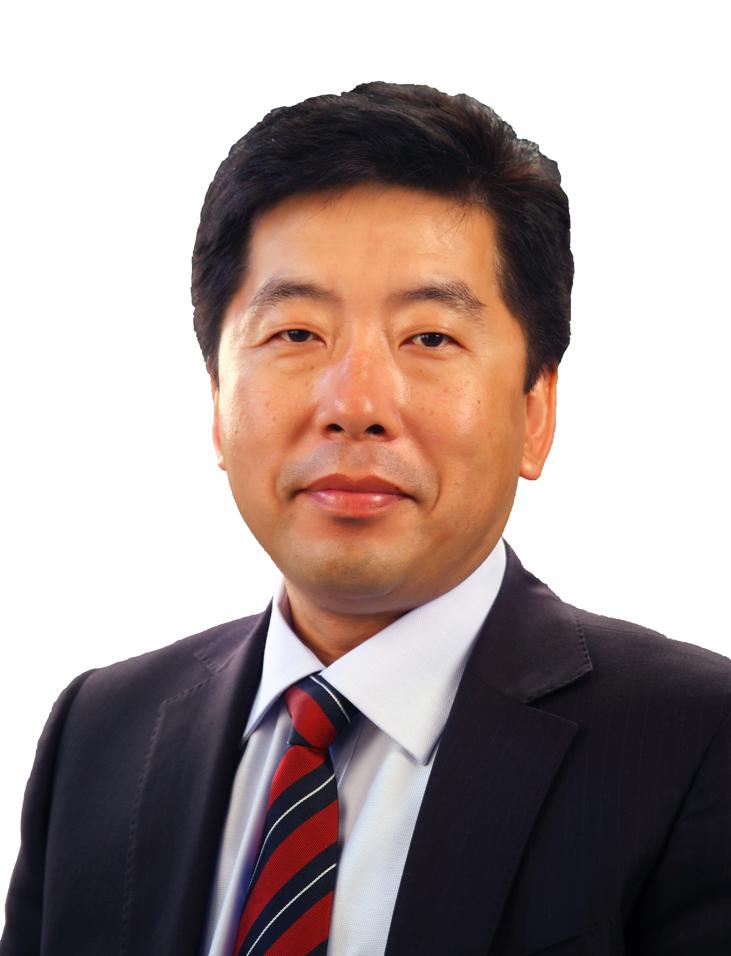}}]{Zhongxuan Luo} received the B.S. degree in Computational Mathematics from Jilin University, China, in 1985, the M.S. degree in Computational
Mathematics from Jilin University in 1988, and the Ph.D. degree in Computational Mathematics from Dalian University of Technology, China, in 1991. He has been a full professor of the School of Mathematical Sciences at Dalian University of Technology since 1997. His research interests include computational geometry and computer vision.
\end{IEEEbiography}

\end{document}